\documentclass{article}

\usepackage{microtype}
\usepackage{graphicx}
\usepackage{subfigure}
\usepackage{booktabs} 
\usepackage{multirow}
\usepackage{color}
\usepackage{colortbl}
\usepackage{enumitem}
\usepackage{pifont}
\usepackage{bbding}
\usepackage[framemethod=TikZ]{mdframed}
\usepackage{placeins}
\usepackage{soul}

\usepackage{hyperref}

\usepackage[accepted]{icml2024}

\usepackage{amsmath}
\usepackage{amssymb}
\usepackage{mathtools}
\usepackage{amsthm}
\usepackage{lipsum}
\usepackage{titletoc}
\usepackage{minitoc}
\usepackage[page,header]{appendix}

\usepackage[capitalize,noabbrev]{cleveref}
\crefname{figure}{Fig.}{Figs.}
\Crefname{figure}{Fig.}{Figs.}
\crefname{equation}{Eq.}{Eqs.}
\Crefname{equation}{Eq.}{Eqs.}
\crefformat{equation}{#2Eq.~(#1)#3}
\Crefformat{equation}{#2Eq.~(#1)#3}

\crefname{table}{Tab.}{Tabs.}
\Crefname{table}{Tab.}{Tabs.}

\crefname{section}{Sec.}{Secs.}
\Crefname{section}{Sec.}{Secs.}

\crefname{lemma}{Lem.}{Lems.}
\Crefname{lemma}{Lem.}{Lems.}

\crefname{remark}{Remark}{Rem.}
\Crefname{remark}{Remark}{Rems.}

\crefname{proposition}{Proposition}{Prop.}
\Crefname{proposition}{Proposition}{Props.}

\crefname{definition}{Definition}{Def.}
\Crefname{definition}{Definition}{Defs.}
\crefname{appendix}{App.}{App.}

\usepackage{tcolorbox}
\usepackage{makecell}
\tcbuselibrary{skins}
\newenvironment{qbox}
{\begin{tcolorbox}[enhanced jigsaw, drop shadow=black!50!white,colback=white, width=0.95\linewidth, center, left=2pt,right=2pt,top=1pt,bottom=1pt]}
{\end{tcolorbox}}

\theoremstyle{plain}
\newtheorem{theorem}{Theorem}[section]
\newtheorem{proposition}[theorem]{Proposition}
\newtheorem{lemma}[theorem]{Lemma}

\theoremstyle{definition}
\newtheorem{definition}[theorem]{Definition}

\theoremstyle{remark}

\definecolor{myblue}{RGB}{235, 244, 246}
\definecolor{mygreen}{RGB}{234, 240, 225}

\newcommand{\MAE}{\mathsf{MAE}}
\newcommand{\F}{\mathsf{F}}
\newcommand{\SMAE}{\mathsf{SI\text{-}MAE}}
\newcommand{\SF}{\mathsf{SI\text{-}F}}
\newcommand{\SAUC}{\mathsf{SI\text{-}AUC}}
\newcommand{\AUC}{\mathsf{AUC}}
\newcommand{\E}{\mathsf{E}}
\newcommand{\CE}{\mathsf{CE}}
\newcommand{\BCE}{\mathsf{BCE}}
\newcommand{\MSE}{\mathsf{MSE}}
\newcommand{\IOU}{\mathsf{IOU}}
\newcommand{\TP}{\mathsf{TP}}
\newcommand{\FP}{\mathsf{FP}}
\newcommand{\FN}{\mathsf{FN}}
\newcommand{\TN}{\mathsf{TN}}
\newcommand{\TPR}{\mathsf{TPR}}
\newcommand{\FPR}{\mathsf{FPR}}
\newcommand{\FNR}{\mathsf{FNR}}

\newcommand{\SI}{\mathsf{SI}}

\usepackage[textsize=tiny]{todonotes}

\icmltitlerunning{Size-invariance Matters: Rethinking Metrics and Losses for Imbalanced Multi-object Salient Object Detection}

\begin{document}

\twocolumn[
\icmltitle{Size-Invariance Matters: Rethinking Metrics and Losses for \\ Imbalanced Multi-object Salient Object Detection}



\icmlsetsymbol{equal}{*}

\begin{icmlauthorlist}
\icmlauthor{Feiran Li}{iie,ucas-scs}
\icmlauthor{Qianqian Xu}{ict}
\icmlauthor{Shilong Bao}{iie,ucas-scs}
\icmlauthor{Zhiyong Yang}{ucas-cs}

\icmlauthor{Runmin Cong}{bju,sdu-cse,misc}
\icmlauthor{Xiaochun Cao}{sysu}
\icmlauthor{Qingming Huang}{ucas-cs,ict,bdkm}
\end{icmlauthorlist}

\icmlaffiliation{ict}{Key Laboratory of Intelligent Information Processing, Institute of Computing Technology, Chinese Academy of Sciences, Beijing, China}
\icmlaffiliation{ucas-cs}{School of Computer Science and Technology, University of Chinese Academy of Sciences, Beijing, China}
\icmlaffiliation{iie}{Institute of Information Engineering, Chinese Academy of Sciences, Beijing, China}
\icmlaffiliation{ucas-scs}{School of Cyber Security, University of Chinese Academy of Sciences, Beijing, China}
\icmlaffiliation{bdkm}{Key Laboratory of Big Data Mining and Knowledge Management, Chinese Academy of Sciences, Beijing, China}
\icmlaffiliation{sysu}{School of Cyber Science and Tech., Shenzhen Campus, Sun Yat-sen University}
\icmlaffiliation{bju}{Institute of Information Science, Beijing Jiaotong University, Beijing, China}
\icmlaffiliation{sdu-cse}{School of Control Science and Engineering, Shandong University, Jinan, China}
\icmlaffiliation{misc}{Key Laboratory of Machine Intelligence and System Control, Ministry of Education, Jinan, China}

\icmlcorrespondingauthor{Qianqian Xu}{xuqianqian@ict.ac.cn}
\icmlcorrespondingauthor{Qingming Huang}{qmhuang@ucas.ac.cn}

\icmlkeywords{Salient Object Detection}

\vskip 0.3in
]



\printAffiliationsAndNotice{}  

\begin{abstract}

This paper explores the size-invariance of evaluation metrics in Salient Object Detection (SOD), especially when multiple targets of diverse sizes co-exist in the same image.
We observe that current metrics are size-sensitive, where larger objects are focused, and smaller ones tend to be ignored. 
We argue that the evaluation should be size-invariant because bias based on size is unjustified without additional semantic information.
In pursuit of this, we propose a generic approach that evaluates each salient object separately and then combines the results, effectively alleviating the imbalance.
We further develop an optimization framework tailored to this goal, achieving considerable improvements in detecting objects of different sizes. 
Theoretically, we provide evidence supporting the validity of our new metrics and present the generalization analysis of SOD. 
Extensive experiments demonstrate the effectiveness of our method. The code is available at \url{https://github.com/Ferry-Li/SI-SOD}.
\end{abstract}

\section{Introduction}
Salient object detection (SOD), also known as salient object segmentation, aims at highlighting visually salient regions in images ~\cite{IDSurvey_2022}. To achieve this, a SOD model typically processes an RGB image to generate a binary mask, marking each pixel as either salient (1) or not (0). Recently, SOD has witnessed great progress in various applications~\cite{tracking_2009,scene_2014,tang2017Tri,li2019deep,zhang2020causal,jiang2023Hierarchical,tang2024remote}.


\begin{figure}[t]
\centering
\subfigure[Average size of objects]{   
\begin{minipage}{0.47\linewidth}
\includegraphics[width=\linewidth]{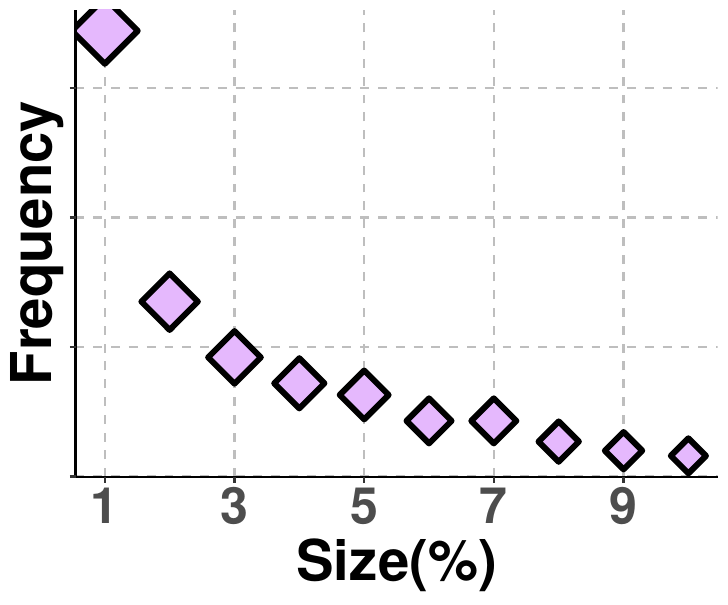}  
\label{fig:msod_area}
\end{minipage}
}
\subfigure[Average number of objects]{   
\begin{minipage}{0.47\linewidth}
\includegraphics[width=\linewidth]{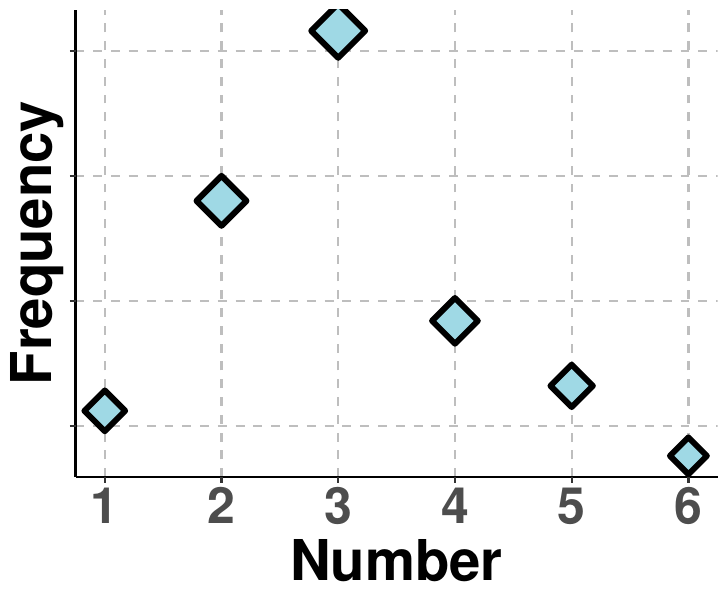}  
\label{fig:msod_num}
\end{minipage}
}
\vspace{-0.3cm}
\caption{Statistics on dataset MSOD. \cref{fig:msod_area} illustrates the widely existing small salient objects,  with \textit{Size(\%)} as the proportion of the size of an object over the whole image.\cref{fig:msod_num} reveals that practical SOD scenarios usually involve multiple salient objects.}
\label{fig:msod_num_area}
\end{figure}

 The progress of SOD primarily depends on two factors. One is the development of sophisticated models (say deep neural networks), which effectively disentangle diverse feature patterns for accurate SOD detection. Notable methods include ~\cite{EDN,Luo_2017_CVPR,MENet,Ma_2021_AAAI,zhang_2021_nips}. The other is the evaluation and selection of the best models for practical applications. 
 Generally, a well-performed SOD model should simultaneously embrace a high \text{True Positive Rate ($\TPR$)} and a low \text{False Positive Rate ($\FPR$)} \cite{Borji_2019_survey}. To this end, various metrics (typically $\MAE$ and $\F$-score) have been widely considered for evaluation and optimization \cite{chen_2021_ijcv,sun_2022_ijcv}. 
 
 In this paper, we argue that current evaluation metrics are size-sensitive, which is not a proper choice for SOD tasks when the sizes of objects in a given image are highly imbalanced. As demonstrated in \cref{fig:msod_num_area}, SOD tasks typically involve \ul{multiple salient objects with diverse sizes}. In this sense, prediction errors would be dominated by those larger objects, leading models to \ul{overlook small salient objects}. Taking $\MAE$ as an example, \cref{fig:eg_a} could merely detect the larger salient object but miss the smaller one on the right, while \cref{fig:eg_b} could successfully capture all salient objects. However, \cref{fig:eg_b} induces a worse $\MAE$ than \cref{fig:eg_a}, which is \textbf{counter-intuitive} to our visual perceptions. Large objects dominate size-sensitive metrics, consequently leading to practical performance degradation because there are many cases where small objects are critical for downstream tasks. For example, in a street view, traffic lights are usually of small size, but they play a significant role in autonomous driving tasks.
 
To address the issues above, we are interested in the following problem:

\begin{figure}[t]
\centering
\subfigure[Image]{
\begin{minipage}{0.22\linewidth}
\includegraphics[width=\linewidth]{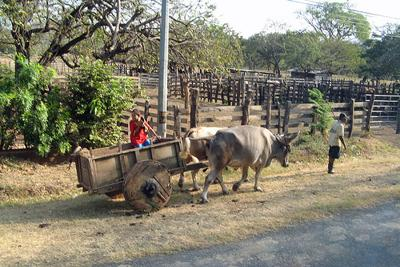}  
\label{fig:eg_image}
\end{minipage}
}
\subfigure[Label]{
\begin{minipage}{0.22\linewidth}
\includegraphics[width=\linewidth]{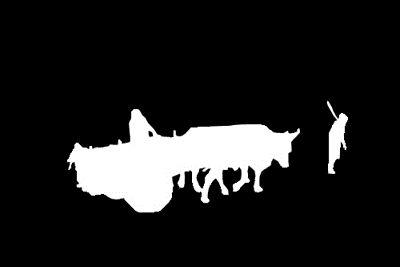}  
\label{fig:eg_gt}
\end{minipage}
}
\subfigure[EDN]{
\begin{minipage}{0.22\linewidth}
\includegraphics[width=\linewidth]{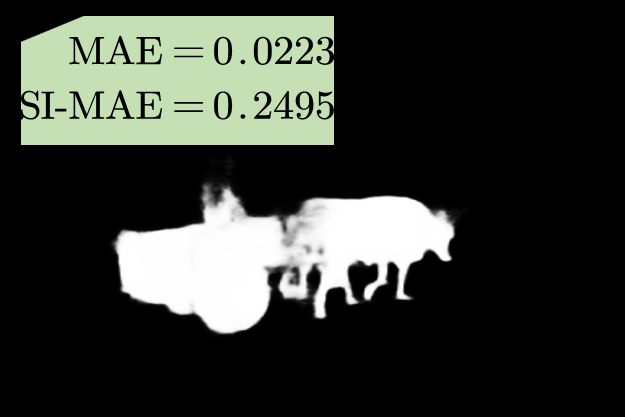}  
\label{fig:eg_a}
\end{minipage}
}
\subfigure[EDN+Ours]{
\begin{minipage}{0.22\linewidth}
\includegraphics[width=\linewidth]{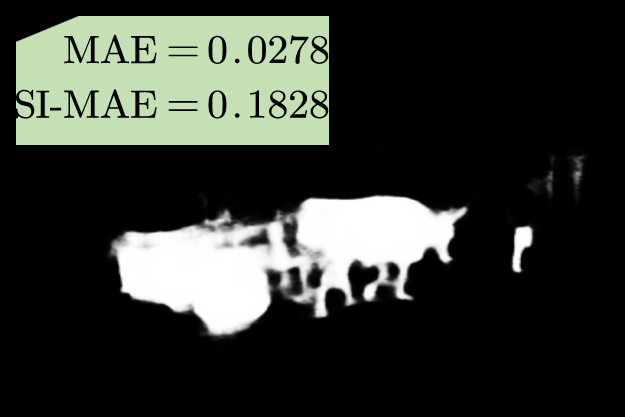}  
\label{fig:eg_b}
\end{minipage}
}
\caption{(c) is the result of backbone EDN~\cite{EDN}, and (d) is the prediction optimized by our approach. (c) detects fewer salient objects, yet enjoys lower $\MAE$ than (d). However, $\SMAE$ can correctly distinguish two detections.}
\label{fig:eg}
\end{figure}

\begin{qbox}
\begin{center}
\textbf{\textit{Can we develop an effective size-invariant criterion for imbalanced multi-object SOD?}}
\end{center}
\end{qbox}
The answer is affirmative in this paper. To begin with, we present a novel unified framework to understand why popular SOD metrics are size-sensitive. Specifically, given an image, we show that common criterion can be reformulated as a weighted (denoted by $\color{blue}{\mathbb{P}_{x_i}}$) sum of multiple independent parts, with each weighted term $\color{blue}{\mathbb{P}_{x_i}}$ being \textbf{highly related to the size} of the corresponding part. This creates an inductive bias toward objects of different sizes. 

Motivated by this, we thus propose a simple yet effective paradigm for size-invariant SOD evaluation. The key idea is to modify the size-related term $\color{blue}{\mathbb{P}_{x_i}}$ into a size-invariant constant, ensuring equal treatment for each salient object regardless of size. Meanwhile, we introduce a generic Size-Invariant SOD (SI-SOD) optimization loss to pursue our size-invariant goal practically.  

To show the effectiveness of our proposed paradigm, we then investigate the generalization performance of the SI-SOD algorithm. To the best of our knowledge, such a problem \textbf{remains barely explored} in the SOD community.  As a result, we find that for composite losses (defined in \cref{Revisting}), the size-invariant loss function leads to a sharper bound than its size-sensitive counterparts.

Finally, extensive experiments over a range of benchmark datasets speak to the efficacy of our proposed method.

\section{Related Work} \label{related_work}
In recent years, SOD achieved considerable progress with elaborate frameworks and well-designed losses. We give a brief overview of SOD methods here and a detailed description of evaluation metrics in \cref{protocol_appendix}.



\textbf{Architecture-focused} methods usually adopt convolutional networks as basic modules since their great success. For example, UCF~\cite{UCF} introduced a reformulated dropout after specific convolutional layers to learn deep uncertain convolutional features. DCL~\cite{DCL} adopted a multi-stream framework, with the pixel-level fully convolutional stream to improve pixel-level accuracy. A common way to extract multi-level features is to design a bottom-up/top-down architecture, which resembles the U-Net~\cite{U-Net}. PiCANet~\cite{PiCANet} proposed a pixel-wise contextual attention network to selectively attend to informative context locations for each pixel and embed global and local networks into a U-Net architecture. RDCPN~\cite{RDCPN} introduced a novel multi-level ROIAlign-based decoder to adaptively aggregate multi-level features for better mask predictions. Similar structures are also utilized in recent works, including EDN~\cite{EDN}, ICON~\cite{ICON}, Bi-Directional~\cite{Bi-Directional}, CANet~\cite{CANet}, etc. ~\cite{piao2019depth} designed a refinement block to fully extract and fuse multi-scale features, successfully achieving excellent performance on most datasets. Based on this, ~\cite{ji2022dmra} further exploited a cascaded hierarchical feature fusion strategy to promote efficient information interaction of multi-level contextual features and efficiently improve contextual represent ability.

\textbf{Multi-source-based} methods have recently become popular. Specifically, both PoolNet~\cite{PoolNet} and MENet~\cite{MENet} conducted joint supervision of salient objects and object boundaries at each side-output. ~\cite{ji2023multispectral} used thermal infrared images as extra input to deal with rainy, overexposure, or low-light occasions, and achieved effective results. Depth information is also widely used in SOD, which is usually named as RGB-D SOD. For instance,~\cite{ji2020accurate,zhang2023c,li2023dvsod,li2023delving} introduced depth map to SOD and significantly improved the detection performance. Furthermore, ~\cite{zhang2019memory,zhang2020lfnet} utilized light field data as an auxiliary for SOD and achieved state-of-the-art performance at that time. Some extensive works such as ~\cite{zhang2021dynamic,ji2023multispectral,li2023dvsod} successfully deal with video SOD tasks exploiting the inter-frame information.

There are also previous works analyzing the evaluation in SOD. ~\cite{Bylinskii2019Measurement} provided a comprehensive analysis of eight different evaluation metrics and their properties. ~\cite{Borji2013Quantitative} performed a comparison of dozens of methods on many datasets to explore the consistency between the model ranking and practical performance. However, little attention has been paid to occasions where multiple salient objects co-exist, which is quiet common in the real world.

\section{A Novel Size-invariant Evaluation Protocol}
\label{Size-invariant Metrics}
In this section, we begin by discussing why the commonly used metrics, such as Mean Absolute Error ($\MAE$) and $\F$-score, are not suitable for evaluating on imbalanced multi-object occasions. We then introduce methods to improve these metrics, aiming for a size-invariant SOD evaluation.

\subsection{Revisiting Current SOD Evaluation Metrics} \label{Revisting}

We start our analysis from standard functions, which could be divided into two groups: \textit{separable} and \textit{composite} functions, expressed as follows:

\begin{mdframed}[hidealllines=true,backgroundcolor=mygreen,innerleftmargin=3pt,innerrightmargin=3pt,leftmargin=-3pt,rightmargin=-3pt]
\begin{definition}[\textit{Separable Function}]\label{def3.1}
Given a predictor $f$, a function $g$ applied to $f$ is separable if the following equation formally holds:
    \begin{equation} \label{eq:separable function}
        g(f(X), Y)=\sum_{i=1}^n w_{X_i}\cdot g(f(X_i), Y_i),
    \end{equation}
    with 
    \begin{equation}
        \bigcup_{i=1}^n X_i = X, \; \bigcap_{i=1}^n X_i = \varnothing,
    \end{equation}
    where $X$ is the input and $Y$ is the ground truth; $(X_1, X_2, \cdots, X_n)$ are $n$ non-intersect parts of $X$; and $w_{X_i}$ is an $X_i$-related weight for the term $g(f(X_i), Y_i)$. 
\end{definition}
\end{mdframed}
According to the definition above, we realize that the point-wise evaluation metrics in the SOD community are separable (say Mean Absolute Error ($\MAE$)~\cite{MAE} and Mean Square Error ($\MSE$)).
\begin{mdframed}[hidealllines=true,backgroundcolor=mygreen,innerleftmargin=3pt,innerrightmargin=3pt,leftmargin=-3pt,rightmargin=-3pt]
    \begin{definition}[\textit{Composite Function}]\label{def3.2}
\textit{Composite Functions} are a series of compositions of separable functions \cref{eq:separable function}, denoted by
 \begin{align*}
    G(f(X), Y)=(g_1\circ g_2 \circ \dots \circ g_T)\left(f(X),Y\right),
 \end{align*}
 where $T$ is the number of compositions.
\end{definition}
\end{mdframed}
According to the definition above, complicated evaluation metrics such as $\F$-score \cite{F_measure}, $\IOU$ \cite{RCNN} and $\AUC$ \cite{AUC_eval} are composite. In what follows, we will discuss each of them respectively. For simplicity, we abbreviate $g(f(X), Y)$ and $G(f(X), Y)$ as $g(f)$ and $G(f)$ for a clear presentation.

\textbf{Current separable metrics are NOT size-invariant.} In SOD, the model $f: \mathbb{R}^S \to \mathbb{R}^S$ takes an image $X$ with label $Y$ as input, aiming to make a binary classification for each pixel, where $S$ is the size of the image and $Y = \{0, 1\}^S$ is the pixel-level ground-truth. In light of this, the image could be naturally divided into $N_c$ parts based on the location of salient objects. 

Therefore, given a certain separable SOD metric $g$, let $g(f_i):=g(f(X_i), Y_i)$, we can rewrite it as \cref{eq:separable function} does:
\begin{equation}
\label{eq:separable-sensitive}
    g(f)=\sum_{i=1}^{N_c} {\color{blue}{\mathbb{P}_{X_i}}} \cdot g(f_i),
\end{equation}
where $\mathbb{P}_{X_i}$ is the \textbf{size-sensitive} weight for the $i$-th part of $X$, which brings about inductive bias in evaluation.

Taking $\MAE$ as an example, the following equation holds:
\begin{equation}
\label{eq:MAE_sensitive}
    \begin{aligned}
        \MAE (f) & = \sum_{i=1}^{N_c}  \frac{\Vert f(X_i)-Y_i \Vert_{1,1}}{S} \\
        & = \sum_{i=1}^{N_c} \frac{S_i}{S} \cdot \frac{\Vert f(X_i)-Y_i \Vert_{1,1} }{S_i} \\
        & = \sum_{i=1}^{N_c} \frac{S_i}{S} \cdot \MAE(f_i) \\
        & = \sum_{i=1}^{N_c} {\color{blue}{\mathbb{P}_{X_i}}} \cdot \MAE(f_i).
    \end{aligned}
\end{equation}
Here we have $\color{blue}{\mathbb{P}_{X_i}=S_i/S}$, where $S_i$ represents the size of the $i$-th part. It is explicitly that the current $\MAE$ metric for SOD is size-sensitive, where \textbf{larger objects would be paid more attention}. Similar results can be drawn for other point-wise metrics in the SOD community.

\textbf{Current composite metrics are NOT size-invariant.} Similarly, we formally rewrite the composite metric $G(f)$ as follows:
\begin{equation}
\label{eq:composite}
    G(f)=\frac{\sum_{i=1}^{N_c} {\color{blue}{\mathbb{P}_{X_i}}} (a_1g_1(f_i)+\cdots+a_Tg_T(f_i))}{\sum_{i=1}^{N_c} {\color{blue}{\mathbb{P}_{X_i}}} (b_1g_1(f_i)+\cdots+b_Tg_T(f_i))},
\end{equation}
where again $g_t(f_i):=g_t(f(X_i),Y_i), t\in [T]$ is a certain separable metric value over the $i$-th part $X_i$; $a_i$ and $b_i$ represent coefficients for different composite functions, and here $\mathbb{P}_{X_i}$ is also a \textbf{size-sensitive} weight for each separable part of $X$.

Specifically, in terms of the widely used $\F$-score \cite{F_measure}, we have:
\begin{equation}
\label{eq:f-sensitive}
    \begin{aligned}
    & \F(f) = \frac{2\sum_{i=1}^{N_c} \TP(f_i)}{\sum_{i=1}^{N_c} [ 2\TP(f_i) + \FP(f_i)+ \FN(f_i)]} \\
    & = \frac{2 \sum_{i=1}^{N_c} \frac{S_i}{S}\cdot \frac{\TP(f_i)}{S_i}  }{\sum_{i=1}^{N_c} [ \frac{S_i}{S} \cdot (2\frac{\TP(f_i)}{S_i} + \frac{\FP(f_i)}{S_i}+ \frac{\FN(f_i)}{S_i})]} \\
    & = \frac{2\sum_{i=1}^{N_c} {\color{blue}{\mathbb{P}_{X_i}}}\cdot \TPR(f_i) }{\sum_{i=1}^{N_c} {\color{blue}{\mathbb{P}_{X_i}}} \cdot (2\TPR(f_i)+\FPR(f_i)+\FNR(f_i))},
    \end{aligned}
\end{equation}
where $\TP(f_i)$, $\FP(f_i)$, $\FN(f_i)$ represent the number of \textbf{T}rue \textbf{P}ositives, \textbf{F}alse \textbf{P}ositives and \textbf{F}alse \textbf{N}egatives within $X_i$, and $\TPR(f_i), \FPR(f_i), \FNR(f_i)$ represent the corresponding \textbf{T}rue \textbf{P}ositive \textbf{R}ate, \textbf{F}alse \textbf{P}ositive \textbf{R}ate, and \textbf{F}alse \textbf{N}egative \textbf{R}ate, which are all separable functions mentioned above. 
In this case, we still have $\color{blue}{\mathbb{P}_{X_i}=S_i/S}$, which is sensitive to the size of salient objects. Similar conclusion also applies to metrics like $\AUC$, with analysis in \cref{wAUC_metric}.

\textbf{Why size-invariance MATTERS?} We have realized that the existing widely adopted metrics would inevitably introduce biased weights for objects of different sizes. With this imbalance, smaller objects are suppressed by larger ones, and therefore are easily ignored in both evaluation and prediction. Unfortunately, as shown in \cref{fig:msod_num_area}, practical SOD tasks usually involve multiple objects of various sizes, including small yet critical ones. For example, \cref{fig:eg_a} totally overlooks a small object, but enjoys a similar $\MAE$ compared to \cref{fig:eg_b}, which contradicts our visual perceptions. To rectify this, we introduce the principles of size-invariant evaluation in the next section.



\subsection{Principles of Size-Invariant Evaluation} \label{principles of SI_eval}
Based on the discussions above, the fundamental limitation of the current evaluation lies in the size-sensitive $\color{blue}{\mathbb{P}_{X_i}}$. Therefore, a principal way to achieve size-invariant evaluation is to eliminate the effect of the weighting term $\color{blue}{\mathbb{P}_{X_i}}$. 
In this paper, we propose a simple yet effective size-invariant protocol:
\begin{equation}
\label{eq:separable-insensitive}
    g_{\SI}(f)=\frac{1}{N_c}\sum_{i=1}^{N_c} {\color{orange}{1}} \cdot g(f_i),
\end{equation}
\begin{equation}
\label{eq:SI-composite}
    G_{\SI}(f)=\frac{1}{N_c}\sum_{i=1}^{N_c} {\color{orange}{1}} \cdot \frac{(a_1g_1(f_i)+\cdots+a_Tg_T(f_i))}{(b_1g_1(f_i)+\cdots+b_Tg_T(f_i))}, 
\end{equation}
where $\color{blue}{\mathbb{P}_{X_i}}$ is replaced by a constant $\color{orange}{1}$. The size-sensitive weight is directly eliminated, and we naturally arrive at size-invariance.

In what follows, we will adopt widely used metrics, i.e., $\MAE$, $\F$-score and $\AUC$, to instantiate our size-invariant principles. Note that our proposed strategy could also be applied to other metrics as mentioned in \cref{SI-SOD}.

\subsubsection{Size-Invariant $\MAE$} \label{SI-MAE}
According to \cref{eq:separable-insensitive}, $\SMAE$ is expressed as follows:
\begin{equation}
\label{naive Size-invariant}
    \SMAE(f)=\frac{1}{N_c} \sum_{i=1}^{N_c} {\color{orange}{1}} \cdot \MAE(f_i).
\end{equation}

Here our primary focus is on dividing the image into $N_c$ parts. Motivated by the success of object detection~\cite{FasterRCNN}~\cite{YOLO}, we segment salient objects into a series of foreground frames by their minimum bounding boxes, and pixels that do not form part of any bounding box are treated as the background.

Ideally, assume that there are $K$ salient objects in an image and let $C_k=\{(a_j,b_j)\}_{j=1}^{M_k}, k\in [K]$ be the coordinate set for the object $k$, then the minimum bounding box for object $k$ could be determined clockwise by the following vertex coordinates: 
\begin{equation} \label{eq:bounding_box}
\begin{aligned}
    X_k^{fore}= & \{(a_k^{min}, b_k^{max}), (a_k^{max}, b_k^{max}), \\ 
    & (a_k^{max}, b_k^{min}), (a_k^{min},b_k^{min})
    \},
\end{aligned}
\end{equation}
where $a_k^{min}, a_k^{max}, b_k^{min}, b_k^{max}$ are the minimum and maximum coordinates in $C_k$, respectively.

Correspondingly, the background frame is defined as follows:
\begin{equation}
    X^{back}_{K+1} = X \setminus F,
\end{equation}
where 
\begin{equation}
    F = \bigcup_{k=1}^K X_k^{fore}
\end{equation}
is the collection of all minimum bounding boxes for salient objects. 

\begin{figure}[t]
\centering
\subfigure[Single-object scenario]{
\begin{minipage}{0.465\linewidth}
\includegraphics[width=\linewidth]{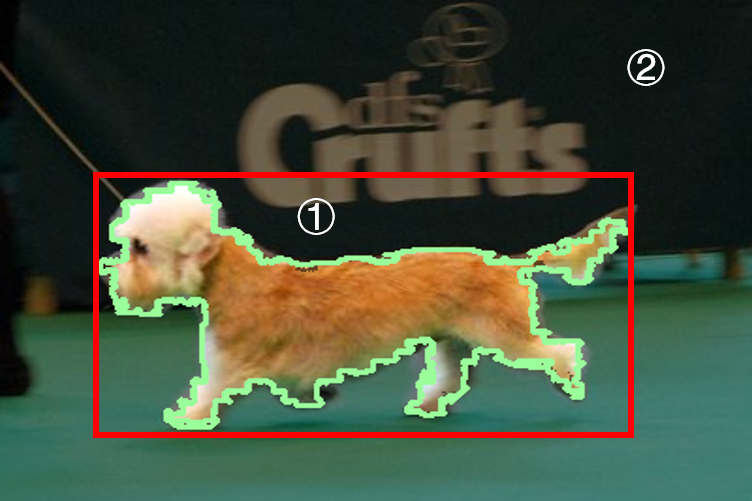}  
\label{fig:object_frame1}
\end{minipage}
}
\subfigure[Multi-object scenario]{
\begin{minipage}{0.47\linewidth}
\includegraphics[width=\linewidth]{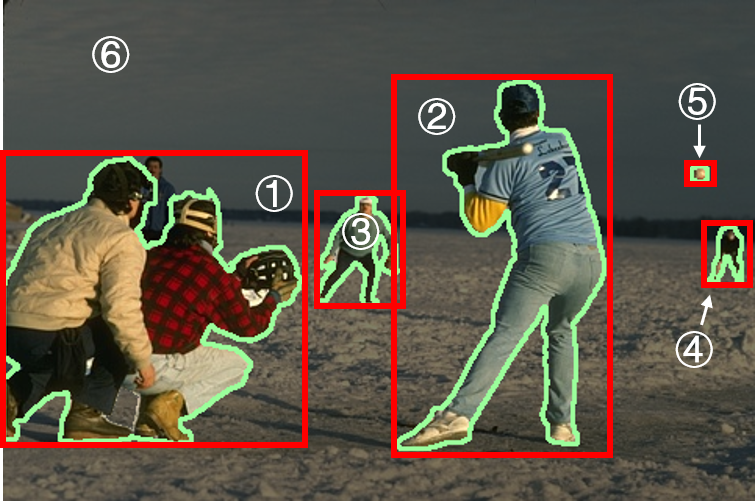}  
\label{fig:object_frame2}
\end{minipage}
}
\caption{Examples of partitions. In \cref{fig:object_frame1}, there is a foreground frame \ding{192} and a background frame \ding{193}. In \cref{fig:object_frame2}, there are five foreground frames from \ding{192} to \ding{196}, and a background frame \ding{197}.} 
\label{fig:eg_frame} 
\end{figure}

However, since there is no instance-level label to distinguish different objects in most practical datasets, we instead regard each connected component composed of salient objects in the saliency map as an independent proxy $C_k$. Some examples of partitions are presented in \cref{fig:eg_frame}, where an image will be divided into $N_c=K+1$ parts, including $K$ foreground frames and a background frame. Please refer to implementation details in \cref{exp_setup} for more details of the connected component.

The bounding boxes are similar to those widely applied in the area of object detection~\cite{Few-Shot-Object-Detection}~\cite{Object-Detection-in-Aerial-Images}. However, object detection makes bounding box regression to match the predicted boxes as close to the ground-truth boxes as possible, while our approach generates the bounding boxes from the ground-truth binary masks and exploits them as auxiliary tools to calculate the loss and metric results around each salient object.


In this way, the goal of $\SMAE$ becomes
\begin{equation}
\label{wMAE_def}
\begin{aligned}
    \SMAE(f)&= \\
    \frac{1}{K+\alpha} & \left[\sum_{k=1}^{K} \MAE(f_k^{fore}) + \alpha \MAE(f_{K+1}^{back})\right],
\end{aligned}
\end{equation}
where a parameter $\alpha$, determined by the ratio of the size of the background and the sum of all foreground frames, namely  $\alpha=\frac{S_{K+1}^{back}}{\sum_{k=1}^{K}S_k^{fore}}$, is further introduced to balance the model attention adaptively. By doing so, the predictor could not only pay equal consideration to salient objects of various sizes, but also impose an appropriate penalty for misclassifications in the background. This plays an important role in reducing the false positives as illustrated in \cref{Ablation Studies}.  

In the following, we make a brief discussion between $\MAE$ and our proposed $\SMAE$, with proof in \cref{prop1_proof}. 

\begin{mdframed}[hidealllines=true,backgroundcolor=myblue,innerleftmargin=3pt,innerrightmargin=3pt,leftmargin=-3pt,rightmargin=-3pt]
\begin{proposition}[Informal] \label{prop1}
Given two different predictors $f_{A}$ and $f_{B}$, the following two possible cases suggest that $\SMAE$ is more effective than $\MAE$ during evaluation.

{\color{blue}{\textbf{Case 1:}}} 
Assume that there is a single salient object (i.e., $K=1$), with two different results from predictors $f_A$ and $f_B$. In this case, there is no imbalance from different sizes of objects, and therefore $\SMAE$ is equivalent to $\MAE$.

{\color{blue}{\textbf{Case 2:}}} Suppose there are two salient objects ($K=2$) where $f_A$ and $f_B$ detect the same amount of salient pixels in an image $X$. Meanwhile, assume that $f_A$ only predicts $C_2$ perfectly while $f_B$ could somewhat recognize $C_1$ and $C_2$ partially. In this case, \underline{$f_B$ should still be better than $f_A$} since $f_A$ totally fails on $C_1$. Unfortunately, if $S_{1}^{fore}<S_{2}^{fore}$, $\MAE(f_A)=\MAE(f_B)$ holds but we have $\SMAE(f_A) > \SMAE(f_B)$.

\end{proposition}
\end{mdframed}

\textbf{Remark.} \cref{fig:eg} provides a toy example for {\color{blue}{\textbf{Case 2}}}. Discussions above support that $\MAE$ is sensitive to the size of objects concerning multiple object cases, yet $\SMAE$ can serve our expectations better. We also extend our analysis to the case with $K \ge 3$, which consistently suggests the efficacy of $\SMAE$. The detailed discussion is attached to \cref{prop1_proof}.

\subsubsection{Size-Invariant Composite Metrics} \label{SI-F}
Here we instantiate our size-invariant principle with common composite metrics, including $\F$-score and $\AUC$.

As to the composite metric $\F$-score, we define $\SF$ as follows:
\begin{equation}
    \label{wF_def}
    \SF(f)=\frac{1}{K}\sum_{i=1}^{K}\F(f_k^{fore}),
\end{equation}
where $X_1^{fore}, \cdots, X_{K}^{fore}$ denote foreground frames. Similar to $\SMAE$, we give a proposition in \cref{wF_metric} to support that in multiple object cases, $\SF$ can serve our expectations better. 


As to another common composite metric $\AUC$, we similarly define $\SAUC$ as follows:
\begin{equation}
    \label{wAUC_def}
    \SAUC(f)=\frac{1}{K}\sum_{i=1}^{K}\AUC(f_k^{fore}),
\end{equation}
where $X_1^{fore}, \cdots, X_{K}^{fore}$ denote foreground frames. The analysis of $\SAUC$ is deferred to \cref{wAUC_metric} due to space limitations. 

\section{How to Practically Pursue Size-Invariance?} \label{SI-SOD}
In previous sections, we outlined how to achieve size-invariant evaluation for SOD. Now this section explores how to directly optimize these size-invariant metrics to promote practical SOD performance.
\subsection{A Generic Size-Invariant Optimization Goal}
Motivated by the principles of the size-invariant evaluation, our optimization goal is expressed as follows:
\begin{equation}
\label{eq:loss}
    \mathcal{L}_{\SI}(f)=\sum_{k=1}^{K} \ell(f_k^{fore}) + \alpha \ell(f_{K+1}^{back}),
\end{equation}
where $\ell(\cdot)$ could be any popular loss in the SOD community (such as $\BCE$ or $\IOU$). For simplicity, we let $\ell(f):=\ell(f(X), Y)$ and $\ell(f_i):= \ell(f(X_i), Y_i)$. Similar to \cref{wMAE_def}, if $\ell(\cdot)$ is separable, we set $\alpha=\frac{S^{back}_{K+1}}{\sum_{k=1}^{K}S_k^{fore}}$; for composite losses like DiceLoss~\cite{DiceLoss} and IOU Loss~\cite{IOULoss}, we set $\alpha=0$ because the $\TPR$ is always 0 in the background. Specifically in \cref{loss_description}, we describe detailed implementations of Size-Invariant Optimization for different backbones discussed in \cref{Experiments}.

As discussed in \cref{principles of SI_eval}, \cref{eq:loss} ensures that the model treats all objects equally regardless of size, thus improving the detection of smaller objects.
We give the following proposition to illustrate the mechanism of SI-SOD, with proof in \cref{re-attention_proof}. 
\begin{mdframed}[hidealllines=true,backgroundcolor=myblue,innerleftmargin=3pt,innerrightmargin=3pt,leftmargin=-3pt,rightmargin=-3pt]
\begin{proposition}[Mechanism of SI-SOD]
\label{re-attention_prop}
Given a separable loss function $\ell(\cdot)$ and its corresponding size-invariant loss $\mathcal{L}_{\SI}(\cdot)$, then for a certain scenario:
\begin{enumerate}
    \item when $S_{i}<\frac{S}{K +\alpha}$, we have $w_{\mathcal{L}_{\SI}}(x_i)>w_{\ell}(x_i)$,
    \item when $S_{i} < S_{j},$ we have $w_{\mathcal{L}_{\SI}}(x_i) > w_{\mathcal{L}_{\SI}}(x_j)$.
\end{enumerate}
where $w_{\mathcal{L}_{\SI}}(x_i)$ is the \textbf{weight} of pixel-level loss in $X_i$ with $\mathcal{L}_{\SI}$, and $w_{\ell}(x_i)$ is the \textbf{weight} of pixel-level loss in $X_i$ with the original loss $\ell$.
\end{proposition}
\end{mdframed}
\textbf{Remark.} \label{re-attention_remark}
Compared to standard losses such as $\BCE$, SI-SOD adaptively adjusts the weight of pixel-level loss to ensure equal treatment on different objects. 
    
\textit{Item} 1 illustrates that smaller objects, which fall below a certain size threshold, will produce more loss. \textit{Item} 2 describes that SI-SOD increases the weight for pixels in smaller salient objects, finally alleviating size-sensitivity.

\subsection{Generalization Bound}
In this section, we theoretically demonstrate that SI-SOD can generalize to common SOD tasks, despite several challenges.

First, SOD is considered as structured prediction~\cite{Carlo_2020_nips, li_2021_nips}, where couplings between output substructures make it difficult to directly apply Rademacher Complexity-based techniques in theoretical analysis. The standard result to bound the empirical Rademacher complexity~\cite{Probability1991} holds when the prediction functions are real-valued. To overcome this, we adopt the vector contraction inequality \cite{Maurer_vector_2016} to extend it from real-valued analysis to vector-valued ones, and consequently reach a \textbf{sharper} result with Lipschitz properties \cite{vector_Contraction}.

Another challenge lies in the diversity of losses, which hinders exploring the generalization properties within a coordinated framework. Therefore, by studying from the view of separable and composite functions respectively, we obtain Lipschitz properties~\cite{p_Lipschitz} for both categories and ultimately achieve a \textbf{unified} conclusion.

We present our conclusions here, and the proof is deferred to \cref{generalization_bound_proof}.
\begin{mdframed}[hidealllines=true,backgroundcolor=myblue,innerleftmargin=3pt,innerrightmargin=3pt,leftmargin=-3pt,rightmargin=-3pt]
\begin{theorem}[\textbf{Generalization Bound for SI-SOD}]\label{generalization_bound}
Assume $\mathcal{F} \subseteq \{ f:\mathcal{X} \to \mathbb{R}^K \}$, where $K=H \times W$ is the pixel count in an image,  $g^{(i)}$ is the risk over $i$-th sample, and is $L$-Lipschitz with respect to the $l_{\infty}$ norm, (i.e. $\Vert g(x)-g(\tilde{x})\Vert_\infty \le L\cdot \Vert x-y \Vert_{\infty}$). When there are $N$ $i.i.d.$ samples, there exists a constant $C>0$ for any $\epsilon > 0$, the following generalization bound holds with probability at least $1-\delta$:
\begin{equation}
\begin{aligned}
    &~ \sup_{f \in \mathcal{F}}(\mathbb{E}[g(f)]-\hat{\mathbb{E}}[g(f)]) \\
    \le&~  C\cdot \frac{L\sqrt{K}}{N} \cdot \max_i \mathfrak{R}_N(\mathcal{F}|_i)\cdot\log^{\frac{3}{2}+\epsilon}\left(\frac{N}{\max_i \mathfrak{R}_N(\mathcal{F}|_i)}\right) \\
    &+3\sqrt{\frac{\log \frac{2}{\delta}}{2N}},
\end{aligned}
\end{equation}
where again $g(f(X,Y)):=g(f)$, $\mathbb{E}[g(f)]$ and $\hat{\mathbb{E}}[g(f)]$ represent the expected risk and empirical risk. $\mathfrak{R}_N(\mathcal{F};x_{1:N})=\max_{x_{1:N}\in \mathcal{X}} \mathfrak{R}(\mathcal{F};x_{1:N})$ denotes the worst-case Rademacher complexity, and we let $\mathfrak{R}_N(\mathcal{F} | i)$ denote its restriction to output coordinate $i$. Specifically, 
\begin{enumerate}[label=\textbf{\textit{ Case \arabic*:}},leftmargin=4em]
    \item For separable loss functions $\ell(\cdot)$, if it is $\mu$-Lipschitz, we have $L=\mu$.
    \item For composite loss functions, when $\ell(\cdot)$ is DiceLoss \cite{DiceLoss}, we have $L=\frac{4}{\rho}$, where $\rho=\min \frac{S_{l}^{1,i}}{S_{l}^i}$, which represents the minimum proportion of the salient object in the $l$-th frame within the $i$-th sample.
\end{enumerate}
\end{theorem}
\end{mdframed}
\textbf{Remark.}
 We reach a bound of $\mathcal{O}(\frac{\sqrt{K} \log N}{N})$, which indicates reliable generalization with a large training set.
 
 Specifically, for case 2, the original composite loss $\ell(\cdot)$, still taking DiceLoss as an example, will result in a $\rho'=\frac{S^{1, i}}{S^i}$, which represents the proportion of salient pixels in an image. It is obvious that $\rho > \rho'$ because SI-SOD enlarges the proportion by reducing the denominator from the whole image to a bounding box, and finally leads to a \textbf{smaller} $L$ and \textbf{a sharper bound}.

\begin{table*}[t]
  \centering
  \caption{Quantitative comparisons on MSOD and DUTS-TE. The better results are shown with \textbf{bold}, and darker color indicates superior results. Metrics with $\uparrow$ mean higher value represents better performance, while $\downarrow$ mean lower value represents better performance.}
  \scalebox{0.85}{
    \begin{tabular}{c|l|cccccccccc}
    \toprule
    \multicolumn{1}{c|}{Dataset} & Methods & $\MAE \downarrow $ & $\SMAE \downarrow$ & $\AUC \uparrow $ &$\SAUC \uparrow $ & $\F_m^\beta \uparrow$ & $\SF_m^{\beta} \uparrow $ & $\F_{max}^\beta \uparrow$ & $\SF_{max}^\beta \uparrow$ & $\E_m \uparrow$  \\ 
    \midrule
    \multirow{10}[8]{*}{MSOD} & PoolNet & 0.0752 & 0.1196 & \cellcolor[rgb]{ .984,  .867,  .792}0.9375 & \cellcolor[rgb]{ .98,  .831,  .729}0.9563 & 0.6645 & 0.6397 & 0.7755 & \cellcolor[rgb]{ .984,  .875,  .796}0.8402 & 0.7529 \\
          & \textbf{+ \textit{Ours}} & \cellcolor[rgb]{ .988,  .922,  .878}\textbf{0.0635} & \cellcolor[rgb]{ .98,  .882,  .812}\textbf{0.0924} & \cellcolor[rgb]{ .98,  .831,  .733}\textbf{0.9553} & \cellcolor[rgb]{ .973,  .796,  .678}\textbf{0.9721} & \cellcolor[rgb]{ .988,  .91,  .859}\textbf{0.7314} & \cellcolor[rgb]{ .984,  .871,  .796}\textbf{0.7467} & \cellcolor[rgb]{ .988,  .89,  .827}\textbf{0.8200} & \cellcolor[rgb]{ .976,  .8,  .682}\textbf{0.8867} & \cellcolor[rgb]{ .988,  .894,  .831}\textbf{0.8286} \\
\cmidrule{2-11}          & LDF   & \cellcolor[rgb]{ .976,  .839,  .745}0.0508 & \cellcolor[rgb]{ .984,  .89,  .827}0.0946 & 0.8719 & \cellcolor[rgb]{ .988,  .894,  .831}0.9246 & \cellcolor[rgb]{ .984,  .875,  .8}0.7589 & \cellcolor[rgb]{ .996,  .965,  .945}0.6691 & \cellcolor[rgb]{ .988,  .906,  .847}0.8144 & 0.7575 & \cellcolor[rgb]{ .988,  .902,  .843}0.8241 \\
          & \textbf{+ \textit{Ours}} & \cellcolor[rgb]{ .976,  .835,  .741}\textbf{0.0506} & \cellcolor[rgb]{ .98,  .867,  .792}\textbf{0.0893} & \cellcolor[rgb]{ .98,  .835,  .741}\textbf{0.9530} & \cellcolor[rgb]{ .98,  .855,  .769}\textbf{0.9441} & \cellcolor[rgb]{ .98,  .847,  .757}\textbf{0.7796} & \cellcolor[rgb]{ .984,  .859,  .776}\textbf{0.7573} & \cellcolor[rgb]{ .98,  .835,  .741}\textbf{0.8415} & \cellcolor[rgb]{ .973,  .796,  .678}\textbf{0.8879} & \cellcolor[rgb]{ .98,  .831,  .733}\textbf{0.8726} \\
\cmidrule{2-11}          & ICON  & \cellcolor[rgb]{ .98,  .863,  .784}0.0545 & \cellcolor[rgb]{ .984,  .89,  .827}0.0945 & \cellcolor[rgb]{ .996,  .949,  .922}0.8973 & \cellcolor[rgb]{ .996,  .965,  .941}0.8909 & \cellcolor[rgb]{ .984,  .863,  .78}0.7687 & \cellcolor[rgb]{ .992,  .925,  .882}0.7029 & \cellcolor[rgb]{ .988,  .898,  .835}0.8178 & \cellcolor[rgb]{ .996,  .969,  .949}0.7789 & \cellcolor[rgb]{ .984,  .867,  .788}0.8487 \\
          & \textbf{+ \textit{Ours}} & \cellcolor[rgb]{ .98,  .855,  .773}\textbf{0.0535} & \cellcolor[rgb]{ .976,  .839,  .749}\textbf{0.0830} & \cellcolor[rgb]{ .98,  .835,  .741}\textbf{0.9537} & \cellcolor[rgb]{ .98,  .839,  .745}\textbf{0.9514} & \cellcolor[rgb]{ .984,  .863,  .78}\textbf{0.7691} & \cellcolor[rgb]{ .98,  .839,  .745}\textbf{0.7738} & \cellcolor[rgb]{ .98,  .847,  .757}\textbf{0.8373} & \cellcolor[rgb]{ .98,  .831,  .733}\textbf{0.8665} & \cellcolor[rgb]{ .98,  .831,  .733}\textbf{0.8742} \\
\cmidrule{2-11}          & GateNet & \cellcolor[rgb]{ .973,  .796,  .678}\textbf{0.0442} & \cellcolor[rgb]{ .976,  .831,  .733}0.0808 & \cellcolor[rgb]{ .984,  .878,  .804}0.9331 & \cellcolor[rgb]{ .988,  .894,  .835}0.9244 & \cellcolor[rgb]{ .976,  .82,  .714}0.8005 & \cellcolor[rgb]{ .984,  .859,  .776}0.7581 & \cellcolor[rgb]{ .976,  .812,  .706}0.8510 & \cellcolor[rgb]{ .984,  .867,  .788}0.8434 & \cellcolor[rgb]{ .976,  .827,  .725}0.8776 \\
          & \textbf{+ \textit{Ours}} & \cellcolor[rgb]{ .973,  .796,  .678}0.0444 & \cellcolor[rgb]{ .973,  .8,  .682}\textbf{0.0734} & \cellcolor[rgb]{ .98,  .851,  .765}\textbf{0.9456} & \cellcolor[rgb]{ .98,  .855,  .773}\textbf{0.9436} & \cellcolor[rgb]{ .973,  .796,  .678}\textbf{0.8157} & \cellcolor[rgb]{ .973,  .796,  .678}\textbf{0.8083} & \cellcolor[rgb]{ .973,  .796,  .678}\textbf{0.8570} & \cellcolor[rgb]{ .976,  .824,  .718}\textbf{0.8724} & \cellcolor[rgb]{ .973,  .796,  .678}\textbf{0.8972} \\
\cmidrule{2-11}          & EDN   & \cellcolor[rgb]{ .973,  .812,  .702}0.0467 & \cellcolor[rgb]{ .973,  .824,  .722}0.0788 & \cellcolor[rgb]{ .988,  .906,  .847}0.9196 & \cellcolor[rgb]{ .988,  .906,  .851}0.9188 & \cellcolor[rgb]{ .98,  .827,  .729}0.7925 & \cellcolor[rgb]{ .98,  .851,  .765}0.7635 & \cellcolor[rgb]{ .98,  .839,  .745}0.8410 & \cellcolor[rgb]{ .984,  .886,  .82}0.8321 & \cellcolor[rgb]{ .98,  .835,  .737}0.8712 \\
          & \textbf{+ \textit{Ours}} & \cellcolor[rgb]{ .973,  .8,  .686}\textbf{0.0453} & \cellcolor[rgb]{ .973,  .796,  .678}\textbf{0.0724} & \cellcolor[rgb]{ .984,  .863,  .784}\textbf{0.9401} & \cellcolor[rgb]{ .984,  .867,  .788}\textbf{0.9387} & \cellcolor[rgb]{ .976,  .812,  .702}\textbf{0.8057} & \cellcolor[rgb]{ .976,  .808,  .698}\textbf{0.7990} & \cellcolor[rgb]{ .976,  .8,  .686}\textbf{0.8555} & \cellcolor[rgb]{ .98,  .839,  .745}\textbf{0.8619} & \cellcolor[rgb]{ .976,  .804,  .69}\textbf{0.8936} \\
    \midrule
    \multirow{10}[8]{*}{DUTS-TE} & PoolNet & 0.0656 & 0.0609 & \cellcolor[rgb]{ .835,  .894,  .941}0.9607 & \cellcolor[rgb]{ .788,  .867,  .937}0.9716 & \cellcolor[rgb]{ .949,  .949,  .949}0.7200 & \cellcolor[rgb]{ .949,  .949,  .949}0.7569 & \cellcolor[rgb]{ .949,  .949,  .949}0.8245 & \cellcolor[rgb]{ .949,  .949,  .949}0.8715 & \cellcolor[rgb]{ .949,  .949,  .949}0.8103 \\
          & \textbf{+ \textit{Ours}} & \cellcolor[rgb]{ .965,  .976,  .988}\textbf{0.0621} & \cellcolor[rgb]{ .945,  .965,  .984}\textbf{0.0562} & \cellcolor[rgb]{ .792,  .871,  .937}\textbf{0.9706} & \cellcolor[rgb]{ .741,  .843,  .933}\textbf{0.9824} & \cellcolor[rgb]{ .898,  .925,  .945}\textbf{0.7479} & \cellcolor[rgb]{ .839,  .894,  .941}\textbf{0.8172} & \cellcolor[rgb]{ .882,  .918,  .945}\textbf{0.8438} & \cellcolor[rgb]{ .788,  .871,  .937}\textbf{0.9029} & \cellcolor[rgb]{ .871,  .91,  .945}\textbf{0.8478} \\
\cmidrule{2-11}          & LDF   & \cellcolor[rgb]{ .773,  .863,  .941}\textbf{0.0419} & \cellcolor[rgb]{ .776,  .863,  .941}\textbf{0.0410} & \cellcolor[rgb]{ .949,  .949,  .949}0.9337 & \cellcolor[rgb]{ .804,  .875,  .941}0.9680 & \cellcolor[rgb]{ .761,  .855,  .937}\textbf{0.8203} & \cellcolor[rgb]{ .835,  .89,  .941}0.8201 & \cellcolor[rgb]{ .776,  .863,  .937}0.8735 & \cellcolor[rgb]{ .906,  .929,  .949}0.8802 & \cellcolor[rgb]{ .796,  .871,  .941}0.8821 \\
          & \textbf{+ \textit{Ours}} & \cellcolor[rgb]{ .792,  .875,  .945}0.0440 & \cellcolor[rgb]{ .792,  .875,  .945}0.0422 & \cellcolor[rgb]{ .8,  .875,  .941}\textbf{0.9690} & \cellcolor[rgb]{ .773,  .859,  .937}\textbf{0.9756} & \cellcolor[rgb]{ .784,  .867,  .937}0.8076 & \cellcolor[rgb]{ .8,  .875,  .941}\textbf{0.8388} & \cellcolor[rgb]{ .776,  .863,  .937}\textbf{0.8736} & \cellcolor[rgb]{ .745,  .847,  .937}\textbf{0.9117} & \cellcolor[rgb]{ .78,  .863,  .937}\textbf{0.8895} \\
\cmidrule{2-11}          & ICON  & \cellcolor[rgb]{ .812,  .886,  .949}0.0461 & \cellcolor[rgb]{ .827,  .894,  .953}0.0454 & \cellcolor[rgb]{ .894,  .922,  .945}0.9469 & \cellcolor[rgb]{ .914,  .933,  .949}0.9424 & \cellcolor[rgb]{ .773,  .859,  .937}\textbf{0.8131} & \cellcolor[rgb]{ .82,  .886,  .941}0.8270 & \cellcolor[rgb]{ .808,  .878,  .941}\textbf{0.8648} & \cellcolor[rgb]{ .898,  .925,  .949}0.8815 & \cellcolor[rgb]{ .788,  .867,  .937}0.8858 \\
          & \textbf{+ \textit{Ours}} & \cellcolor[rgb]{ .808,  .882,  .949}\textbf{0.0454} & \cellcolor[rgb]{ .804,  .882,  .949}\textbf{0.0435} & \cellcolor[rgb]{ .824,  .886,  .941}\textbf{0.9640} & \cellcolor[rgb]{ .792,  .871,  .937}\textbf{0.9706} & \cellcolor[rgb]{ .792,  .871,  .937}0.8031 & \cellcolor[rgb]{ .796,  .875,  .941}\textbf{0.8395} & \cellcolor[rgb]{ .816,  .882,  .941}0.8629 & \cellcolor[rgb]{ .827,  .886,  .941}\textbf{0.8958} & \cellcolor[rgb]{ .776,  .863,  .937}\textbf{0.8921} \\
\cmidrule{2-11}          & GateNet & \cellcolor[rgb]{ .741,  .843,  .933}\textbf{0.0383} & \cellcolor[rgb]{ .745,  .843,  .933}0.0380 & \cellcolor[rgb]{ .827,  .886,  .941}0.9629 & \cellcolor[rgb]{ .831,  .89,  .941}0.9619 & \cellcolor[rgb]{ .741,  .843,  .933}\textbf{0.8292} & \cellcolor[rgb]{ .773,  .863,  .937}0.8519 & \cellcolor[rgb]{ .741,  .843,  .933}\textbf{0.8835} & \cellcolor[rgb]{ .784,  .867,  .937}0.9041 & \cellcolor[rgb]{ .749,  .847,  .937}\textbf{0.9053} \\
          & \textbf{+ \textit{Ours}} & \cellcolor[rgb]{ .753,  .851,  .933}0.0399 & \cellcolor[rgb]{ .741,  .843,  .933}\textbf{0.0375} & \cellcolor[rgb]{ .812,  .878,  .941}\textbf{0.9663} & \cellcolor[rgb]{ .8,  .875,  .941}\textbf{0.9692} & \cellcolor[rgb]{ .765,  .855,  .937}0.8185 & \cellcolor[rgb]{ .741,  .843,  .933}\textbf{0.8687} & \cellcolor[rgb]{ .776,  .863,  .937}0.8743 & \cellcolor[rgb]{ .745,  .847,  .937}\textbf{0.9116} & \cellcolor[rgb]{ .749,  .847,  .937}0.9038 \\
\cmidrule{2-11}          & EDN   & \cellcolor[rgb]{ .745,  .843,  .933}\textbf{0.0389} & \cellcolor[rgb]{ .753,  .851,  .933}0.0388 & \cellcolor[rgb]{ .839,  .894,  .941}0.9600 & \cellcolor[rgb]{ .835,  .89,  .941}0.9611 & \cellcolor[rgb]{ .745,  .847,  .937}\textbf{0.8288} & \cellcolor[rgb]{ .765,  .855,  .937}0.8565 & \cellcolor[rgb]{ .773,  .859,  .937}0.8752 & \cellcolor[rgb]{ .796,  .871,  .941}0.9017 & \cellcolor[rgb]{ .753,  .851,  .937}0.9033 \\
          & \textbf{+ \textit{Ours}} & \cellcolor[rgb]{ .749,  .847,  .933}0.0392 & \cellcolor[rgb]{ .745,  .847,  .933}\textbf{0.0381} & \cellcolor[rgb]{ .816,  .882,  .941}\textbf{0.9658} & \cellcolor[rgb]{ .8,  .875,  .941}\textbf{0.9687} & \cellcolor[rgb]{ .749,  .847,  .937}0.8260 & \cellcolor[rgb]{ .745,  .847,  .937}\textbf{0.8672} & \cellcolor[rgb]{ .769,  .859,  .937}\textbf{0.8765} & \cellcolor[rgb]{ .741,  .843,  .933}\textbf{0.9119} & \cellcolor[rgb]{ .741,  .843,  .933}\textbf{0.9072} \\
    \bottomrule
    \end{tabular}%
    }
  \label{tab:exp_result}
\end{table*}%

\section{Experiments} \label{Experiments}
In this section, we describe some details of the experiments and present our results.
\textbf{Due to space limitations, please refer to \cref{Experiments_appendix} for an extended version.}

\subsection{Experimental Setups} \label{exp_setup}

\textbf{Datasets.} Eight datasets, DUTS~\cite{DUTS}, ECSSD~\cite{ECSSD}, DUT-OMRON~\cite{DUT-OMRON}, HKU-IS~\cite{HKU-IS}, MSOD~\cite{MSOD}, PASCAL-S~\cite{ECSSD}, SOD~\cite{SOD} and XPIE~\cite{XPIE}, are included in the experiment. Following common practice, we train our network on the DUTS training set (DUTS-TR) and test it on the DUTS test set (DUTS-TE) and the other seven datasets. Detailed introductions on these datasets are deferred to \cref{dataset_appendix}.

\textbf{Competitors.} 
To demonstrate the effectiveness of size-invariant loss, we integrate it into five state-of-the-art backbones: EDN \cite{EDN}, ICON \cite{ICON}, GateNet \cite{GateNet}, LDF \cite{LDF}, PoolNet \cite{PoolNet}. EDN, ICON, and LDF utilize DiceLoss or IOULoss to handle the potential imbalanced distribution, and ICON specifically focuses on the macro-integrity, which are summarized at \cref{competitor_appendix} with details.
Specifically, \ul{we modify the original loss functions into their corresponding size-invariant versions} following \cref{eq:loss}, and re-train the network with the same setting. Correspondingly, we also compare the time cost of our method with original optimization frameworks, which is deferred at \cref{time_cost}.

\textbf{Evaluation Metrics.} 
Apart from our proposed metrics $\SMAE$, $\SF$ and $\SAUC$, we also include common metrics such as Mean Absolute Error($\MAE$), max F-measure($\F_{max}^\beta$), mean F-measure($\F_m^\beta$) and $\AUC$.


Another widely used metric is $\E_m$ introduced by \cite{Emeasure}, which is a newly proposed metric considering both global and local information. Definitions and calculations of all metrics are deferred to \cref{protocol_appendix}.

\textbf{Implementation Details.}
We carried out the experiments on a single GeForce RTX 3090. To ensure fairness,  both the original and modified backbones are trained under identical settings. All images are resized into $384 \times 384$ for training and testing, and the ResNet50~\cite{resnet} pre-trained on ImageNet~\cite{ImageNet} is loaded. Specific settings and optimization details for each backbone are deferred at \cref{details_appendix} and \cref{loss_description}.

We preprocess the dataset with the package \textit{skimage}, which identifies connected components with the ground-truth mask. Then we obtain the minimum bounding box following \cref{eq:bounding_box}. Note that all procedures can also be done during training without any preprocessing. 

\subsection{Overall Performance}
As mentioned above, we re-train the backbones with our size-invariant loss for a fair comparison. \cref{tab:exp_result} shows the results on MSOD and DUTS-TE. The result is shown in a pair of backbones before and after applying our size-invariant loss, with the superior result highlighted in \textbf{bold}.

\begin{figure}[t]
\centering
\subfigure[MSOD]{   
\begin{minipage}{0.47\linewidth}
\includegraphics[width=\linewidth]{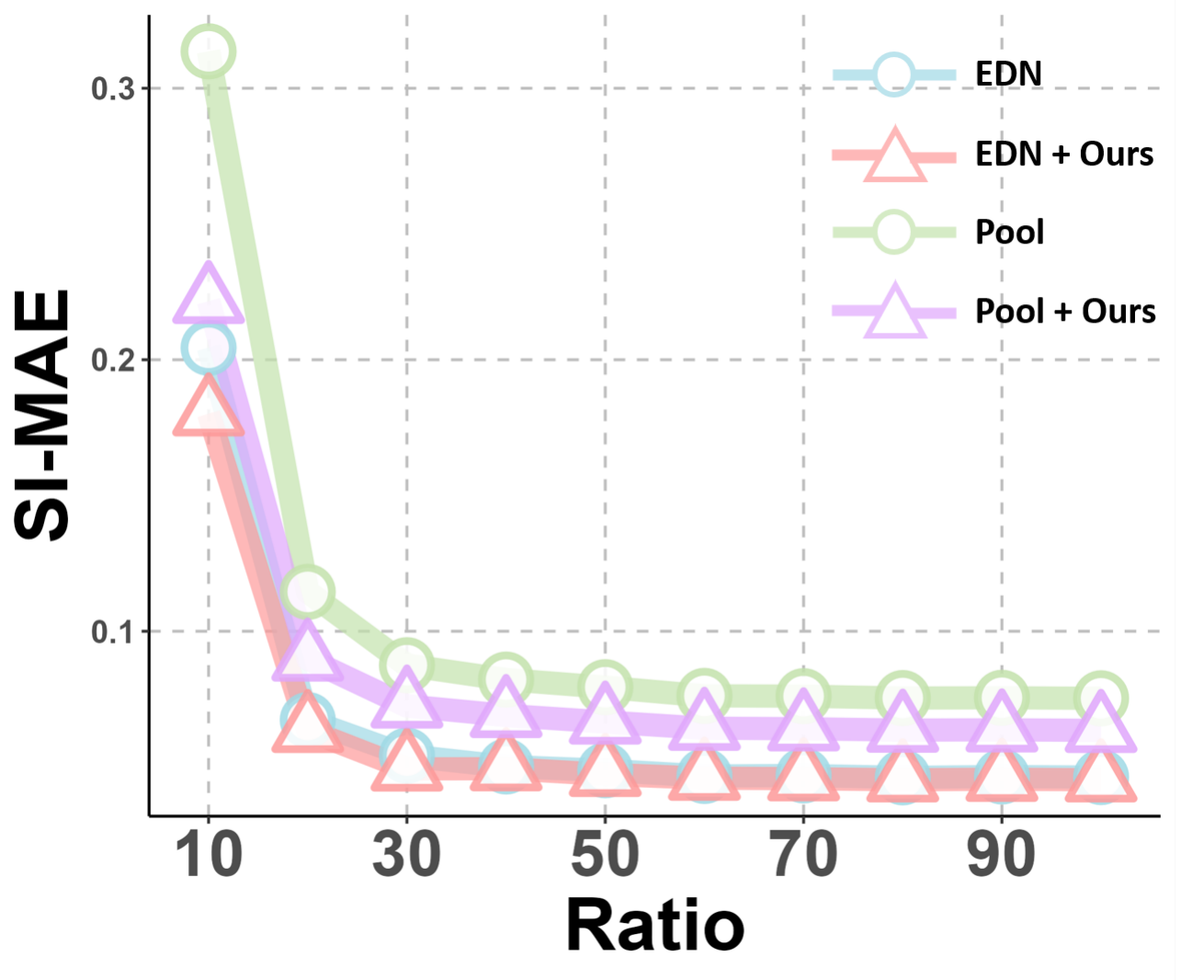}  
\label{fig:EDN_Pool_msod_ratio_line}
\end{minipage}
}
\subfigure[DUTS]{   
\begin{minipage}{0.47\linewidth}
\includegraphics[width=\linewidth]{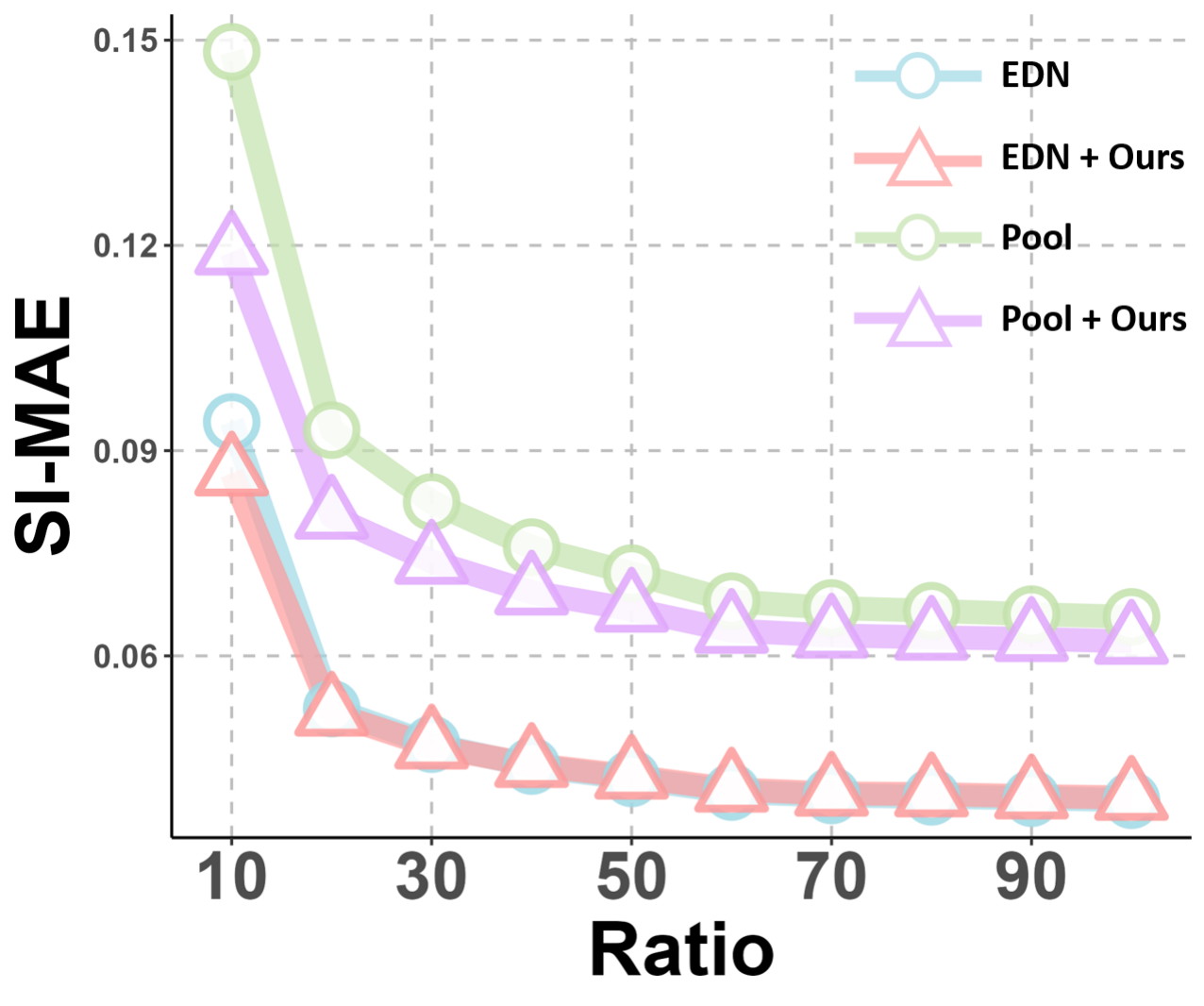}  
\label{fig:EDN_Pool_DUTS_ratio_line}
\end{minipage}
}
\vspace{-0.3cm}
\caption{$\SMAE$ performance on objects with different sizes on two representative datasets, with EDN and PoolNet as backbones.}
\label{fig:fine-analysis-EDN-Pool-ratio} 
\vspace{-0.3cm}
\end{figure}

Since samples in MSOD contain multiple salient objects, it naturally arises that small objects can be overlooked due to the imbalance. Therefore, all backbones with our loss achieve considerable improvements on nearly all metrics, even including the original $\MAE$. Averagely, our method outperforms other frameworks by around 0.012, 0.038, 0.070, 0.065, 0.038 on $\SMAE$, $\SAUC$, $\SF_m^\beta$, $\SF_{max}^\beta$ and $\E_m$, respectively.

\cref{tab:exp_result} also shows the performance on DUTS-TE. Our method achieves better results on almost all size-invariant metrics and $\E_m$, and stays competitive in terms of original $\MAE$ and $\F$-score. This justifies that the size-invariant loss achieves similar performance on single-object scenarios, suggesting the superior generalization ability of our loss. Averagely, our method outperforms other frameworks by around 0.001, 0.012, 0.024, 0.020, 0.011 on $\SMAE$, $\SAUC$, $\SF_m$, $\SF_{max}^\beta$, $\E_m$ on DUTS-TE. Results on other datasets are deferred to \cref{tab:exp_result_appendix}.

\begin{figure*}[t]
    \centering
    \includegraphics[width=\linewidth]{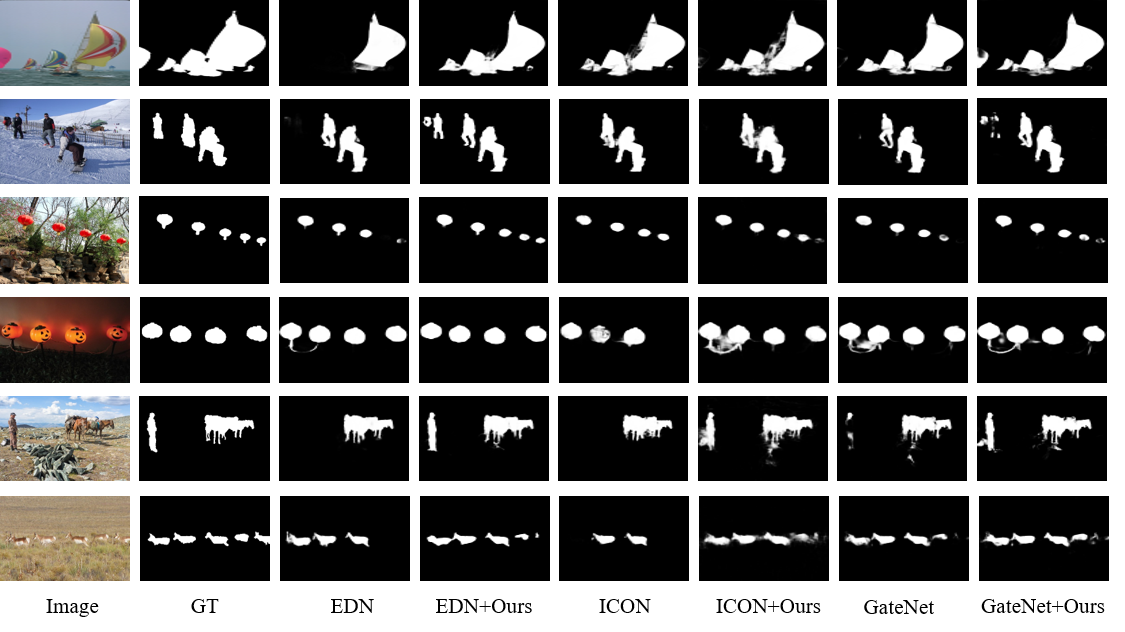}
    \vspace{-0.3cm}
    \caption{Qualitative comparison on different backbones.}
    \label{fig:vis}
    \vspace{-0.3cm}
\end{figure*}

\cref{fig:vis} shows the qualitative comparison on different backbones. While the original backbones may fail to detect all the salient objects in some hard samples, our method can significantly improve the detection on multi-object occasions. For example, in the 1st image, EDN only finds the largest sailboat on the right but fails to detect two smaller targets on the left, while ours additionally detects two small sailboats. In the 4th image, EDN detects fewer false positive pixels and ICON detects one more salient object with our loss. In the 3rd, 5th, and 6th images, all backbones detect more salient objects at the right part after using our loss. More qualitative comparisons are deferred to \cref{Qualitative_appendix}.

\subsection{Fine-grained Analysis}
\subsubsection{Performance with Respect to Sizes}
We conduct size-relevant analysis on five datasets. As it is the size of salient objects that our method focuses on, we divide all salient objects into ten groups according to their proportion to the entire image, ranging from [0\%, 10\%], [10\%,20\%], and finally up to [90\%, 100\%]. We evaluate the performance within each group, and here we only take foreground frames into account to concentrate on the detection performance of salient objects with different sizes.

\begin{figure}[t]
    \centering
    \subfigure[MSOD]{   
    \begin{minipage}{0.47\linewidth}
    \includegraphics[width=\linewidth]{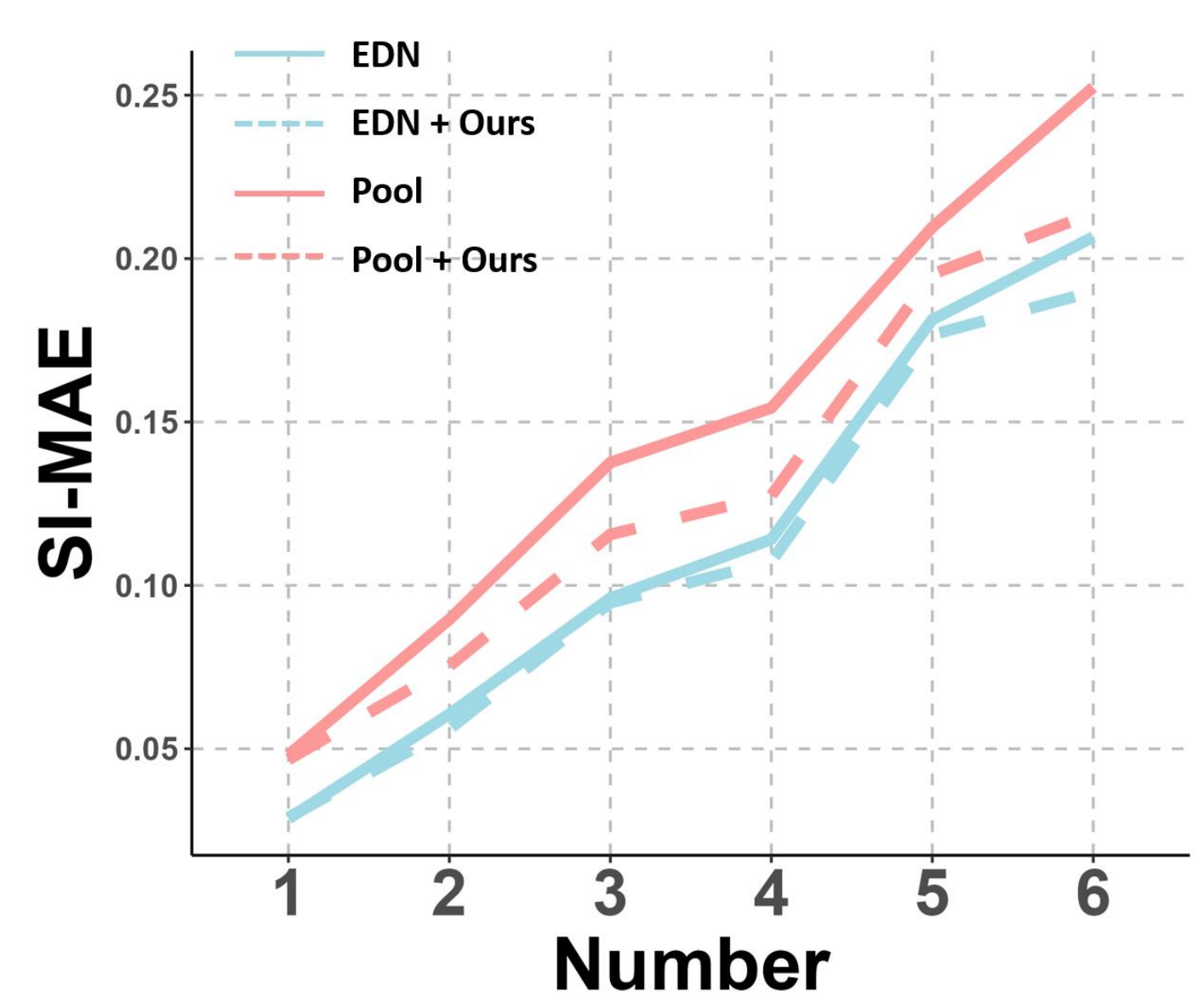}  
    \label{fig:EDN_Pool_msod_num_line}
    \end{minipage}
    }
    \subfigure[DUTS]{   
    \begin{minipage}{0.47\linewidth}
   \includegraphics[width=\linewidth]{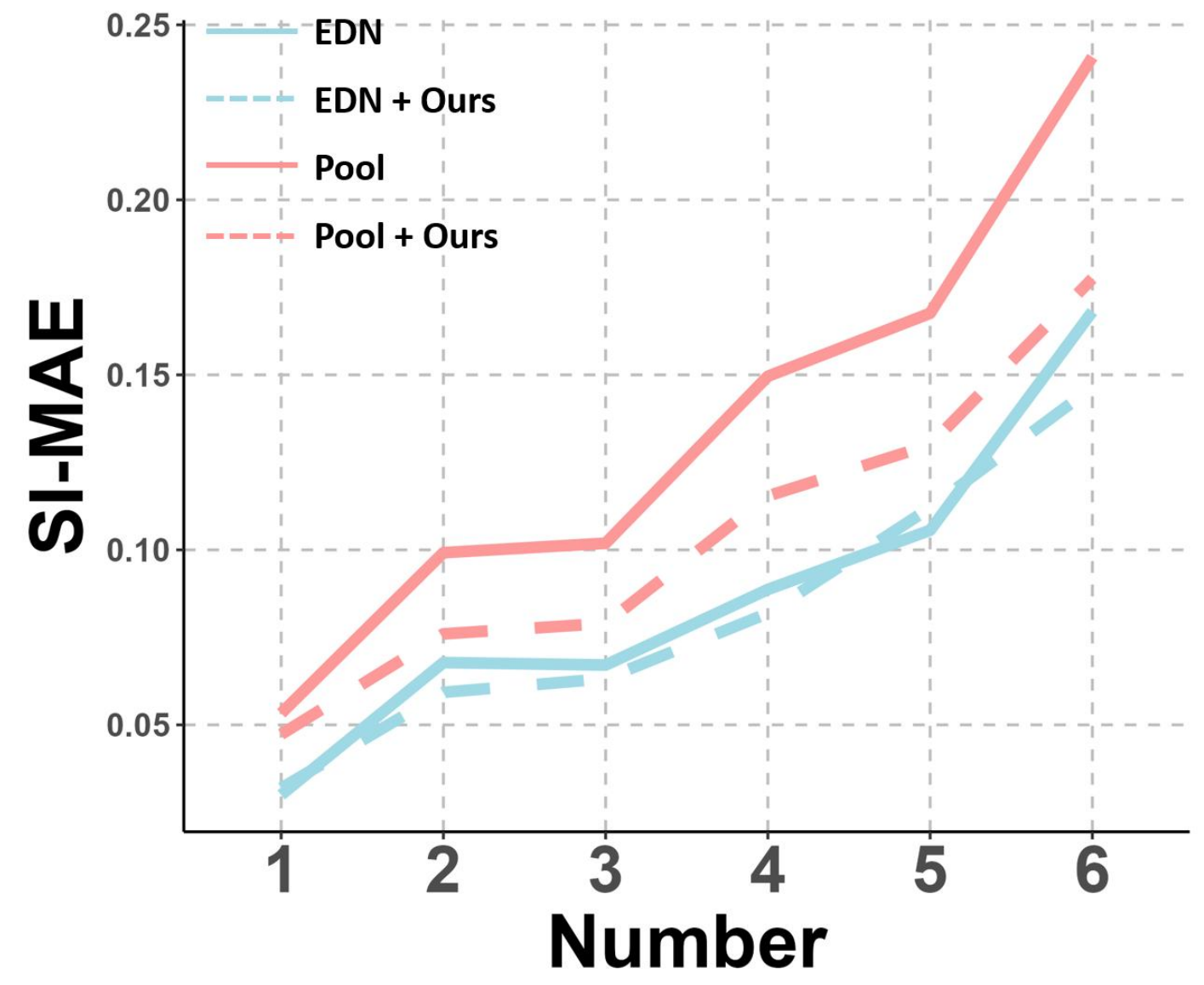}
    \label{fig:EDN_Pool_DUTS_num_line}
    \end{minipage}
    }
    \vspace{-0.3cm}
    \caption{$\SMAE$ performance with different object numbers on two representative datasets, with EDN and PoolNet as backbones.}
    \label{fig:fine-analysis-EDN-Pool-num}
    \vspace{-0.5cm}
\end{figure}

From \cref{fig:fine-analysis-EDN-Pool-ratio}, we observe that all backbones perform well on larger objects but show remarkable improvements on smaller objects when using our method. This aligns with our objective to enhance the detection of smaller objects. Specifically, for objects with size in [0\%, 10\%] of the image, our method outperforms the previous backbone, say EDN, by around 0.024 on $\SMAE$ on the MSOD dataset. As seen in \cref{fig:msod_area}, small-size salient objects usually account for the majority, therefore such improvement firmly speaks to our progress. This is not reflected by size-sensitive metrics like $\MAE$, but can be directly revealed by our proposed $\SMAE$.
Performance analysis with respect to the object size on other backbones and datasets is deferred to \cref{size-fine-grained_appendix}.

\subsubsection{Performance with Respect to Object Numbers}
We also conduct number-relevant analysis on five datasets to evaluate the performance on single-object and multi-object scenarios, as shown in \cref{fig:fine-analysis-EDN-Pool-num}. With the number of salient objects increasing, the SOD tasks are getting imbalanced, where some objects are more likely to be ignored. Therefore, we divide all samples into several groups according to the number of salient objects in the image.

Generally, our method shows substantial improvements in multi-object scenarios and remains competitive in single-object cases, which again justifies the generalization and universality of our method. Specifically, for samples with greater than or equal to two salient objects, EDN gains an improvement by around 0.007 on $\SMAE$ on the MSOD dataset after employing our size-invariant loss.
Performance analysis with respect to the object numbers on other backbones and datasets is deferred to \cref{number-fine-grained_appendix}.

\begin{figure}[t]
\centering
\subfigure[MSOD]{   
\begin{minipage}{0.44\linewidth}
\includegraphics[width=\linewidth]{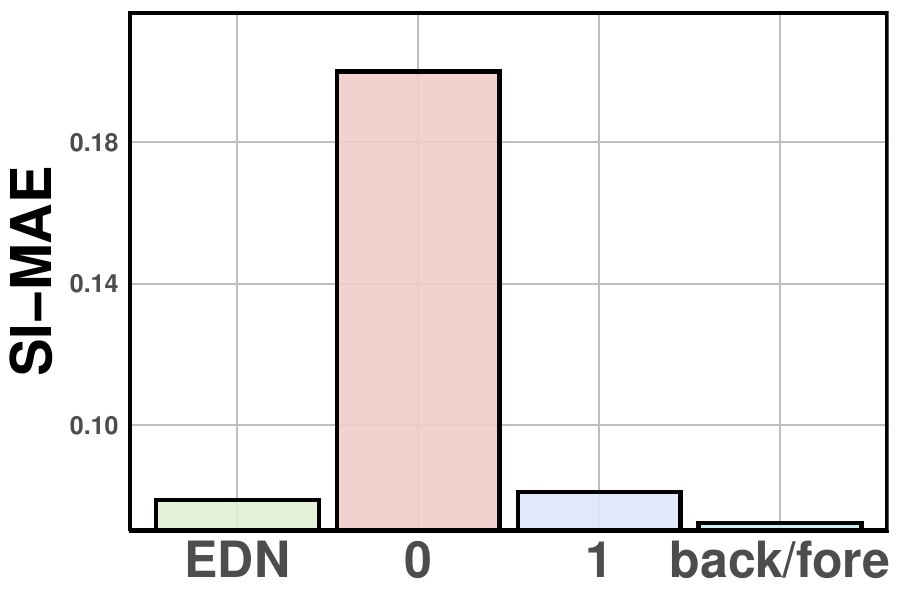}  
\label{fig:EDN_msod_ablation}
\end{minipage}
}
\subfigure[DUTS]{   
\begin{minipage}{0.44\linewidth}
\includegraphics[width=\linewidth]{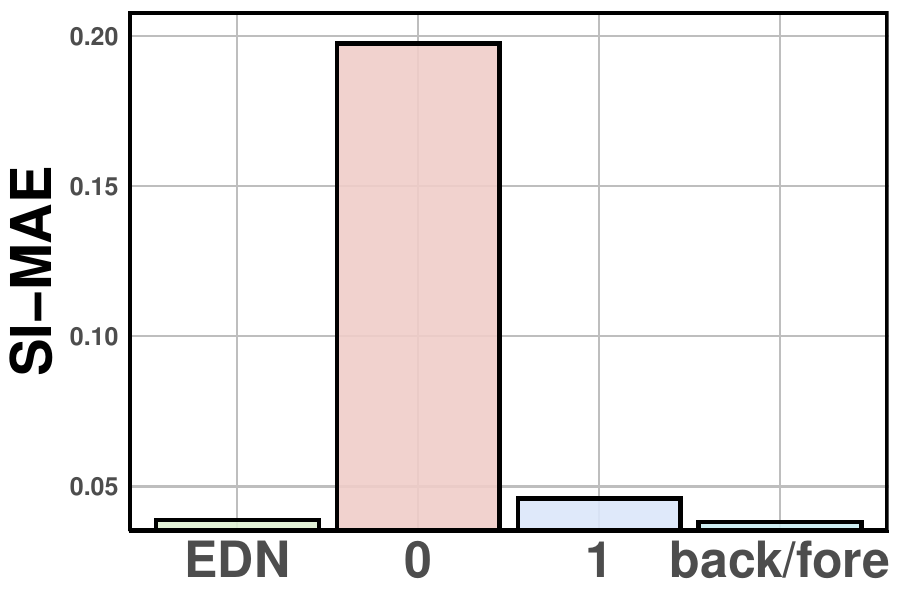}  
\label{fig:EDN_DUTS_ablation}
\end{minipage}
}
\caption{$\SMAE$ performance on two representative datasets with different value of $\alpha$. EDN, 0, 1, and back/fore represent the original backbone EDN, $\alpha=0$, $\alpha=1$ and $\alpha=\frac{S^{back}}{S^{fore}}$, respectively.}
\label{fig:ablation_SIMAE}
\vspace{-0.5cm}
\end{figure}

\subsubsection{Ablation Studies} \label{Ablation Studies}
To investigate how the parameter $\alpha$ works, we conduct ablation studies on $\alpha$ to verify its effectiveness. Here we set $\alpha$ among $0, 1, \frac{S^{back}}{S^{fore}}$. $\alpha=0$ indicates that we do not consider the background frame and pay all attention to foreground frames, while $\alpha=1$ means that we consider the background frame equally as other foreground frames, and $\alpha=\frac{S^{back}}{S^{fore}}$ is exactly our method.
\cref{fig:ablation_SIMAE} illustrates the ablations on dataset MSOD and DUTS with the backbone EDN. 
$\alpha=0$ induces an extreme result with a high score within foreground frames and a low score in the background frame because it solely focuses on foreground detection. $\alpha=1$ alleviates the phenomenon, and surpasses the original framework on some metrics, but still predicts too many false positives, due to the slight penalty on the error within the background frame. Experiments on other datasets also speak to the efficacy of the $\alpha$. More detailed results are deferred to \cref{ablation_appendix}.

\section{Conclusion}
In this paper, we explore the size-invariance in SOD tasks. When multiple objects of various sizes co-exist, we observe that current evaluation metrics are size-sensitive, where larger objects are focused and smaller objects are likely overlooked. To rectify this, we introduce a generic approach to achieve size-invariance. Specifically, we propose $\SMAE$ and $\SF$, which evaluate each salient object separately before merging their results. We further design an optimization framework directly toward this goal, which can adaptively balance the weights to ensure equal treatment on different objects. Theoretically, we provide evidence to support our proposed metrics and present the generalization analysis for our SI-SOD optimization loss. Comprehensive experiments consistently demonstrate the efficacy of our method.

\section*{Acknowledgements}
This work was supported in part by the National Key R\&D Program of China under Grant 2018AAA0102000, in part by National Natural Science Foundation of China: 62236008, U21B2038, U23B2051, U2001202, 61931008, 62122075, 61976202, 62206264 and 92370102, in part by Youth Innovation Promotion Association CAS, in part by the Strategic Priority Research Program of the Chinese Academy of Sciences, Grant No. XDB0680000, in part by the Innovation Funding of ICT, CAS under Grant No.E000000, in part by the Taishan Scholar Project of Shandong Province under Grant tsqn202306079.

\section*{Impact Statement}
We propose a general SOD method to deal with the potential bias toward small objects. For fairness-sensitive scenarios, it might be helpful to improve fairness for minority groups.

\bibliography{main}
\bibliographystyle{icml2024}


\newpage
\appendix
\onecolumn
\definecolor{app_blue}{RGB}{0,20,115}
\textcolor{white}{dasdsa}
\section*{\textcolor{app_blue}{\Large{Contents}}}

\titlecontents{section}[0em]{\color{app_blue}\bfseries}{\thecontentslabel. }{}{\hfill\contentspage}

\titlecontents{subsection}[1.5em]{\color{app_blue}}{\thecontentslabel. }{}{\titlerule*[0.75em]{.}\contentspage} 

\startcontents[sections]

\printcontents[sections]{l}{1}{\setcounter{tocdepth}{3}}
\newpage

\section{Evaluation Metrics for SOD} \label{protocol_appendix}
Different from usual classification tasks where we calculate accuracy on the image level, SOD requires evaluation pixel by pixel. Other pixel-level tasks like semantic segmentation adopt $\mathsf{mIOU}$, which utilizes the mean $\IOU$ over all classes as the metric. However, in SOD all salient objects are labeled with 1 without further class labels. Therefore, there is no representative metric, and the following are commonly utilized:

\textbf{$\MAE$~\cite{MAE}.} It measures the average absolute error pixel-wise. The prediction is normalized to $[0,1]$ when calculating the errors from the ground truth. It is defined as:
\begin{equation}
    \MAE = \frac{1}{W \times H} \sum_{i=1}^W \sum_{j=1}^H |f_N(X)_{(i,j)}-Y_{(i,j)}|,
\end{equation}
where $f_N(X)$ and $Y$ are the normalized prediction map and the saliency map, respectively.

\textbf{$\F$-score~\cite{F_measure}.} It is designed to deal with imbalanced distribution and comprehensively considers both precision and recall. The original $\F$-score is defined as follows:
\begin{equation}
    \F = \frac{2 \times Precision \times Recall}{Precision + Recall},
\end{equation}
where 
\begin{equation}
    Precision = \frac{\TP}{\TP + \FP}, \; Recall = \frac{\TP}{\TP+\FN},
\end{equation}
where $\TP,\TN,\FP,\FN$ are \textbf{T}rue \textbf{P}ositive, \textbf{T}rue \textbf{N}egative, \textbf{F}alse \textbf{P}ositive and \textbf{F}alse \textbf{N}egative. A set of thresholds is applied to generate the binary result when calculating the metrics above.

According to empirical settings \cite{EDN}, \cite{PiCANet_2018}, \cite{Amulet_2017}, \cite{PoolNet}, we adopt $\F^\beta$ as previous works do~\cite{F-beta}:
\begin{equation}
    \F_{\beta}=\frac{(1+\beta^2) \times Precision \times Recall}{\beta^2 \times Precision + Recall},
\end{equation}
and set $\beta^2=0.3$ to emphasize the importance of precision, following \cite{EDN}, \cite{PiCANet_2018}, \cite{Amulet_2017}, \cite{PoolNet}, etc.

\textbf{$\AUC$.~\cite{AUC_eval}} As SOD is essentially a binary classification task, it is natural that $\AUC$ is suitable for this problem. $\AUC$ considers both $\TPR$ and $\FPR$, and is insensitive to data distribution~\cite{pAUC}.
Geometrically, it can be calculated as follows:
\begin{equation}
    \AUC = \int_{x=0}^1 \TPR(\FPR^{-1}(x))dx.
\end{equation}
It is equivalent to the Wilcoxon test of ranks~\cite{AUC_rank}, and an unbiased estimator of $\AUC$ can be expressed as:
\begin{equation}
    \AUC = \mathbb{E}_{P_+, P_-} [\ell_{0,1}(f(x^+)-f(x^-))],
\end{equation}
where $\ell_{0,1}(\cdot)$ denotes the 0-1 loss.

\textbf{$\E_m$.~\cite{Emeasure}} It considers the match of global and local similarities simultaneously. It is specially designed for binary map evaluation and has been widely used in recent years.
$\E_m$ is defined as follows:
\begin{equation}
    \E_m=\frac{1}{w \times h} \sum_{x=1}^w \sum_{y=1}^h \phi_{FM}(x,y),
\end{equation}
where $w$ and $h$ are the height and width of the image, and 
\begin{equation}
    \phi_{FM} = f(\xi_{FM}),
\end{equation}
where $f(\cdot)$ is a convex function. Here we set $f(x)=\frac{1}{2}(1+x)^2$ as \cite{Emeasure} suggested. $\xi$ is computed as:
\begin{equation}
    \xi_{FM}=\frac{2\phi_{GT} \circ \phi_{GT}}{\phi_{GT} \circ \phi_{GT} + \phi_{FM} \circ \phi_{FM}},
\end{equation}
where $\circ$ represents Hadamard production, and $\phi_I=I-\mu_I \cdot \mathbb{A}$, with $I$ as the input, $\mu_I$ as the global mean value, and $\mathbb{A}$ an all-ones matrix.

Toward the objectives above, most SOD methods are trained with two types of loss functions: pixel-level loss and region-level loss. The former focuses on pixel-level accuracy, and the latter aims at promoting regional performance. 

\textbf{Pixel-level} loss includes binary cross-entropy ($\BCE$), mean square error ($\MSE$), etc. $\BCE$ is the most widely used loss function in SOD because it is essentially a binary classification task for each pixel. It is also reasonable to regard it as a regression task with $\MSE$ considering that there are few pixels labeled between 0 and 1. Specifically, GateNet~\cite{GateNet} and RDSN~\cite{RDSN} employ $\MSE$ as the loss function, while most of the other methods utilize $\BCE$.
Specifically, they are defined as follows:
\begin{equation}
\begin{aligned}
    \BCE &= \frac{1}{N} \sum_{i=1}^N[-{(p_i\log(g_i) + (1 - p_i)\log(1 - g_i))}], \\
    \MSE &= \frac{1}{N} \sum_{i=1}^N (p_i - g_i)^2,
\end{aligned} 
\end{equation}
where $p_i$ and $g_i$ is the prediction and ground-truth for $i$-th pixel.

\textbf{Region-level} loss can vary throughout different methods. Some widely used loss functions include DiceLoss~\cite{DiceLoss}, and IOULoss~\cite{IOULoss}. Both these losses consider the performance in a region, instead of focusing on certain pixels, which can therefore improve the performance from a higher level. DiceLoss is defined as follows:
\begin{equation}
    \mathsf{DiceLoss}=1-\frac{2\cdot \sum_{i}^N p_i \cdot g_i}{\sum_{i}^N p_i^2+\sum_{i}^N g_i^2},
\end{equation}
where the sums run over the $N$ pixels, and $p_i, g_i$ represent the prediction and ground truth, respectively. 

IOULoss is computed as follows, which is slightly different from \cite{IOULoss}:
\begin{equation}
    \mathsf{IOULoss}=1-\frac{\sum_i^N (p_i \cdot g_i)}{\sum_i^N (p_i+d_i)-\sum_i^N (p_i \cdot g_i)},
\end{equation}
where $p_i$ and $g_i$ represent the prediction and ground truth.

There are also other region-level losses, such as SSIM~\cite{SSIM}. Generally, they focus on regional detection performance and are therefore robust against imbalanced distribution.

\section{$\SF$ Metric} \label{wF_metric}
\addcontentsline{sections}{section}{\numberline{}$\SF$ Metric}

Similar to $\SMAE$, $\SF$ is expressed as follows according to \cref{eq:separable-insensitive}:
\begin{equation}
\begin{aligned}
    & \SF(f)\\
    =\;&\frac{1}{N_c} \sum_{i=1}^{N_c}  {\color{orange}{1}} \cdot \frac{2 \cdot \TPR(f_i)}{2\TPR(f_i)+\FPR(f_i)+\FNR(f_i)} \\
    = \;& \frac{1}{N_c} \sum_{i=1}^{N_c} \F(f_i),
\end{aligned}
\end{equation}
where $\color{blue}{\mathbb{P}_{X_i}}$ is replaced by {\color{orange}{1}}. The definition of object frames is the same as that in $\SMAE$. It is worth noting that we do not consider the background frame here and just leave it to $\SMAE$ because there are no salient pixels and the true positive rate is always 0.

Based on the discussions above, we now define the new metric $\SF$ as follows:
\begin{equation}
    \SF(f)=\frac{1}{K}\sum_{i=1}^{K}\F(f_k^{fore}),
\end{equation}
where $X_1^{fore}, \cdots, X_{k}^{fore}$ denote foreground frames.

Also, we give the following proposition to demonstrate the effectiveness of $\SF$ when there are multiple salient objects, with proof deferred to \cref{prop3_proof}:
\begin{mdframed}[hidealllines=true,backgroundcolor=myblue,innerleftmargin=3pt,innerrightmargin=3pt,leftmargin=-3pt,rightmargin=-3pt]
\begin{proposition}[Informal] \label{prop3} Given two different predictors $f_A$ and $f_B$, the following case suggest that $\SF$ is more effective than $\F$ during evaluation.

Suppose there are two salient objects ($K=2$) where $f_A$ and $f_B$ detect the same amount of salient pixels for an image $X$. Meanwhile, assume that $f_A$ only predicts $C_2$ perfectly while $f_B$ could somewhat recognize $C_1$ and $C_2$ partially. In this case, \underline{$f_B$ should still be better than $f_A$} since the latter totally fails on $C_1$. Unfortunately, $\F(f_A)=\F(f_B)$ holds but we have $\SF(f_A) < \SF(f_B)$.

\end{proposition} 
\end{mdframed}

\textbf{Remark.}
\cref{fig:eg} provides a toy example. Discussions above support that $\F$-score is sensitive to the sizes of objects in multi-object cases, yet $\SF$ can serve our expectations better.

\section{$\SAUC$ Metric} \label{wAUC_metric}
Normally, AUC is defined as 
\begin{equation}
    \mathsf{AUC}=\int_0^1 \mathsf{TPR}_f(\mathsf{FPR^{-1}}_f(t))dt,
\end{equation} 
where $f$ is the predictor, $t$ is the probability threshold, $\mathsf{TPR},\mathsf{FPR}$ are the true positive rate and false positive rate, respectively. In the equation above, the integral makes it hard to analyze. Therefore, to further investigate the potential issue for $\AUC$, we adopt another form:
\begin{equation}
    \mathsf{AUC}(f) = \sum_{i=1}^{n^+} \sum_{j=1}^{n^-} \frac{\mathbb{I}(f(X_i^+)>f(X_j^-))}{n^+ n^-},
\end{equation}
where $\mathbb{I}(x)=1$ when $x$ is True, $X_i^+,X_i^-$ are sampled from salient and non-salient pixels.

Following Eq.(5) in the main text, we let
\begin{equation}
 g(f_k) = \sum_{i=1}^{n_k^+} \sum_{j=1}^{n_k^-} \frac{\mathbb{I}(f(X_{k,i}^+)>f(X_{k,j}^-))}{n_k^+ n_k^-},
\end{equation}
where $n_k^+, n_k^-$ are the number of salient and non-salient pixels within the $k$-th part, $X_{k,i}^+$, $X_{k,j}^-$ are respectively sampled from salient and non-salient pixels within the $k$-th part, and $g(f_k)$ is actually the $\AUC$ value within the $k$-th part. Then following the definition of composite functions, we have:
\begin{equation}
    \mathsf{AUC}(f) = \sum_{k=1}^{K}g(f_k) \frac{n_k^+ n_k^-}{n^+n^-} =\sum_{k=1}^{K}g(f_k) \frac{S_k'}{S'},
\end{equation}
where $K$ is the number of foreground frames, $S'=n^+n^-$ and $S_k'=n^+_kn^-_k$, which is also a size-related term. However, it also depends on the distribution of salient pixels, and is therefore more size-robust than $\MAE,\F$-score to some extent. To thoroughly eliminate the influence of size, we can define $\SAUC$, similar to $\SF$:
\begin{equation}
    \mathsf{SI\text{-}AUC}(f)= \frac{1}{K} \sum_{k=1}^{K}\textsf{AUC}(f_k^{fore}).
\end{equation}

\section{Proof for Propositions of Size-Invariant Metrics}
\subsection{Proof and extension for \cref{prop1}}\label{prop1_proof}

\textbf{Restate of \cref{prop1}.} 
Given two different predictors $f_{A}$ and $f_{B}$, the following two possible cases suggest that $\SMAE$ is more effective than $\MAE$ during evaluation.

{\color{blue}{\textbf{Case 1:}}} 
Assume that there is a single salient object (i.e., $K=1$), with two different results from predictors $f_A$ and $f_B$. In this case, there is no imbalance from different sizes of objects, and therefore $\SMAE$ is equivalent to $\MAE$.

{\color{blue}{\textbf{Case 2:}}} Suppose there are two salient objects ($K=2$) where $f_A$ and $f_B$ detect the same amount of salient pixels in an image $X$. Meanwhile, assume that $f_A$ only predicts $C_2$ perfectly while $f_B$ could somewhat recognize $C_1$ and $C_2$ partially. In this case, \underline{$f_B$ should still be better than $f_A$} since the latter totally fails on $C_1$. Unfortunately, if $S_{1}^{fore}<S_{2}^{fore}$, $\MAE(f_A)=\MAE(f_B)$ holds but we have $\SMAE(f_A) > \SMAE(f_B)$.

\begin{proof}
    For {\color{blue}{\textbf{Case 1}}}, as there is only one salient object, we can easily divide the image into $X_1^{fore}$ and $X_2^{back}$, and the weight $\alpha=S_2^{back}/S_1^{fore}=(S-S_1^{fore})/S_1^{fore}$. Therefore, we have the following:
    \begin{equation}
    \begin{aligned}
        \MAE(f)&=\frac{\Vert f(X_1^{fore})-Y_1^{fore}\Vert_{1,1} + \Vert f(X_2^{back})-Y_2^{back}\Vert_{1,1}}{S}, \\
        \SMAE(f)&=\frac{1}{1+\alpha} \cdot  \left[ \frac{\Vert f(X_1^{fore})-Y_1^{fore}\Vert_{1,1}}{S_1^{fore}} + \alpha \cdot \frac{\Vert f(X_2^{back})-Y_2^{back}\Vert_{1,1}}{S_2^{back}}  \right] \\
        &=\frac{1}{1+\frac{S-S_1^{fore}}{S_1^{fore}}} \cdot \left[  \frac{\Vert f(X_1^{fore})-Y_1^{fore}\Vert_{1,1}}{S_1^{fore}} + \frac{S_2^{back}}{S_1^{fore}} \cdot \frac{\Vert f(X_2^{back})-Y_2^{back}\Vert_{1,1}}{S_2^{back}}  \right] \\
        &=\frac{S_1^{fore}}{S} \cdot \left[  \frac{\Vert f(X_1^{fore})-Y_1^{fore}\Vert_{1,1} + \Vert f(X_2^{back})-Y_2^{back}\Vert_{1,1}}{S_1^{fore}} \right] \\
        &=\frac{\Vert f(X_1^{fore})-Y_1^{fore}\Vert_{1,1} + \Vert f(X_2^{back})-Y_2^{back}\Vert_{1,1}}{S} \\
        &=\MAE(f).
    \end{aligned}
    \end{equation}

For {\color{blue}{\textbf{Case 2}}}, we first suppose $\rho\in [0,1]$ of $C_2$ is correctly detected by $f_B$. For convenience, we consider errors just in foreground frames. According to the proposition, $f_A$ and $f_B$ detect the same amount of salient pixels, and therefore it is clear that $\MAE(f_A)=\MAE(f_B)$. As there are two salient objects, we have $\alpha=S_3^{back} / (S_1^{fore}+S_2^{fore})$, and thus $\SMAE$ is calculated as follows:
\begin{equation}
    \begin{aligned}
        \SMAE(f_A) &= \frac{1}{2+\alpha} \cdot \frac{|C_1|}{S_1}, \\
        \SMAE(f_B) &= \frac{1}{2+\alpha} \cdot \left[ \frac{|C_2|(1-\rho)}{S_2} + \frac{|C_1|-|C_2|(1-\rho)}{S_1}  \right],
    \end{aligned}
\end{equation}
then when $S_1^{fore}<S_2^{fore}$, we have:
\begin{equation}
\begin{aligned}
    \SMAE(f_B)-\SMAE(f_A) & = \frac{1}{2+\alpha} \cdot \left[ \frac{|C_2|(1-\rho)}{S_2} + \frac{|C_1|-|C_2|(1-\rho)}{S_1} - \frac{|C_1|}{S_1} \right] \\
    &= \frac{1}{2+\alpha} \cdot \left[  \frac{|C_2|(1-\rho)}{S_2} - \frac{|C_2|(1-\rho)}{S_1} \right] \\
    &= \frac{|C_2|(1-\rho)}{2+\alpha}\cdot \left( \frac{1}{S_2} - \frac{1}{S_1} \right) \\
    & <0.
\end{aligned}
\end{equation}
This completes the proof.
    
\end{proof}

\textbf{Extension of \cref{prop1}.}
Still, suppose $f_A, f_B$ detect the same amount of salient pixels in an image $X$. We denote $K$ as the number of salient objects in image $X$, with $S_1\le S_2 \le \cdots \le S_K$. Meanwhile, assume $f_A$ only predicts $C_{m+1},\cdots,C_K$ perfectly while $f_B$ could recognize $C_1,\dots,C_K$ partially. In this case, $f_B$ should still be better than $f_A$ since the latter totally fails on $C_1,\cdots,C_m$. Unfortunately when $S_{m+1}$ is sufficiently large, which means that larger objects dominate smaller ones, $\mathsf{MAE}(f_A)=\mathsf{MAE}(f_B)$ holds while $\mathsf{SI\text{-}MAE}(f_A)>\mathsf{SI\text{-}MAE}(f_B)$.

\begin{proof}
    We first suppose $[\rho_1,\rho_2,\cdots,\rho_K],\rho_i\in [0,1]$ of $C_1,C_2,\cdots,C_K$ are correctly detected by $f_B$. For convenience, we consider errors just in foreground frames. According to the proposition above, $f_A$ and $f_B$ detect the same amount of salient pixels, and therefore $\mathsf{MAE}(f_A)=\mathsf{MAE}(f_B)$. Furthermore, we have the equation below:
    \begin{equation}
        \sum_{i=1}^K \rho_i|C_i|=\sum_{i=m+1}^K|C_i|.
    \end{equation}
As there are $K$ salient objects, we have $\alpha=\frac{S_{K+1}^{back}}{\sum_{i=1}^KS_i^{fore}}$, and thus $\mathsf{SI\text{-}MAE}$ is calculated as follows:
\begin{equation}
    \begin{aligned}
         \mathsf{SI\text{-}MAE}(f_A)&=\frac{1}{K+\alpha}\sum_{i=1}^m\frac{|C_i|}{S_i}, \\
         \mathsf{SI\text{-}MAE}(f_B)&=\frac{1}{K+\alpha}\sum_{i=1}^K \frac{(1-\rho_i)|C_i|}{S_i}.
    \end{aligned}
\end{equation}
For clear representation, we denote $t=(t_1,\cdots,t_K)$ where $t_i=\rho_i|C_i|$ and $S=(S_1,\cdots,S_K)$.

Therefore, when $S_{m+1}>\frac{\left\langle t,\textbf{1}\right\rangle}{\left\langle t,\frac{1}{S}\right\rangle}$, we have:
\begin{equation}
\begin{aligned}
    \mathsf{SI\text{-}MAE}(f_B)-\mathsf{SI\text{-}MAE}(f_A)&=\frac{1}{K+\alpha}\left[\sum_{i=1}^K\frac{(1-\rho_i)|C_i|}{S_i}-\sum_{i=1}^m\frac{|C_i|}{S_i} \right] \\
    &=\frac{1}{K+\alpha} \left[\sum_{i=m+1}^K \frac{|C_i|}{S_i}-\sum^{K}_{i=1}\frac{\rho_i|C_i|}{S_i} \right] \\
    & \le \frac{1}{K+\alpha} \left[\frac{\sum_{i=m+1}^K |C_i|}{S_{m+1}}-\sum^{K}_{i=1}\frac{\rho_i|C_i|}{S_i}\right] \\
    &=\frac{1}{(K+\alpha)} \left[ \frac{\sum _{i=1}^K\rho_i|C_i|}{S _{m+1}}- \sum^{K} _{i=1}\frac{\rho_i|C_i|}{S_i}\right] \\
    &=\frac{1}{(K+\alpha)}\left[\frac{<t,1 >}{S _{m+1}}-<t,\frac{1}{S}>\right] \\
    &<0.
\end{aligned}
\end{equation}
 
\end{proof}

\subsection{Proof for \cref{prop3}}
\textbf{Restate of \cref{prop3}.} Given two different predictors $f_A$ and $f_B$, the following case suggest that $\SF$ is more effective than $\F$-score during evaluation.

Suppose there are two salient objects ($K=2$) where $f_A$ and $f_B$ detect the same amount of salient pixels for an image $X$. Meanwhile, assume that $f_A$ only predicts $C_2$ perfectly while $f_B$ could somewhat recognize $C_1$ and $C_2$ partially. In this case, \underline{$f_B$ should still be better than $f_A$} since the latter totally fails on $C_1$. Unfortunately, $\F(f_A)=\F(f_B)$ holds but we have $\SF(f_A) < \SF(f_B)$.

\begin{proof} \label{prop3_proof}  
We first suppose $\rho\in [0,1]$ of $C_2$ is correctly detected by $f_B$. For convenience, we consider errors just in foreground frames. According to the proposition, $f_A$ and $f_B$ detect the same amount of salient pixels, and therefore it is clear that $\MAE(f_A)=\MAE(f_B)$. As there are two salient objects, we have $\alpha=S_3^{back} / (S_1^{fore}+S_2^{fore})$, and thus $\SF$ is calculated as follows:
\begin{equation}
    \begin{split}
        \SF(f_A)&=\frac{1}{2}\cdot(1+0)=\frac{1}{2}, \\
        \SF(f_B)&=\frac{1}{2}\cdot\left[\frac{2\rho}{1+\rho}+\frac{2(1-\rho)|C_2|}{|C_1|+(1-\rho)|C_2|}\right],
    \end{split}
\end{equation}
Therefore, we have $\SF(f_B) > \SF(f_A)$ because
\begin{equation}
    \begin{split}
        \SF(f_B)-\SF(f_A) &= \frac{1}{2}\cdot\left[\frac{2\rho}{1+\rho}+\frac{2(1-\rho) |C_2|}{|C_1|+(1-\rho)|C_2|}\right]-\frac{1}{2} \\
        &=\frac{1}{2}\cdot\left[\frac{1-\rho}{1+\rho}+\frac{2(1-\rho) |C_2|}{|C_1|+(1-\rho) |C_2|}\right] \\
        &=\frac{1-\rho}{2}\cdot\left[\frac{1}{1+\rho}+\frac{2 |C_2|}{|C_1|+(1-\rho) |C_2|}\right] \\
        & >0.
    \end{split}
\end{equation}
This completes the proof.
\end{proof}

\section{Proof for Properties of SI-SOD Loss}
\subsection{Proof for \cref{re-attention_prop}}
\textbf{Restate of \cref{re-attention_prop}} \label{re-attention_proof}
Given a separable loss function $\ell(\cdot)$ and its corresponding size-invariant loss $\mathcal{L}_{\SI}(\cdot)$, then:
\begin{enumerate}
    \item when $S_{i}<\frac{S}{K +\alpha}$, we have $w_{\mathcal{L}_{\SI}}(x_i)>w_{\ell}(x_i)$,
    \item when $S_{i} < S_{j},$ we have $w_{\mathcal{L}_{\SI}}(x_i) > w_{\mathcal{L}_{\SI}}(x_j)$.
\end{enumerate}
where $w_{\mathcal{L}_{\SI}}(x_i)$ is the \textbf{weight} of pixel-level loss in $X_i$ with $\mathcal{L}_{\SI}$, and $w_{\ell}(x_i)$ is the \textbf{weight} of pixel-level loss in $X_i$ with the original loss $\ell$.

\begin{proof}
Common separable loss functions include Cross Entropy($\CE$), Mean Absolute Error($\MAE$), Mean Square Error($\MSE$), etc. 

For Item 1, we have the following as mentioned in \cref{Revisting}:
\begin{equation}
    \ell(f)=\frac{1}{S} \sum_{i=1}^{S} \ell(f(x_i), y_i),
\end{equation}
where the weight of pixel-level loss $w_{\ell}(x_i) = 1/S$.

When using the size-invariant loss $\mathcal{L}_{\SI}$, we have:
\begin{equation}
\begin{aligned}
    \mathcal{L}_{\SI}(f) &= \frac{1}{K+\alpha} \left[  \sum_{i=1}^K  \ell(f_i) +\alpha \ell(f_{K+1}) \right] \\
    &= \frac{1}{K+\alpha}\left[ \sum_{i=1}^K  \frac{\sum_{j \in X_i} \ell(f(x_j), y_j)}{S_i} + \alpha\frac{\sum_{j \in X_{K+1}} \ell(f(x_j), y_j)}{S_{K+1}}\right],
\end{aligned}
\end{equation}
where the weight of foreground pixel-level loss $w_{\mathcal{L}_{\SI}}(x_i)=\frac{1}{(K+\alpha)S_i}$. Therefore, when $S_{i}<\frac{S}{K +\alpha}$, we have $w_{\mathcal{L}_{\SI}}(x_i)>w_{\ell}(x_i)$.

For Item 2, it is obvious that $w_{\mathcal{L}_{\SI}}(x_i)$ is in proportion to $1/{S_i}$. Therefore, when $S_{i} < S_{j}$, we have $w_{\mathcal{L}_{\SI}}(x_i) > w_{\mathcal{L}_{\SI}}(x_j)$.
\end{proof} 

\subsection{Proof for \cref{generalization_bound}} \label{generalization_bound_proof}

\subsubsection{Proof for Technical Lemmas}
In this subsection, we present key lemmas that are important for the proof of \cref{generalization_bound}.
\begin{lemma}
\label{lemma1}
The empirical Rademacher complexity of function $g$ with respect to the predictor $f$ is defined as:
\begin{equation}
    \hat{\mathfrak{R}}_{\mathcal F}(g)=\mathbb{E}_{\sigma}[\sup _{f \in \mathcal F}\frac{1}{N} \sum_{i=1}^{N} \sigma_ig(f^{(i)})].
\end{equation}
where $\mathcal{F} \subseteq \{f: \mathcal{X} \to \mathbb{R}^K\}$ is a family of predictors, and $N$ refers to the size of the dataset, and $\sigma_i$s are independent uniform random variables taking values in $\{-1, +1\}$. The random variables $\sigma_i$ are called Rademacher variables.
\end{lemma}

\begin{lemma}
\label{lemma2}
Let $\mathbb{E}[g]$ and $\hat{\mathbb{E}}[g]$ represent the expected risk and empirical risk, and $\mathcal{F} \subseteq \{ f:\mathcal{X} \to \mathbb{R}^K \}$. Then with probability at least $1-\delta$ over the draw of an i.i.d. sample S of size $m$, the generalization bound holds:
\begin{equation}
    \sup_{f \in \mathcal{F}}\big(\mathbb{E}[g(f)]-\hat{\mathbb{E}}[g(f)] \big) \le 2\hat{\mathfrak{R}}_{\mathcal F}(g)+3\sqrt{\frac{\log \frac{2}{\delta}}{2m}}.
\end{equation}
\end{lemma}

\begin{lemma}
\label{lemma3}
\cite{vector_Contraction} Assume $\mathcal{F} \subseteq \{ f:\mathcal{X} \to \mathbb{R}^K \}$, and $(\sigma_1, \cdots, \sigma_n)$ is a sequence of i.i.d Rademacher random variables. When $\phi_1, \cdots, \phi_n$ are $L$-Lipschitz with respect to the $\ell_{\infty}$ norm, there exists a constant $C>0$ for any $\delta>0$, such that if $|\phi_t(f(x))| \vee \Vert f(x) \Vert_\infty \le \beta$, the following holds:
\begin{equation}
    \mathfrak{R}(\phi \circ \mathcal{F} ; x_{1:n}) \le C \cdot L \sqrt K \cdot \max_i \mathfrak{R}_n(\mathcal{F}|_9)\cdot\log^{\frac{3}{2}+\delta}\left(\frac{\beta n}{\max_i \mathfrak{R}_n(\mathcal{F}|_i)}\right).
\end{equation}
where
\begin{equation}
    \mathfrak{R}(\phi \circ \mathcal{F} ; x_{1:n}) = \mathbb{E}_{\epsilon} \sup_{f\in \mathcal{F}} \sum_{t=1}^n \epsilon_t f(x_t),
\end{equation}
where $\mathcal{F}$ is a class of functions $f:\mathcal{X} \to \mathbb{R}$, and $\epsilon=(\epsilon_1, \cdots, \epsilon_n)$ is a sequence of $i.i.d.$ Rademacher random variables.
Here we set $\beta=1$ because it is obvious that $\Vert f(x) \Vert_\infty \le 1$ and $\phi_t(x) \le 1$. Therefore, 
\begin{equation}
    \mathfrak{R}(\phi \circ \mathcal{F} ; x_{1:n}) \le C \cdot L \sqrt K \cdot \max_i \mathfrak{R}_n(\mathcal{F}|_9)\cdot\log^{\frac{3}{2}+\delta}\left(\frac{n}{\max_i \mathfrak{R}_n(\mathcal{F}|_i)}\right).
\end{equation}
\end{lemma}

\begin{lemma}
\label{lemma4}
When $g(\cdot)$ is Lipschitz continuous, the following holds:
\begin{equation}
    \Vert g(x)-g(\tilde{x})\Vert_{\infty} \le \sup \Vert\nabla_x g\Vert_p  \cdot \Vert x-\tilde x\Vert_q,
\end{equation}
where $\frac{1}{p}+\frac{1}{q}=1$.

\begin{proof}
\begin{equation}
\begin{aligned}
    |g(x)-g(\tilde{x})| &=~ \left|\int_0^1 \left\langle\nabla g(\tau x + (1-\tau)\tilde{x}), x-\tilde{x} \right\rangle d\tau\right| \\
    & \le~ \sup_{ x \in \mathcal{X}} \big[ \Vert \nabla g \Vert_p \big] \cdot  \big\Vert x-\tilde{x}\big\Vert_q 
\end{aligned}
\end{equation}
\end{proof}

Specifically, when $p=1$ and $q=\infty$, we have
\begin{equation}
    \Vert g(x)-g(\tilde{x})\Vert_{\infty} \le \sup \Vert\nabla_x g\Vert_1  \cdot \Vert x-\tilde x\Vert_\infty.
\end{equation}
\end{lemma}

\begin{lemma}
\label{lemma5}
Common composite functions are $p$-Lipschitz, as \cite{p_Lipschitz} stated:
\begin{definition}[$p$-Lipschitzness]
    $\Phi(u,v,p)$ is said to be $p$-Lipschitz if:
\begin{equation}
    | \Phi(u,v,p)-\Phi(u',v',p') | \le U_p |u-u'| + V_p |v-v'| + P_p |p-p'|,
\end{equation}
where $\Phi(u(h), v(h), p)$ is defined as:
\begin{equation}
    u(h) = \text{TP}(h), \qquad v(h)=\mathbb{P}(h=1), \text{and} \quad p=\mathbb{P}(y=1).
\end{equation}
\end{definition}
Any metric being a function of the confusion matrix can be parameterized in this way.
According to \cite{p_Lipschitz}, Accuracy, AM, F-score, Jaccard, G-Mean, and AUC are all $p$-Lipschitz.
\end{lemma}

\subsubsection{Proof for the Generalization Bound}
\textbf{Restate of \cref{generalization_bound}} Assume $\mathcal{F} \subseteq \{ f:\mathcal{X} \to \mathbb{R}^K \}$, where $K=H \times W$ is the number of pixels in an image,  $g^{(i)}$ is the risk over $i$-th sample, and is $L$-Lipschitz with respect to the $l_{\infty}$ norm, (i.e. $\Vert g(x)-g(\tilde{x})\Vert_\infty \le L\cdot \Vert x-y \Vert_{\infty}$). When there are $N$ $i.i.d.$ samples, there exists a constant $C>0$ for any $\epsilon > 0$, the following generalization bound holds with probability at least $1-\delta$:
\begin{equation}
\begin{aligned}
    &~ \sup_{f \in \mathcal{F}}(\mathbb{E}[g(f)]-\hat{\mathbb{E}}[g(f)]) \\
    \le&~  C\cdot \frac{L\sqrt{K}}{N} \cdot \max_i \mathfrak{R}_N(\mathcal{F}|_i)\cdot\log^{\frac{3}{2}+\epsilon}\left(\frac{N}{\max_i \mathfrak{R}_N(\mathcal{F}|_i)}\right) \\
    &+3\sqrt{\frac{\log \frac{2}{\delta}}{2N}},
\end{aligned}
\end{equation}
where again $g(f(X,Y)):=g(f)$, $\mathbb{E}[g(f)]$ and $\hat{\mathbb{E}}[g(f)]$ represent the expected risk and empirical risk, and $\mathfrak{R}_N(\mathcal{F}|_i)=\max_{x_{1:N}\in \mathcal{X}} \mathfrak{R}(\mathcal{F};x_{1:N})$ denotes the worst-case Rademacher complexity. Specifically, 
\begin{enumerate}[label=\textbf{\textit{ Case \arabic*:}},leftmargin=4em]
    \item For separable loss functions $\ell(\cdot)$, if it is $\mu$-Lipschitz, we have $L=\mu$.
    \item For composite loss functions, when $\ell(\cdot)$ is DiceLoss \cite{DiceLoss}, we have $L=\frac{4}{\rho}$, where $\rho=\min \frac{S_{l}^{1,i}}{S_{l}^i}$, which represents the minimum proportion of the salient object in the $l$-th frame within the $i$-th sample.
\end{enumerate}
\begin{proof}
In this subsection, we give the proof combining lemmas above. 

Firstly, the empirical risk over the dataset is:
\begin{equation}
    \hat{\mathbb E}[g(f)]=\frac{1}{N}\sum_{i=1}^N g^{(i)}(f(X),Y),
\end{equation}
where $X, Y$ are the prediction and ground truth, and
\begin{equation}
    g^{(i)}(f(X),Y)=\frac{1}{|N_l^i|} \sum_{l \in [N_c^i]} \ell (f^{(i)}(X_{N_l}),Y^{(i)}_{N_l}),
\end{equation}
where $N_l$ stands for $l$-th foreground frame, and therefore $\ell(\cdot)$ is a matrix element function.

Combing \cref{lemma1}, \cref{lemma2} and \cref{lemma3}, with probability at least $1-\delta$, we have:
\begin{equation}
\begin{aligned}
    \sup_{f \in \mathcal{F}}(\mathbb{E}[g(f)]-\hat{\mathbb{E}}[g(f)]) & \le 2\hat{\mathfrak{R}}_{\mathcal F}(g)+3\sqrt{\frac{\log \frac{2}{\delta}}{2N}} \\
    & = 2\mathbb{E}_{\sigma}\left[\sup _{f \in \mathcal F}\frac{1}{N} \sum_{i=1}^{N} \sigma_ig(f^{(i)})\right]+3\sqrt{\frac{\log \frac{2}{\delta}}{2N}} \\
    & \le C\cdot \frac{L\sqrt{K}}{N} \cdot \max_i \mathfrak{R}_n(\mathcal{F}|_i)\cdot\log^{\frac{3}{2}+\epsilon}\left(\frac{n}{\max_i \mathfrak{R}_n(\mathcal{F}|_i)}\right)+3\sqrt{\frac{\log \frac{2}{\delta}}{2N}}.
\end{aligned}
\end{equation}

For \textbf{\textit{Case 1}} separable loss functions $\ell(\cdot)$, we have the following equation according to definition \cref{eq:separable function}:
\begin{equation}
    g^{(i)}(f)=\frac{1}{N_c^{(i)}}\sum_{l \in [N_c^i]} \frac{1}{|N_l^i|} \sum_{(j,k) \in N_{l}^i} \ell (f_{j,k}^{(i)}, Y_{j,k}^{(i)}).
\end{equation}
Assume $\ell(\cdot)$ is $\mu$-Lipschitz, then for Lipschitz continuous of $g^{(i)}$, we have:
\begin{equation}
    \begin{aligned}
g^{(i)}(f)-g^{(i)}(\tilde{f})
&=\frac{1}{N_c^{(i)}}\sum_{l \in [N_c^i]} \frac{1}{|N_l^i|} \sum_{(j,k) \in N_{l}^i}  \left(\ell(f_{j,k}^{(i)}, Y_{j,k}^{(i)}) - \ell(\tilde{f}_{j,k}^{(i)}, Y_{j,k}^{(i)})\right) \\
& \le \frac{1}{N_c^{(i)}}\sum_{l \in [N_c^i]} \frac{1}{|N_l^i|} \sum_{(j,k) \in N_{l}^i}  \max_{(j,k)} \left|\left(\ell (f_{j,k}^{(i)}, Y_{j,k}^{(i)}) - \ell (\tilde{f}_{j,k}^{(i)}, Y_{j,k}^{(i)})\right)\right| \\
& \le \frac{1}{N_c^{(i)}}\sum_{l \in [N_c^i]} \frac{1}{|N_l^i|} \sum_{(j,k) \in N_{l}^i} \mu \max_{(j, k)}\left|f_{j,k}^{(i)}-\tilde{f}_{j,k}^{(i)}\right| \\
& = \mu \Vert f^{(i)}-\tilde{f}^{(i)}\Vert_{\infty},
\end{aligned}
\end{equation}
where we use $f_{j,k}^{(i)}$ to denote $f^{(i)}(X_{(j,k)\in N_l^i})$ as abbreviation. We can always bound $\ell (f_{j,k}^{(i)}, Y_{j,k}^{(i)}) - \ell (\tilde{f}_{j,k}^{(i)}, Y_{j,k}^{(i)})$ with the maximum element $\max_{(j,k)} |(\ell(f_{j,k}^{(i)}, Y_{j,k}^{(i)}) - \ell(\tilde{f}_{j,k}^{(i)}, Y_{j,k}^{(i)}))|$ because there are finite pixels in an image.
Therefore, for separable loss $g^{(i)}$, we let $L=\mu$, and complete the proof.

For \textbf{\textit{Case 2}} composite loss, when $\ell(\cdot)$ is DiceLoss \cite{DiceLoss}, we have the following equation:
\begin{equation}
    g^{(i)}(f)=\frac{1}{N_c^i} \sum_{l=1}^{N_c^i} \left[1-\frac{2 \sum_{(j, k) \in N_l^i} Y^{(i)}_{j,k}\cdot f^{(i)}_{j,k}}{\sum_{(j, k) \in N_l^i} Y^{(i)}_{j,k}  + \sum_{(j, k) \in N_l^i}f^{(i)}_{j,k}}\right].
\end{equation}
Considering that formally $\text{DiceLoss}=1-\F$, combining \cref{lemma4} and \cref{lemma5}, we turn to solve $\sup \Vert\nabla_x g\Vert_1  \cdot \Vert$ instead of directly pursuing the Lipschitz constant with respect to $\ell_\infty$ norm. Therefore, we can find the Lipschitz continuous of $g^{(i)}$:
\begin{equation}
    \begin{aligned}
\left\Vert\frac{\partial g^{(i)}}{\partial f^{(i)}_{j,k}}\right\Vert_1 
&= 2\cdot\left|\frac{Y_{j,k}^{(i)}\cdot\left(\sum_{(j, k) \in N_l^i} Y^{(i)}_{j,k}  + \sum_{(j, k) \in N_l^i}f^{(i)}_{j,k}\right)-\sum_{(j, k) \in N_l^i} Y^{(i)}_{j,k}\cdot f^{(i)}_{j,k}}{\left(\sum_{(j, k) \in N_l^i} Y^{(i)}_{j,k}  + \sum_{(j, k) \in N_l^i}f^{(i)}_{j,k}\right)^2}\right| \\
& \le 2\cdot \left(\left|\frac{Y^{(i)}_{j,k}}{\sum_{(j, k) \in N_l^i} Y^{(i)}_{j,k}  + \sum_{(j, k) \in N_l^i}f^{(i)}_{j,k}}\right| + \left|\frac{\sum_{(j, k) \in N_l^i} Y^{(i)}_{j,k}\cdot f^{(i)}_{j,k}}{(\sum_{(j, k) \in N_l^i} Y^{(i)}_{j,k}  + \sum_{(j, k) \in N_l^i}f^{(i)}_{j,k})^2}\right|\right) \\
& \le 2\cdot\left(\frac{1}{\sum_{(j, k) \in N_l^i} Y^{(i)}_{j,k}  + \sum_{(j, k) \in N_l^i}f^{(i)}_{j,k}} + \frac{\sum_{(j, k) \in N_l^i} Y^{(i)}_{j,k}}{\left(\sum_{(j, k) \in N_l^i} Y^{(i)}_{j,k}  + \sum_{(j, k) \in N_l^i}f^{(i)}_{j,k}\right)^2}\right) \\
& \le 2\cdot \left(\frac{1}{\sum_{(j, k) \in N_l^i} Y^{(i)}_{j,k}} + \frac{\sum_{(j, k) \in N_l^i} Y^{(i)}_{j,k}}{\left(\sum_{(j, k) \in N_l^i} Y^{(i)}_{j,k} \right)^2}\right)\\
&= \frac{4}{\sum_{(j, k) \in N_l^i} Y^{(i)}_{j,k}} \\
&=\frac{4}{S_{l}^{1,i}},
\end{aligned}
\end{equation}
where $S_{l}^{1,i}$ stands for the area of the object in $i$-th frame, and $S_{l}^{i}$ stands for the area of the $i$-th frame.
Therefore, for a frame in  the image,
\begin{equation}
    \begin{aligned}
\Vert\nabla g^{(i)}\Vert_1 &=\frac{1}{N_c^i}\sum_{l=1}^{N_c^i}
\sum_{(j,k) \in N_l^{i}}|\nabla g^{(i)}_{j,k}| \\
&= \frac{1}{N_c^i}\sum_{l=1}^{N_c^i} \left\Vert\frac{\partial g^{(i)}}{\partial f^{(i)}_{j,k}}\right\Vert_1  \cdot S_l^i \\
& \le \frac{1}{N_c^i}\sum_{l=1}^{N_c^i} 4\cdot \frac{S_l^i}{S_{l}^{1,i}} \\
& \le \frac{4}{\rho},
\end{aligned}
\end{equation}   
where $0<\rho \le \frac{S_{l}^{1,i}}{S_{l}^i}$, which depicts the threshold, how much proportion the object occupies in the corresponding frame. Therefore, taking DiceLoss as an example of composite loss, we let $L=\frac{4}{\rho}$, and complete the proof.
\end{proof}

\section{Additional Experiment Settings} \label{Experiments_appendix}
In this section, we make a supplementation to \cref{Experiments}.

\subsection{Datasets} \label{dataset_appendix}
Here we give more detailed introductions to the five datasets used in the experiments, as shown in \cref{tab:dataset_stat}.

\begin{table}[ht]
    \centering
    \scalebox{0.8}{
    \begin{tabular}{c|c|c}
    \toprule
    Dataset &  Scale & Characteristics \\
    \midrule
    DUTS~\cite{DUTS}    & 10,553 + 5,019 & Training set (10,553), as well as test set (5,019), is provided. \\
    ECSSD~\cite{ECSSD} & 1,000 & Semantically meaningful but structurally complex contents are included.  \\
    DUT-OMRON~\cite{DUT-OMRON} & 5,168 & It is characterized by complex background and diverse contents. \\
    HKU-IS~\cite{HKU-IS} & 4,447 & Far more multiple disconnected objects are included. \\ 
    MSOD~\cite{MSOD} & 300 & It consists of the most challenging multi-object scenarios with 1342 objects in total. \\
    PASCAL-S~\cite{ECSSD} & 850 & Images are from PASCAL VOC 2010 validation set with multiple salient objects. \\
    SOD~\cite{SOD} & 300 & Many images have more than one salient object that is similar to the background. \\
    XPIE~\cite{XPIE} & 10,000 & It covers many complex scenes with different numbers, sizes and positions of salient objects. \\
    \bottomrule
    \end{tabular}
    }
    \caption{Statistics on Datasets.}
    \label{tab:dataset_stat}
\end{table}

\textbf{DUTS} is a widely used large-scale dataset, consisting of the training set DUTS-TR including 10,553 images, and the test set DUTS-TE including 5,019 images. All the images are sampled from the ImageNet DET training and test set~\cite{ImageNet}, and some test images are also collected from the SUN data set~\cite{SUN}. It is common practice that SOD models are trained on DUTS-TR and tested on other datasets.

\textbf{ECSSD}, namely \textbf{E}xtended \textbf{C}omplex \textbf{S}cene \textbf{S}aliency \textbf{D}ataset, consists of 1,000 images with complex scenes, presenting textures and structures. One of the characteristics is that this dataset includes many semantically meaningful but structurally complex images. The images are acquired from the internet and 5 helpers are asked to produce the ground truth masks individually.

\textbf{DUT-OMRON} is composed of 5,168 images with complex backgrounds and diverse content. Images in this dataset have one or more salient objects and a relatively complex background. 

\textbf{HKU-IS} has 4,447 images with relatively more multi-object scenarios. Particularly, around 50\% images in this dataset have multiple disconnected salient objects, far beyond the three datasets above.

\textbf{MSOD} contains the most challenging multi-object scenes across the common SOD datasets. It comprises 300 test images with 1342 total objects. The dataset comprises a variety of object classes and a varied number of these objects across the image.

\textbf{PASCAL-S} is a dataset for salient object detection consisting of a set of 850 images from PASCAL VOC 2010 validation set with multiple salient objects on the scenes.

\textbf{SOD} consists of 300 images, constructed from ~\cite{SOD_construct}. Many images have more than one salient object that is similar to the background or touches image boundaries.

\textbf{XPIE} contains 10,000 images with pixel-wise masks of salient objects. It covers many complex scenes with different numbers, sizes and positions of salient objects.

\subsection{Competitors} \label{competitor_appendix}
Here we give a more detailed summary of the backbones mentioned in the experiments.

\textbf{PoolNet}~\cite{PoolNet} is a widely used baseline. They design various pooling-based modules for the first time to assist in improving the performance of SOD. Specifically, the model consists of two primary modules based on the feature pyramid networks, namely a global guidance module (GGM), and a feature aggregation module (FAM). GGM is an individual module, where high-level semantic information can be transmitted to all pyramid layers, and FAM aims at capturing local context information at different scales and then combining them with different weights. 

\textbf{LDF}~\cite{LDF} contains a level decoupling procedure and a feature interaction network. It decomposes the saliency label into the body map and detail map to supervise the model. The feature interaction network is introduced to make full use of the complementary information between branches. Both branches iteratively exchange information to produce more precise saliency maps.

\textbf{GateNet}~\cite{GateNet} proposes a gated network to adaptively control the amount of information flowing into the decoder. The multilevel gate units help to balance the contribution of each encoder and suppress the information in non-salient regions. An ASPP module is exploited to capture richer context information. With a dual-branch architecture, it forms a residual structure, which complements each other to generate better results.

\textbf{ICON}~\cite{ICON} proposes micro-integrity and macro-integrity, aiming to focus on whole-part relevance within a single salient object and to identify all salient objects within the given image scene. It is composed of three parts: diverse feature aggregation, integrity channel enhancement, and part-whole verification. 

\textbf{EDN}~\cite{EDN} directly utilizes an extreme down-sampling technique to capture effective high-level features for SOD and achieved competitive performance with high computational efficiency. The proposed Extremely Downsampled Block is to learn a global view of the whole image. It only introduces a tiny computational overhead but achieves competitive performance with a fast inference speed.

\subsection{Implementation Details} \label{details_appendix}
The experiment platform is Ubuntu 18.04.5 LTS with Intel(R) Xeon(R) Gold 6246R CPU @ 3.40GHz. We implement our method with Python 3.8.13 and torch 2.0.1. For specific backbones, we report the settings as follows:
\begin{itemize}
    \item For EDN, following ~\cite{EDN}, Adam optimizer with parameters $\beta_1=0.9$, and $\beta_2=0.99$ is adopted, with weight decay $10^{-4}$ and batch size 36. The initial learning rate is set to $5 \times 10^{-5}$ with a poly learning rate strategy and the training lasts for 100 epochs in total.
    \item For ICON, we use the SGD optimizer, with the initial learning rate as $10^{-2}$, the weight decay as $10^{-4}$, and the momentum as $0.9$. The batch size is set to 36, and the training lasts 100 epochs.
    \item For GateNet~\cite{GateNet}, we fully follow ~\cite{GateNet} to utilize the SGD optimizer, with the initial learning rate as $10^{-3}$, momentum as $0.9$ and the weight decay as $5 \times 10^{-4}$. The batch size is set to 12, and the network is iterated within $10^5$ times.
    \item For LDF, as it is a two-stage framework, we integrate our method into the second stage, as most previous works do. Specifically, we use the SGD optimizer with the initial learning rate as $5 \times 10^{-2}$, momentum as $0.9$ and the weight decay as $5 \times 10 ^{-4}$. The batch size is set to 32 and the training in the second stage lasts 40 epochs.
    \item For PoolNet, we fully follow ~\cite{PoolNet} to utilize the Adam optimizer with the initial learning rate as $5\times 10^{-5}$, and the weight decay as $5 \times 10^{-4}$. The batch size to $1$, and the network is iterated every 10 gradients are accumulated. The training lasts 24 epochs.
\end{itemize}

\subsection{Optimization Details for Different Backbones} \label{loss_description}
Following \cref{eq:loss}, here we give a detailed description of the implementation of optimization for different backbones.
\begin{itemize}
    \item For EDN, the original loss function is
    \begin{equation}
        \mathcal{L} = \mathsf{BCE}(f) + \mathsf{DiceLoss}(f),
    \end{equation}
    and we modify it into
    \begin{equation}
        \mathcal{L} = \sum_{k=1}^K [\mathsf{BCE}(f_k^{fore}) + \mathsf{DiceLoss}(f_k^{fore})] + \alpha \mathsf{BCE}(f_{K+1}^{back}).
    \end{equation}
    Specifically, DiceLoss is defined as:
    \begin{equation}
        \mathsf{DiceLoss}(\cdot)=\frac{2\cdot \sum_{i}^N p_i \cdot g_i}{\sum_{i}^N p_i^2+\sum_{i}^N g_i^2}, 
    \end{equation}
    where the sums run over the $N$ pixels, and $p_i, g_i$ represent the prediction and ground truth, respectively.

    \item For ICON, the original loss function is
    \begin{equation}
         \mathcal{L} = \mathsf{BCE}(f) + \mathsf{IOULoss}(f),
    \end{equation}
    and we modify it into
    \begin{equation}
        \mathcal{L} = \sum_{k=1}^K [\mathsf{BCE}(f_k^{fore}) + \mathsf{IOULoss}(f_k^{fore})] + \alpha \mathsf{BCE}(f_{K+1}^{back}),
    \end{equation}
    where IOULoss is defined as
    \begin{equation}
        \mathsf{IOULoss}(\cdot)=1-\frac{\sum_i^N (p_i \cdot g_i)}{\sum_i^N (p_i+d_i)-\sum_i^N (p_i \cdot g_i)},
    \end{equation}
    where $p_i$ and $g_i$ represent the prediction and ground truth.

    \item For GateNet, the original loss function is
    \begin{equation}
        \mathcal{L}=\ell(f),
    \end{equation}
    where $\ell(\cdot)$ is anyone among the binary cross entropy, mean square error, and L1 loss. We modify it into
    \begin{equation}
        \mathcal{L} = \sum_{k=1}^K \ell(f_k^{fore}) + \alpha \ell(f_{K+1}^{back}).
    \end{equation}

    \item For LDF, the original loss function is the same as that in ICON. Therefore, we adopt the same modification.

    \item For PoolNet, the original loss function is the same as that in GateNet. Therefore, we adopt the same modification.
\end{itemize}

\clearpage

\section{Additional Experiment Analysis}
\subsection{Quantitative comparisons} \label{Quantitative_appendix}
Here we display the quantitative results on other datasets.
\begin{table}[htbp]
  \centering
  \caption{Quantitative comparisons on ECSSD, DUT-OMRON, and HKU-IS. The better results are shown with \textbf{bold}, and darker color indicates superior results. Metrics with $\uparrow$ mean higher value represents better performance, while $\downarrow$ mean lower value represents better performance.}
  \scalebox{0.85}{
    \begin{tabular}{c|l|ccccccccc}
    \toprule
    \multicolumn{1}{c|}{Dataset} & Methods & $\MAE \downarrow $ & $\SMAE \downarrow$ & $\AUC \uparrow $ &$\SAUC \uparrow $ & $\F_m^\beta \uparrow$ & $\SF_m^{\beta} \uparrow $ & $\F_{max}^\beta \uparrow$ & $\SF_{max}^\beta \uparrow$ & $\E_m \uparrow$  \\ 
\cmidrule{1-1}\cmidrule{2-11}    \multicolumn{1}{c|}{\multirow{10}[9]{*}{ECSSD}} & PoolNet & 0.0632 & 0.0467 & \cellcolor[rgb]{ .867,  .929,  .824} 0.9785 & \cellcolor[rgb]{ .839,  .914,  .788} 0.9817 & 0.8453 & 0.8630 & 0.9205 & 0.9309 & 0.8813 \\
    \multicolumn{1}{c|}{} & \textbf{+ \textit{Ours}} & \cellcolor[rgb]{ .953,  .973,  .937} \textbf{0.0575} & \cellcolor[rgb]{ .945,  .969,  .929} \textbf{0.0421} & \cellcolor[rgb]{ .824,  .906,  .765} \textbf{0.9839} & \cellcolor[rgb]{ .776,  .878,  .706} \textbf{0.9893} & \cellcolor[rgb]{ .957,  .976,  .941} \textbf{0.8588} & \cellcolor[rgb]{ .902,  .949,  .871} \textbf{0.8907} & \cellcolor[rgb]{ .945,  .973,  .929} \textbf{0.9287} & \cellcolor[rgb]{ .851,  .922,  .804} \textbf{0.9477} & \cellcolor[rgb]{ .933,  .965,  .914} \textbf{0.8989} \\
\cmidrule{2-11}    \multicolumn{1}{c|}{} & LDF   & \cellcolor[rgb]{ .851,  .918,  .804} \textbf{0.0450} & \cellcolor[rgb]{ .851,  .918,  .804} \textbf{0.0336} & 0.9618 & \cellcolor[rgb]{ .875,  .933,  .835} 0.9774 & \cellcolor[rgb]{ .831,  .91,  .776} \textbf{0.8949} & \cellcolor[rgb]{ .875,  .933,  .835} 0.8984 & \cellcolor[rgb]{ .89,  .941,  .855} 0.9366 & \cellcolor[rgb]{ .902,  .949,  .871} 0.9421 & \cellcolor[rgb]{ .871,  .929,  .827} 0.9153 \\
    \multicolumn{1}{c|}{} & \textbf{+ \textit{Ours}} & \cellcolor[rgb]{ .871,  .929,  .831} 0.0476 & \cellcolor[rgb]{ .871,  .929,  .827} 0.0353 & \cellcolor[rgb]{ .855,  .922,  .808} \textbf{0.9799} & \cellcolor[rgb]{ .878,  .933,  .839} \textbf{0.9769} & \cellcolor[rgb]{ .839,  .914,  .788} 0.8920 & \cellcolor[rgb]{ .843,  .918,  .792} \textbf{0.9072} & \cellcolor[rgb]{ .867,  .929,  .824} \textbf{0.9404} & \cellcolor[rgb]{ .812,  .898,  .753} \textbf{0.9522} & \cellcolor[rgb]{ .835,  .914,  .784} \textbf{0.9236} \\
\cmidrule{2-11}    \multicolumn{1}{c|}{} & ICON  & \cellcolor[rgb]{ .804,  .894,  .745} \textbf{0.0395} & \cellcolor[rgb]{ .808,  .894,  .749} \textbf{0.0300} & \cellcolor[rgb]{ .953,  .976,  .937} 0.9677 & \cellcolor[rgb]{ .898,  .945,  .867} 0.9746 & \cellcolor[rgb]{ .8,  .89,  .737} \textbf{0.9036} & \cellcolor[rgb]{ .839,  .914,  .788} 0.9084 & \cellcolor[rgb]{ .863,  .925,  .82} \textbf{0.9407} & \cellcolor[rgb]{ .875,  .933,  .835} 0.9452 & \cellcolor[rgb]{ .82,  .902,  .765} 0.9279 \\
    \multicolumn{1}{c|}{} & \textbf{+ \textit{Ours}} & \cellcolor[rgb]{ .839,  .91,  .788} 0.0436 & \cellcolor[rgb]{ .839,  .914,  .792} 0.0328 & \cellcolor[rgb]{ .89,  .941,  .855} \textbf{0.9756} & \cellcolor[rgb]{ .875,  .933,  .835} \textbf{0.9774} & \cellcolor[rgb]{ .847,  .918,  .796} 0.8905 & \cellcolor[rgb]{ .835,  .91,  .78} \textbf{0.9096} & \cellcolor[rgb]{ .91,  .953,  .882} 0.9338 & \cellcolor[rgb]{ .847,  .918,  .8} \textbf{0.9481} & \cellcolor[rgb]{ .82,  .902,  .761} \textbf{0.9281} \\
\cmidrule{2-11}    \multicolumn{1}{c|}{} & GateNet & \cellcolor[rgb]{ .792,  .886,  .725} 0.0378 & \cellcolor[rgb]{ .796,  .886,  .733} 0.0288 & \cellcolor[rgb]{ .875,  .933,  .835} 0.9773 & \cellcolor[rgb]{ .875,  .933,  .835} 0.9773 & \cellcolor[rgb]{ .776,  .878,  .706} \textbf{0.9098} & \cellcolor[rgb]{ .796,  .89,  .733} 0.9199 & \cellcolor[rgb]{ .824,  .906,  .769} \textbf{0.9463} & \cellcolor[rgb]{ .816,  .898,  .753} \textbf{0.9520} & \cellcolor[rgb]{ .788,  .886,  .722} 0.9358 \\
    \multicolumn{1}{c|}{} & \textbf{+ \textit{Ours}} & \cellcolor[rgb]{ .776,  .878,  .71} \textbf{0.0362} & \cellcolor[rgb]{ .78,  .878,  .71} \textbf{0.0274} & \cellcolor[rgb]{ .867,  .929,  .824} \textbf{0.9784} & \cellcolor[rgb]{ .812,  .898,  .749} \textbf{0.9853} & \cellcolor[rgb]{ .788,  .886,  .718} 0.9075 & \cellcolor[rgb]{ .776,  .878,  .706} \textbf{0.9253} & \cellcolor[rgb]{ .851,  .922,  .804} 0.9423 & \cellcolor[rgb]{ .843,  .914,  .792} 0.9489 & \cellcolor[rgb]{ .776,  .878,  .706} \textbf{0.9387} \\
\cmidrule{2-11}    \multicolumn{1}{c|}{} & EDN   & \cellcolor[rgb]{ .78,  .878,  .71} 0.0363 & \cellcolor[rgb]{ .776,  .878,  .706} 0.0271 & \cellcolor[rgb]{ .882,  .937,  .843} \textbf{0.9767} & \cellcolor[rgb]{ .89,  .941,  .855} \textbf{0.9754} & \cellcolor[rgb]{ .78,  .882,  .714} \textbf{0.9089} & \cellcolor[rgb]{ .816,  .902,  .757} 0.9147 & \cellcolor[rgb]{ .776,  .878,  .706} \textbf{0.9531} & \cellcolor[rgb]{ .776,  .878,  .706} \textbf{0.9560} & \cellcolor[rgb]{ .796,  .89,  .733} 0.9338 \\
    \multicolumn{1}{c|}{} & \textbf{+ \textit{Ours}} & \cellcolor[rgb]{ .776,  .878,  .706} \textbf{0.0358} & \cellcolor[rgb]{ .776,  .878,  .706} \textbf{0.0269} & \cellcolor[rgb]{ .886,  .937,  .847} 0.9762 & \cellcolor[rgb]{ .871,  .929,  .831} 0.9778 & \cellcolor[rgb]{ .784,  .882,  .714} 0.9084 & \cellcolor[rgb]{ .792,  .886,  .725} \textbf{0.9216} & \cellcolor[rgb]{ .831,  .91,  .776} 0.9456 & \cellcolor[rgb]{ .792,  .89,  .729} 0.9543 & \cellcolor[rgb]{ .784,  .882,  .714} \textbf{0.9375} \\
\midrule
    \multicolumn{1}{c|}{\multirow{10}[9]{*}{DUT-OMRON}} & PoolNet & 0.0727 & 0.0625 & \cellcolor[rgb]{ 1,  .937,  .733} 0.9403 & \cellcolor[rgb]{ 1,  .914,  .647} 0.9595 & 0.6757 & \cellcolor[rgb]{ 1,  .996,  .984} 0.7463 & \cellcolor[rgb]{ 1,  .988,  .941} 0.7919 & \cellcolor[rgb]{ 1,  .941,  .757} 0.8738 & 0.7807 \\
    \multicolumn{1}{c|}{} & \textbf{+ \textit{Ours}} & \cellcolor[rgb]{ 1,  .984,  .937} \textbf{0.0699} & \cellcolor[rgb]{ 1,  .973,  .898} \textbf{0.0585} & \cellcolor[rgb]{ 1,  .929,  .71} \textbf{0.9457} & \cellcolor[rgb]{ 1,  .902,  .6} \textbf{0.9687} & \cellcolor[rgb]{ 1,  .973,  .89} \textbf{0.6979} & \cellcolor[rgb]{ 1,  .957,  .824} \textbf{0.7847} & \cellcolor[rgb]{ 1,  .929,  .698} \textbf{0.8093} & \cellcolor[rgb]{ 1,  .906,  .616} \textbf{0.8927} & \cellcolor[rgb]{ 1,  .961,  .843} \textbf{0.8123} \\
\cmidrule{2-11}    \multicolumn{1}{c|}{} & LDF   & \cellcolor[rgb]{ 1,  .902,  .6} \textbf{0.0540} & \cellcolor[rgb]{ 1,  .902,  .6} \textbf{0.0464} & 0.8834 & \cellcolor[rgb]{ 1,  .929,  .714} 0.9452 & \cellcolor[rgb]{ 1,  .953,  .8} 0.7154 & 0.7416 & \cellcolor[rgb]{ 1,  .988,  .941} 0.7918 & 0.8413 & \cellcolor[rgb]{ 1,  .969,  .863} 0.8080 \\
    \multicolumn{1}{c|}{} & \textbf{+ \textit{Ours}} & \cellcolor[rgb]{ 1,  .922,  .694} 0.0584 & \cellcolor[rgb]{ 1,  .922,  .686} 0.0499 & \cellcolor[rgb]{ 1,  .945,  .773} \textbf{0.9327} & \cellcolor[rgb]{ 1,  .922,  .675} \textbf{0.9535} & \cellcolor[rgb]{ 1,  .953,  .8} \textbf{0.7156} & \cellcolor[rgb]{ 1,  .961,  .839} \textbf{0.7813} & \cellcolor[rgb]{ 1,  .965,  .851} \textbf{0.7982} & \cellcolor[rgb]{ 1,  .902,  .6} \textbf{0.8944} & \cellcolor[rgb]{ 1,  .941,  .753} \textbf{0.8299} \\
\cmidrule{2-11}    \multicolumn{1}{c|}{} & ICON  & \cellcolor[rgb]{ 1,  .933,  .729} \textbf{0.0601} & \cellcolor[rgb]{ 1,  .937,  .749} \textbf{0.0525} & \cellcolor[rgb]{ 1,  .965,  .851} 0.9160 & \cellcolor[rgb]{ 1,  .949,  .784} 0.9296 & \cellcolor[rgb]{ 1,  .914,  .651} \textbf{0.7450} & \cellcolor[rgb]{ 1,  .949,  .784} 0.7940 & \cellcolor[rgb]{ 1,  .929,  .702} \textbf{0.8092} & \cellcolor[rgb]{ 1,  .969,  .875} 0.8584 & \cellcolor[rgb]{ 1,  .922,  .671} \textbf{0.8456} \\
    \multicolumn{1}{c|}{} & \textbf{+ \textit{Ours}} & \cellcolor[rgb]{ 1,  .957,  .824} 0.0646 & \cellcolor[rgb]{ 1,  .953,  .824} 0.0554 & \cellcolor[rgb]{ 1,  .941,  .757} \textbf{0.9358} & \cellcolor[rgb]{ 1,  .933,  .725} \textbf{0.9423} & \cellcolor[rgb]{ 1,  .961,  .827} 0.7098 & \cellcolor[rgb]{ 1,  .933,  .714} \textbf{0.8108} & 0.7874 & \cellcolor[rgb]{ 1,  .929,  .714} \textbf{0.8798} & \cellcolor[rgb]{ 1,  .937,  .729} 0.8342 \\
\cmidrule{2-11}    \multicolumn{1}{c|}{} & GateNet & \cellcolor[rgb]{ 1,  .906,  .616} \textbf{0.0548} & \cellcolor[rgb]{ 1,  .906,  .624} \textbf{0.0475} & \cellcolor[rgb]{ 1,  .953,  .808} 0.9246 & \cellcolor[rgb]{ 1,  .953,  .808} 0.9246 & \cellcolor[rgb]{ 1,  .922,  .675} \textbf{0.7403} & \cellcolor[rgb]{ 1,  .941,  .745} 0.8031 & \cellcolor[rgb]{ 1,  .922,  .667} \textbf{0.8116} & \cellcolor[rgb]{ 1,  .949,  .784} 0.8704 & \cellcolor[rgb]{ 1,  .922,  .667} 0.8464 \\
    \multicolumn{1}{c|}{} & \textbf{+ \textit{Ours}} & \cellcolor[rgb]{ 1,  .922,  .682} 0.0580 & \cellcolor[rgb]{ 1,  .922,  .69} 0.0501 & \cellcolor[rgb]{ 1,  .945,  .769} \textbf{0.9329} & \cellcolor[rgb]{ 1,  .949,  .788} \textbf{0.9293} & \cellcolor[rgb]{ 1,  .925,  .694} 0.7363 & \cellcolor[rgb]{ 1,  .914,  .635} \textbf{0.8300} & \cellcolor[rgb]{ 1,  .941,  .761} 0.8049 & \cellcolor[rgb]{ 1,  .922,  .671} \textbf{0.8852} & \cellcolor[rgb]{ 1,  .914,  .643} \textbf{0.8510} \\
\cmidrule{2-11}    \multicolumn{1}{c|}{} & EDN   & \cellcolor[rgb]{ 1,  .906,  .624} \textbf{0.0551} & \cellcolor[rgb]{ 1,  .914,  .647} 0.0484 & \cellcolor[rgb]{ 1,  .949,  .788} 0.9292 & \cellcolor[rgb]{ 1,  .937,  .733} 0.9407 & \cellcolor[rgb]{ 1,  .906,  .608} \textbf{0.7529} & \cellcolor[rgb]{ 1,  .922,  .667} 0.8224 & \cellcolor[rgb]{ 1,  .918,  .667} \textbf{0.8117} & \cellcolor[rgb]{ 1,  .929,  .714} 0.8798 & s \\
    \multicolumn{1}{c|}{} & \textbf{+ \textit{Ours}} & \cellcolor[rgb]{ 1,  .91,  .635} 0.0557 & \cellcolor[rgb]{ 1,  .91,  .647} \textbf{0.0483} & \cellcolor[rgb]{ 1,  .937,  .745} \textbf{0.9382} & \cellcolor[rgb]{ 1,  .925,  .686} \textbf{0.9507} & \cellcolor[rgb]{ 1,  .902,  .6} 0.7544 & \cellcolor[rgb]{ 1,  .902,  .6} \textbf{0.8381} & \cellcolor[rgb]{ 1,  .902,  .6} 0.8163 & \cellcolor[rgb]{ 1,  .91,  .627} \textbf{0.8912} & \cellcolor[rgb]{ 1,  .902,  .6} \textbf{0.8594} \\
    \midrule
    \multirow{10}[10]{*}{HKU-IS} & PoolNet & 0.0526 & 0.0537 & \cellcolor[rgb]{ .969,  .902,  .906} 0.9804 & \cellcolor[rgb]{ .965,  .886,  .89} 0.9842 & 0.8294 & 0.8316 & 0.9066 & \cellcolor[rgb]{ .996,  .98,  .98} 0.9199 & 0.8816 \\
          & \textbf{+ \textit{Ours}} & \cellcolor[rgb]{ .988,  .965,  .965} \textbf{0.0464} & \cellcolor[rgb]{ .98,  .945,  .949} \textbf{0.0440} & \cellcolor[rgb]{ .965,  .886,  .89} \textbf{0.9847} & \cellcolor[rgb]{ .957,  .867,  .871} \textbf{0.9891} & \cellcolor[rgb]{ .988,  .965,  .965} \textbf{0.8501} & \cellcolor[rgb]{ .976,  .925,  .929} \textbf{0.8770} & \cellcolor[rgb]{ .988,  .957,  .957} \textbf{0.9188} & \cellcolor[rgb]{ .965,  .89,  .894} \textbf{0.9395} & \cellcolor[rgb]{ .984,  .945,  .949} \textbf{0.9081} \\
\cmidrule{2-11}          & LDF   & \cellcolor[rgb]{ .965,  .894,  .898} \textbf{0.0333} & \cellcolor[rgb]{ .965,  .902,  .902} 0.0355 & \cellcolor[rgb]{ 1,  .988,  .992} 0.9575 & 0.9544 & \cellcolor[rgb]{ .969,  .894,  .898} \textbf{0.8868} & \cellcolor[rgb]{ .976,  .929,  .929} 0.8759 & \cellcolor[rgb]{ .976,  .925,  .929} 0.9263 & \cellcolor[rgb]{ .992,  .973,  .973} 0.9217 & \cellcolor[rgb]{ .973,  .914,  .918} 0.9234 \\
          & \textbf{+ \textit{Ours}} & \cellcolor[rgb]{ .965,  .902,  .902} 0.0346 & \cellcolor[rgb]{ .965,  .894,  .898} \textbf{0.0344} & \cellcolor[rgb]{ .969,  .898,  .902} \textbf{0.9810} & \cellcolor[rgb]{ .965,  .89,  .89} \textbf{0.9839} & \cellcolor[rgb]{ .969,  .906,  .906} 0.8815 & \cellcolor[rgb]{ .969,  .902,  .906} \textbf{0.8910} & \cellcolor[rgb]{ .973,  .91,  .914} \textbf{0.9306} & \cellcolor[rgb]{ .961,  .871,  .875} \textbf{0.9437} & \cellcolor[rgb]{ .969,  .894,  .898} \textbf{0.9320} \\
\cmidrule{2-11}          & ICON  & \cellcolor[rgb]{ .965,  .902,  .902} \textbf{0.0346} & \cellcolor[rgb]{ .969,  .91,  .914} 0.0374 & \cellcolor[rgb]{ .992,  .973,  .976} 0.9616 & \cellcolor[rgb]{ .988,  .965,  .965} 0.9641 & \cellcolor[rgb]{ .969,  .898,  .898} \textbf{0.8854} & \cellcolor[rgb]{ .98,  .933,  .937} 0.8722 & \cellcolor[rgb]{ .98,  .937,  .941} 0.9232 & 0.9151 & \cellcolor[rgb]{ .969,  .906,  .906} 0.9277 \\
          & \textbf{+ \textit{Ours}} & \cellcolor[rgb]{ .969,  .906,  .91} 0.0357 & \cellcolor[rgb]{ .969,  .902,  .906} \textbf{0.0361} & \cellcolor[rgb]{ .969,  .898,  .902} \textbf{0.9815} & \cellcolor[rgb]{ .973,  .91,  .91} \textbf{0.9787} & \cellcolor[rgb]{ .973,  .91,  .914} 0.8788 & \cellcolor[rgb]{ .973,  .906,  .91} \textbf{0.8898} & \cellcolor[rgb]{ .976,  .929,  .929} \textbf{0.9260} & \cellcolor[rgb]{ .969,  .902,  .906} \textbf{0.9369} & \cellcolor[rgb]{ .969,  .894,  .898} \textbf{0.9315} \\
\cmidrule{2-11}          & GateNet & \cellcolor[rgb]{ .965,  .89,  .894} 0.0326 & \cellcolor[rgb]{ .965,  .89,  .894} 0.0338 & \cellcolor[rgb]{ .976,  .918,  .922} 0.9762 & 0.9830 & \cellcolor[rgb]{ .961,  .878,  .878} 0.8961 & \cellcolor[rgb]{ .965,  .89,  .894} 0.8993 & \cellcolor[rgb]{ .969,  .894,  .898} \textbf{0.9346} & \cellcolor[rgb]{ .969,  .894,  .898} 0.9385 & \cellcolor[rgb]{ .961,  .878,  .882} 0.9394 \\
          & \textbf{+ \textit{Ours}} & \cellcolor[rgb]{ .957,  .871,  .875} \textbf{0.0292} & \cellcolor[rgb]{ .957,  .867,  .871} \textbf{0.0293} & \cellcolor[rgb]{ .973,  .91,  .914} \textbf{0.9785} & \cellcolor[rgb]{ .973,  .91,  .914} 0.9779 & \cellcolor[rgb]{ .961,  .871,  .875} \textbf{0.8995} & \cellcolor[rgb]{ .957,  .867,  .871} \textbf{0.9122} & \cellcolor[rgb]{ .969,  .894,  .898} 0.9345 & \cellcolor[rgb]{ .961,  .871,  .875} \textbf{0.9441} & \cellcolor[rgb]{ .957,  .867,  .871} \textbf{0.9443} \\
\cmidrule{2-5}\cmidrule{7-11}          & EDN   & \cellcolor[rgb]{ .957,  .867,  .871} \textbf{0.0279} & \cellcolor[rgb]{ .957,  .867,  .871} 0.0294 & \cellcolor[rgb]{ .976,  .922,  .925} 0.9750 & \cellcolor[rgb]{ .976,  .925,  .929} 0.9738 & \cellcolor[rgb]{ .957,  .867,  .871} \textbf{0.9004} & \cellcolor[rgb]{ .965,  .886,  .89} 0.9017 & \cellcolor[rgb]{ .957,  .867,  .871} \textbf{0.9417} & \cellcolor[rgb]{ .969,  .906,  .906} 0.9364 & \cellcolor[rgb]{ .961,  .871,  .875} 0.9429 \\
          & \textbf{+ \textit{Ours}} & \cellcolor[rgb]{ .957,  .871,  .875} 0.0287 & \cellcolor[rgb]{ .957,  .867,  .871} \textbf{0.0289} & \cellcolor[rgb]{ .973,  .914,  .914} \textbf{0.9776} & \cellcolor[rgb]{ .973,  .91,  .914} \textbf{0.9780} & \cellcolor[rgb]{ .961,  .871,  .875} 0.8986 & \cellcolor[rgb]{ .961,  .878,  .882} \textbf{0.9072} & \cellcolor[rgb]{ .965,  .886,  .886} 0.9375 & \cellcolor[rgb]{ .957,  .867,  .871} \textbf{0.9443} & \cellcolor[rgb]{ .961,  .871,  .875} \textbf{0.9442} \\
    \bottomrule
    \end{tabular}
    }
  \label{tab:exp_result_appendix}
\end{table}%
\FloatBarrier
\newpage

\begin{table}[htbp]
  \centering
  \caption{Quantitative comparisons on SOD, PASCAL-S, and XPIE. The better results are shown with \textbf{bold}, and darker color indicates superior results. Metrics with $\uparrow$ mean higher value represents better performance, while $\downarrow$ mean lower value represents better performance.}
  \scalebox{0.85}{
    \begin{tabular}{c|l|ccccccccc}
    \toprule
    \multicolumn{1}{c|}{Dataset} & Methods & $\MAE \downarrow $ & $\SMAE \downarrow$ & $\AUC \uparrow $ &$\SAUC \uparrow $ & $\F_m^\beta \uparrow$ & $\SF_m^{\beta} \uparrow $ & $\F_{max}^\beta \uparrow$ & $\SF_{max}^\beta \uparrow$ & $\E_m \uparrow$  \\ 
\midrule
    \multirow{10}[8]{*}{SOD} & PoolNet & 0.1353 & 0.1219 & \cellcolor[rgb]{ .984,  .886,  .816} 0.9000 & \cellcolor[rgb]{ .984,  .871,  .796} 0.9039 & 0.6974 & 0.6685 & 0.8356 & \cellcolor[rgb]{ .992,  .925,  .882} 0.8370 & 0.7208 \\
          & +Ours & \cellcolor[rgb]{ .988,  .941,  .906} \textbf{0.1235} & \cellcolor[rgb]{ .988,  .925,  .882} \textbf{0.1097} & \cellcolor[rgb]{ .98,  .835,  .741} \textbf{0.9143} & \cellcolor[rgb]{ .973,  .796,  .678} \textbf{0.9251} & \cellcolor[rgb]{ .992,  .933,  .89} \textbf{0.7365} & \cellcolor[rgb]{ .988,  .898,  .839} \textbf{0.7245} & \cellcolor[rgb]{ .996,  .957,  .929} \textbf{0.8437} & \cellcolor[rgb]{ .976,  .808,  .698} \textbf{0.8657} & \cellcolor[rgb]{ .988,  .914,  .859} \textbf{0.7654} \\
\cmidrule{2-11}          & LDF   & \cellcolor[rgb]{ .973,  .796,  .678} \textbf{0.0940} & \cellcolor[rgb]{ .973,  .796,  .678} \textbf{0.0884} & \cellcolor[rgb]{ .988,  .906,  .855} 0.8934 & \cellcolor[rgb]{ .992,  .941,  .906} 0.8839 & \cellcolor[rgb]{ .976,  .82,  .714} \textbf{0.7991} & \cellcolor[rgb]{ .98,  .835,  .741} \textbf{0.7584} & \cellcolor[rgb]{ .976,  .808,  .694} \textbf{0.8706} & \cellcolor[rgb]{ .992,  .918,  .867} 0.8391 & \cellcolor[rgb]{ .976,  .824,  .722} \textbf{0.8091} \\
          & +Ours & \cellcolor[rgb]{ .98,  .871,  .796} 0.1093 & \cellcolor[rgb]{ .98,  .871,  .796} 0.1009 & \cellcolor[rgb]{ .984,  .867,  .784} \textbf{0.9058} & \cellcolor[rgb]{ .98,  .851,  .761} \textbf{0.9102} & \cellcolor[rgb]{ .984,  .863,  .78} 0.7755 & \cellcolor[rgb]{ .984,  .871,  .792} 0.7404 & \cellcolor[rgb]{ .984,  .871,  .796} 0.8593 & \cellcolor[rgb]{ .973,  .796,  .678} \textbf{0.8684} & \cellcolor[rgb]{ .984,  .871,  .796} 0.7858 \\
\cmidrule{2-11}          & ICON  & \cellcolor[rgb]{ .98,  .851,  .769} 0.1058 & \cellcolor[rgb]{ .98,  .863,  .784} 0.0995 & \cellcolor[rgb]{ .996,  .961,  .933} 0.8785 & 0.8661 & \cellcolor[rgb]{ .98,  .831,  .729} 0.7931 & \cellcolor[rgb]{ .984,  .859,  .776} 0.7461 & \cellcolor[rgb]{ .984,  .886,  .816} \textbf{0.8567} & 0.8182 & \cellcolor[rgb]{ .98,  .847,  .757} 0.7984 \\
          & +Ours & \cellcolor[rgb]{ .973,  .816,  .714} \textbf{0.0987} & \cellcolor[rgb]{ .973,  .816,  .71} \textbf{0.0920} & \cellcolor[rgb]{ .98,  .855,  .773} \textbf{0.9083} & \cellcolor[rgb]{ .98,  .855,  .769} \textbf{0.9088} & \cellcolor[rgb]{ .976,  .82,  .714} \textbf{0.7992} & \cellcolor[rgb]{ .976,  .804,  .69} \textbf{0.7759} & \cellcolor[rgb]{ .988,  .89,  .827} 0.8557 & \cellcolor[rgb]{ .984,  .867,  .788} \textbf{0.8513} & \cellcolor[rgb]{ .973,  .796,  .678} \textbf{0.8220} \\
\cmidrule{2-11}          & GateNet & \cellcolor[rgb]{ .973,  .816,  .714} 0.0987 & \cellcolor[rgb]{ .976,  .835,  .737} 0.0949 & \cellcolor[rgb]{ .992,  .941,  .906} 0.8834 & \cellcolor[rgb]{ 1,  .98,  .965} 0.8727 & \cellcolor[rgb]{ .98,  .831,  .733} 0.7928 & \cellcolor[rgb]{ .98,  .851,  .761} 0.7514 & \cellcolor[rgb]{ .984,  .867,  .788} 0.8602 & \cellcolor[rgb]{ .996,  .965,  .941} 0.8274 & \cellcolor[rgb]{ .98,  .851,  .765} 0.7949 \\
          & +Ours & \cellcolor[rgb]{ .973,  .808,  .694} \textbf{0.0965} & \cellcolor[rgb]{ .973,  .812,  .702} \textbf{0.0910} & \cellcolor[rgb]{ .988,  .906,  .851} \textbf{0.8935} & \cellcolor[rgb]{ .992,  .925,  .882} \textbf{0.8880} & \cellcolor[rgb]{ .976,  .8,  .686} \textbf{0.8094} & \cellcolor[rgb]{ .973,  .796,  .678} \textbf{0.7792} & \cellcolor[rgb]{ .98,  .851,  .765} \textbf{0.8626} & \cellcolor[rgb]{ .984,  .878,  .808} \textbf{0.8483} & \cellcolor[rgb]{ .976,  .804,  .69} \textbf{0.8190} \\
\cmidrule{2-11}          & EDN   & \cellcolor[rgb]{ .98,  .871,  .796} 0.1093 & \cellcolor[rgb]{ .98,  .871,  .796} 0.1009 & \cellcolor[rgb]{ .996,  .945,  .914} 0.8823 & \cellcolor[rgb]{ 1,  .984,  .973} 0.8716 & \cellcolor[rgb]{ .98,  .855,  .769} 0.7795 & \cellcolor[rgb]{ .988,  .886,  .82} 0.7313 & \cellcolor[rgb]{ .973,  .796,  .678} \textbf{0.8723} & \cellcolor[rgb]{ .992,  .922,  .875} 0.8379 & \cellcolor[rgb]{ .984,  .867,  .792} 0.7873 \\
          & +Ours & \cellcolor[rgb]{ .973,  .816,  .71} \textbf{0.0982} & \cellcolor[rgb]{ .973,  .816,  .714} \textbf{0.0922} & \cellcolor[rgb]{ .992,  .922,  .875} \textbf{0.8892} & \cellcolor[rgb]{ .996,  .949,  .918} \textbf{0.8818} & \cellcolor[rgb]{ .973,  .796,  .678} \textbf{0.8110} & \cellcolor[rgb]{ .976,  .812,  .698} \textbf{0.7728} & \cellcolor[rgb]{ .976,  .824,  .722} 0.8677 & \cellcolor[rgb]{ .992,  .922,  .875} \textbf{0.8382} & \cellcolor[rgb]{ .976,  .8,  .686} \textbf{0.8207} \\
\midrule
    \multirow{10}[8]{*}{PASCAL-S} & PoolNet & 0.0944 & 0.0716 & \cellcolor[rgb]{ .827,  .894,  .957} 0.9462 & \cellcolor[rgb]{ .827,  .898,  .957} 0.9612 & 0.7556 & \cellcolor[rgb]{ .933,  .961,  .984} 0.8495 & 0.7932 & \cellcolor[rgb]{ .996,  .996,  1} 0.8915 & 0.8165 \\
          & +Ours & \cellcolor[rgb]{ .976,  .984,  .992} \textbf{0.0919} & \cellcolor[rgb]{ .969,  .98,  .992} \textbf{0.0690} & \cellcolor[rgb]{ .741,  .843,  .933} \textbf{0.9535} & \cellcolor[rgb]{ .741,  .843,  .933} \textbf{0.9722} & \cellcolor[rgb]{ .969,  .98,  .992} \textbf{0.7651} & \cellcolor[rgb]{ .918,  .949,  .98} \textbf{0.8520} & \cellcolor[rgb]{ .882,  .929,  .973} \textbf{0.8346} & \cellcolor[rgb]{ .792,  .875,  .949} \textbf{0.9103} & \cellcolor[rgb]{ .933,  .961,  .984} \textbf{0.8359} \\
\cmidrule{2-11}          & LDF   & \cellcolor[rgb]{ .773,  .859,  .941} \textbf{0.0662} & \cellcolor[rgb]{ .769,  .859,  .941} \textbf{0.0502} & \cellcolor[rgb]{ .855,  .914,  .965} 0.9437 & \cellcolor[rgb]{ .89,  .933,  .973} 0.9536 & \cellcolor[rgb]{ .808,  .882,  .953} \textbf{0.8117} & \cellcolor[rgb]{ .929,  .957,  .984} 0.8504 & \cellcolor[rgb]{ .765,  .859,  .941} \textbf{0.8759} & \cellcolor[rgb]{ .792,  .875,  .949} 0.9106 & \cellcolor[rgb]{ .792,  .875,  .949} \textbf{0.8742} \\
          & +Ours & \cellcolor[rgb]{ .808,  .882,  .949} 0.0705 & \cellcolor[rgb]{ .804,  .878,  .949} 0.0532 & \cellcolor[rgb]{ .792,  .875,  .949} \textbf{0.9492} & \cellcolor[rgb]{ .82,  .89,  .953} \textbf{0.9626} & \cellcolor[rgb]{ .82,  .894,  .957} 0.8075 & \cellcolor[rgb]{ .925,  .957,  .98} \textbf{0.8509} & \cellcolor[rgb]{ .78,  .867,  .945} 0.8710 & \cellcolor[rgb]{ .741,  .843,  .933} \textbf{0.9150} & \cellcolor[rgb]{ .804,  .882,  .949} 0.8716 \\
\cmidrule{2-11}          & ICON  & \cellcolor[rgb]{ .831,  .898,  .953} \textbf{0.0735} & \cellcolor[rgb]{ .835,  .902,  .957} \textbf{0.0565} & 0.9308 & 0.9391 & \cellcolor[rgb]{ .796,  .878,  .949} \textbf{0.8142} & 0.8406 & \cellcolor[rgb]{ .8,  .878,  .949} \textbf{0.8642} & 0.8908 & \cellcolor[rgb]{ .816,  .89,  .953} \textbf{0.8682} \\
          & +Ours & \cellcolor[rgb]{ .875,  .922,  .965} 0.0791 & \cellcolor[rgb]{ .875,  .922,  .965} 0.0599 & \cellcolor[rgb]{ .831,  .898,  .957} \textbf{0.9456} & \cellcolor[rgb]{ .835,  .902,  .961} \textbf{0.9603} & \cellcolor[rgb]{ .867,  .918,  .969} 0.7946 & \cellcolor[rgb]{ .914,  .949,  .98} \textbf{0.8521} & \cellcolor[rgb]{ .824,  .894,  .957} 0.8554 & \cellcolor[rgb]{ .875,  .925,  .969} \textbf{0.9028} & \cellcolor[rgb]{ .831,  .898,  .957} 0.8641 \\
\cmidrule{2-11}          & GateNet & \cellcolor[rgb]{ .741,  .843,  .933} \textbf{0.0622} & \cellcolor[rgb]{ .741,  .843,  .933} \textbf{0.0473} & \cellcolor[rgb]{ .812,  .886,  .953} 0.9474 & \cellcolor[rgb]{ .875,  .925,  .969} 0.9554 & \cellcolor[rgb]{ .741,  .843,  .933} \textbf{0.8298} & \cellcolor[rgb]{ .776,  .863,  .945} 0.8707 & \cellcolor[rgb]{ .757,  .855,  .937} \textbf{0.8794} & \cellcolor[rgb]{ .773,  .863,  .945} 0.9121 & \cellcolor[rgb]{ .741,  .843,  .933} \textbf{0.8882} \\
          & +Ours & \cellcolor[rgb]{ .773,  .863,  .941} 0.0665 & \cellcolor[rgb]{ .773,  .863,  .941} 0.0504 & \cellcolor[rgb]{ .808,  .886,  .953} \textbf{0.9478} & \cellcolor[rgb]{ .851,  .91,  .965} \textbf{0.9582} & \cellcolor[rgb]{ .773,  .863,  .941} 0.8217 & \cellcolor[rgb]{ .741,  .843,  .933} 0.8750 & \cellcolor[rgb]{ .776,  .867,  .945} 0.8724 & \cellcolor[rgb]{ .757,  .851,  .937} \textbf{0.9138} & \cellcolor[rgb]{ .757,  .855,  .941} 0.8839 \\
\cmidrule{2-11}          & EDN   & \cellcolor[rgb]{ .761,  .855,  .937} 0.0649 & \cellcolor[rgb]{ .761,  .855,  .937} 0.0494 & \cellcolor[rgb]{ .875,  .925,  .969} 0.9419 & \cellcolor[rgb]{ .914,  .949,  .98} 0.9506 & \cellcolor[rgb]{ .776,  .863,  .945} 0.8207 & \cellcolor[rgb]{ .984,  .992,  .996} 0.8431 & \cellcolor[rgb]{ .741,  .843,  .933} \textbf{0.8841} & \cellcolor[rgb]{ .812,  .886,  .953} 0.9086 & \cellcolor[rgb]{ .792,  .875,  .949} 0.8750 \\
          & +Ours & \cellcolor[rgb]{ .757,  .851,  .937} \textbf{0.0644} & \cellcolor[rgb]{ .757,  .851,  .937} \textbf{0.0491} & \cellcolor[rgb]{ .831,  .898,  .957} \textbf{0.9456} & \cellcolor[rgb]{ .878,  .925,  .969} \textbf{0.9551} & \cellcolor[rgb]{ .757,  .855,  .937} \textbf{0.8260} & \cellcolor[rgb]{ .792,  .875,  .949} \textbf{0.8684} & \cellcolor[rgb]{ .769,  .859,  .941} 0.8757 & \cellcolor[rgb]{ .78,  .867,  .945} \textbf{0.9114} & \cellcolor[rgb]{ .753,  .851,  .937} \textbf{0.8859} \\
\midrule
    \multirow{10}[9]{*}{XPIE} & PoolNet & 0.0622 & 0.0505 & \cellcolor[rgb]{ .835,  .91,  .78} 0.9667 & \cellcolor[rgb]{ .831,  .91,  .78} 0.9771 & 0.7904 & 0.8242 & 0.8710 & \cellcolor[rgb]{ .992,  .996,  .988} 0.9103 & 0.8494 \\
          & +Ours & \cellcolor[rgb]{ .973,  .984,  .965} \textbf{0.0599} & \cellcolor[rgb]{ .961,  .976,  .949} \textbf{0.0476} & \cellcolor[rgb]{ .776,  .878,  .706} \textbf{0.9733} & \cellcolor[rgb]{ .776,  .878,  .706} \textbf{0.9857} & \cellcolor[rgb]{ .957,  .976,  .945} \textbf{0.8042} & \cellcolor[rgb]{ .886,  .937,  .847} \textbf{0.8662} & \cellcolor[rgb]{ .953,  .973,  .933} \textbf{0.8786} & \cellcolor[rgb]{ .82,  .902,  .765} \textbf{0.9308} & \cellcolor[rgb]{ .918,  .957,  .89} \textbf{0.8738} \\
\cmidrule{2-11}          & LDF   & \cellcolor[rgb]{ .796,  .886,  .729} \textbf{0.0428} & \cellcolor[rgb]{ .788,  .882,  .722} 0.0347 & \cellcolor[rgb]{ .855,  .922,  .812} 0.9641 & \cellcolor[rgb]{ .878,  .933,  .839} 0.9700 & \cellcolor[rgb]{ .808,  .898,  .745} \textbf{0.8520} & \cellcolor[rgb]{ .835,  .91,  .78} \textbf{0.8844} & \cellcolor[rgb]{ .796,  .89,  .733} \textbf{0.9015} & \cellcolor[rgb]{ .808,  .898,  .749} 0.9324 & \cellcolor[rgb]{ .804,  .894,  .745} \textbf{0.9054} \\
          & +Ours & \cellcolor[rgb]{ .827,  .906,  .773} 0.0458 & \cellcolor[rgb]{ .82,  .902,  .765} \textbf{0.0372} & \cellcolor[rgb]{ .804,  .894,  .745} \textbf{0.9701} & \cellcolor[rgb]{ .824,  .906,  .769} \textbf{0.9785} & \cellcolor[rgb]{ .824,  .906,  .769} 0.8467 & \cellcolor[rgb]{ .839,  .914,  .788} 0.8824 & \cellcolor[rgb]{ .831,  .91,  .776} 0.8965 & \cellcolor[rgb]{ .776,  .878,  .706} \textbf{0.9360} & \cellcolor[rgb]{ .82,  .902,  .761} 0.9013 \\
\cmidrule{2-11}          & ICON  & \cellcolor[rgb]{ .827,  .906,  .773} \textbf{0.0459} & \cellcolor[rgb]{ .831,  .91,  .78} \textbf{0.0381} & 0.9468 & 0.9503 & \cellcolor[rgb]{ .812,  .898,  .749} \textbf{0.8514} & \cellcolor[rgb]{ .875,  .933,  .835} 0.8699 & \cellcolor[rgb]{ .875,  .933,  .831} \textbf{0.8903} & 0.9090 & \cellcolor[rgb]{ .827,  .906,  .773} \textbf{0.8994} \\
          & +Ours & \cellcolor[rgb]{ .867,  .925,  .827} 0.0498 & \cellcolor[rgb]{ .867,  .925,  .824} 0.0405 & \cellcolor[rgb]{ .839,  .914,  .788} \textbf{0.9662} & \cellcolor[rgb]{ .851,  .922,  .804} \textbf{0.9740} & \cellcolor[rgb]{ .859,  .925,  .812} 0.8359 & \cellcolor[rgb]{ .855,  .922,  .808} \textbf{0.8774} & \cellcolor[rgb]{ .91,  .953,  .882} 0.8847 & \cellcolor[rgb]{ .906,  .949,  .878} \textbf{0.9204} & \cellcolor[rgb]{ .827,  .906,  .773} 0.8988 \\
\cmidrule{2-11}          & GateNet & \cellcolor[rgb]{ .78,  .878,  .71} \textbf{0.0414} & \cellcolor[rgb]{ .776,  .878,  .706} \textbf{0.0339} & \cellcolor[rgb]{ .878,  .933,  .839} 0.9615 & \cellcolor[rgb]{ .906,  .949,  .875} 0.9656 & \cellcolor[rgb]{ .776,  .878,  .706} \textbf{0.8614} & \cellcolor[rgb]{ .812,  .898,  .753} 0.8924 & \cellcolor[rgb]{ .792,  .886,  .725} \textbf{0.9024} & \cellcolor[rgb]{ .831,  .91,  .776} 0.9296 & \cellcolor[rgb]{ .788,  .886,  .718} 0.9107 \\
          & +Ours & \cellcolor[rgb]{ .796,  .886,  .733} 0.0429 & \cellcolor[rgb]{ .788,  .882,  .722} 0.0347 & \cellcolor[rgb]{ .851,  .918,  .8} \textbf{0.9649} & \cellcolor[rgb]{ .871,  .929,  .827} \textbf{0.9713} & \cellcolor[rgb]{ .796,  .89,  .729} 0.8563 & \cellcolor[rgb]{ .776,  .878,  .706} \textbf{0.9046} & \cellcolor[rgb]{ .827,  .906,  .773} 0.8969 & \cellcolor[rgb]{ .78,  .882,  .714} \textbf{0.9356} & \cellcolor[rgb]{ .78,  .882,  .71} \textbf{0.9125} \\
\cmidrule{2-11}          & EDN   & \cellcolor[rgb]{ .776,  .878,  .706} \textbf{0.0409} & \cellcolor[rgb]{ .776,  .878,  .706} 0.0337 & \cellcolor[rgb]{ .894,  .941,  .859} 0.9598 & \cellcolor[rgb]{ .914,  .953,  .886} 0.9642 & \cellcolor[rgb]{ .788,  .886,  .722} 0.8584 & \cellcolor[rgb]{ .847,  .918,  .8} 0.8793 & \cellcolor[rgb]{ .776,  .878,  .706} \textbf{0.9044} & \cellcolor[rgb]{ .863,  .925,  .82} 0.9256 & \cellcolor[rgb]{ .808,  .898,  .749} 0.9043 \\
          & +Ours & \cellcolor[rgb]{ .78,  .882,  .714} 0.0416 & \cellcolor[rgb]{ .776,  .878,  .706} \textbf{0.0337} & \cellcolor[rgb]{ .859,  .922,  .812} \textbf{0.9640} & \cellcolor[rgb]{ .875,  .933,  .831} \textbf{0.9707} & \cellcolor[rgb]{ .784,  .886,  .718} \textbf{0.8590} & \cellcolor[rgb]{ .796,  .89,  .729} \textbf{0.8986} & \cellcolor[rgb]{ .835,  .91,  .784} 0.8959 & \cellcolor[rgb]{ .816,  .898,  .753} \textbf{0.9317} & \cellcolor[rgb]{ .776,  .878,  .706} \textbf{0.9132} \\
\bottomrule   \end{tabular}}
  \label{tab:exp_result_appendix_2}%
\end{table}%
 \newpage

\subsection{Qualitative comparisons}  \label{Qualitative_appendix}
Here we present some visualization examples on the effect of our size-invariant loss.
\begin{figure*}[htbp]
    \centering
    \includegraphics[width=\linewidth]{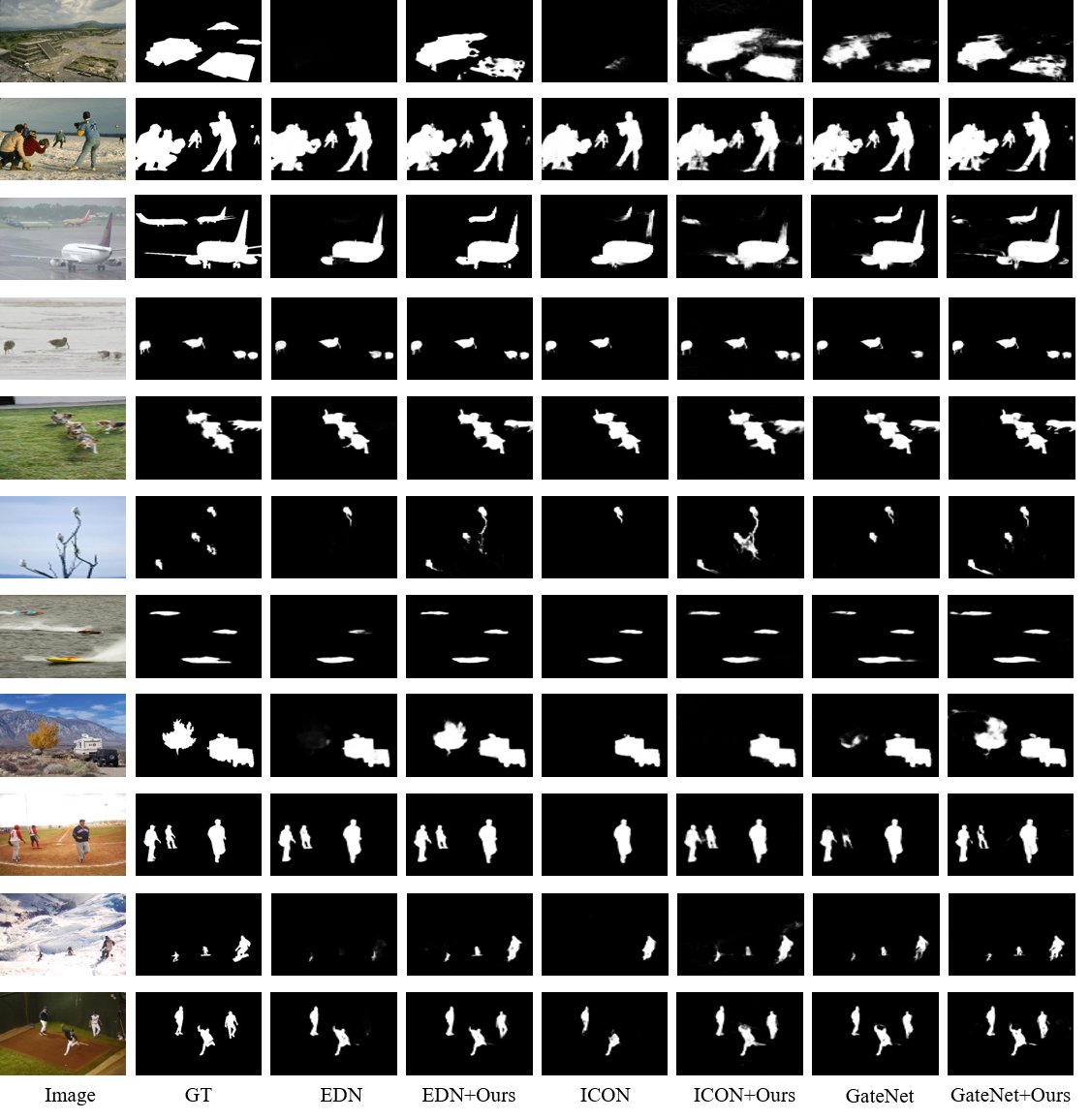}
    \vspace{-0.5cm}
    \caption{Qualitative comparison on different backbones.}
    \label{fig:vis_appendix}
\vspace{-2cm}
\end{figure*}
\FloatBarrier
\newpage
\subsection{Performance with Respect to Sizes} \label{size-fine-grained_appendix}
Here we expand the size-relevant fine-grained analysis to other backbones and benchmarks.
\begin{figure*}[ht]
\centering
\subfigure[MSOD]{   
\begin{minipage}{0.185\linewidth}
\includegraphics[width=\linewidth]{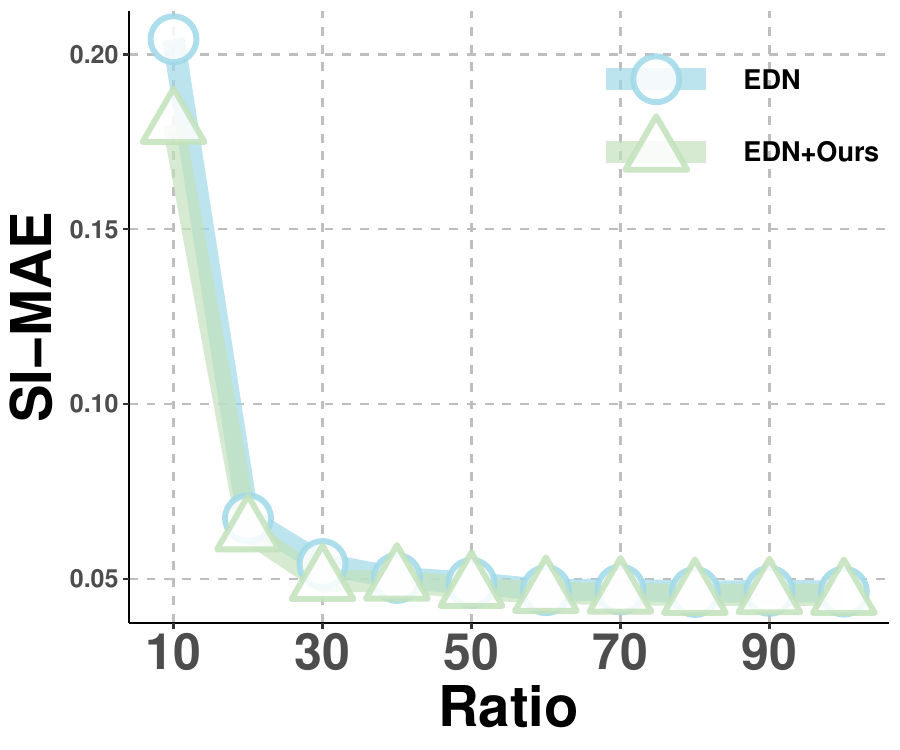}  
\label{fig:EDN_msod_ratio_line-appendix}
\end{minipage}
}
\subfigure[DUTS]{   
\begin{minipage}{0.185\linewidth}
\includegraphics[width=\linewidth]{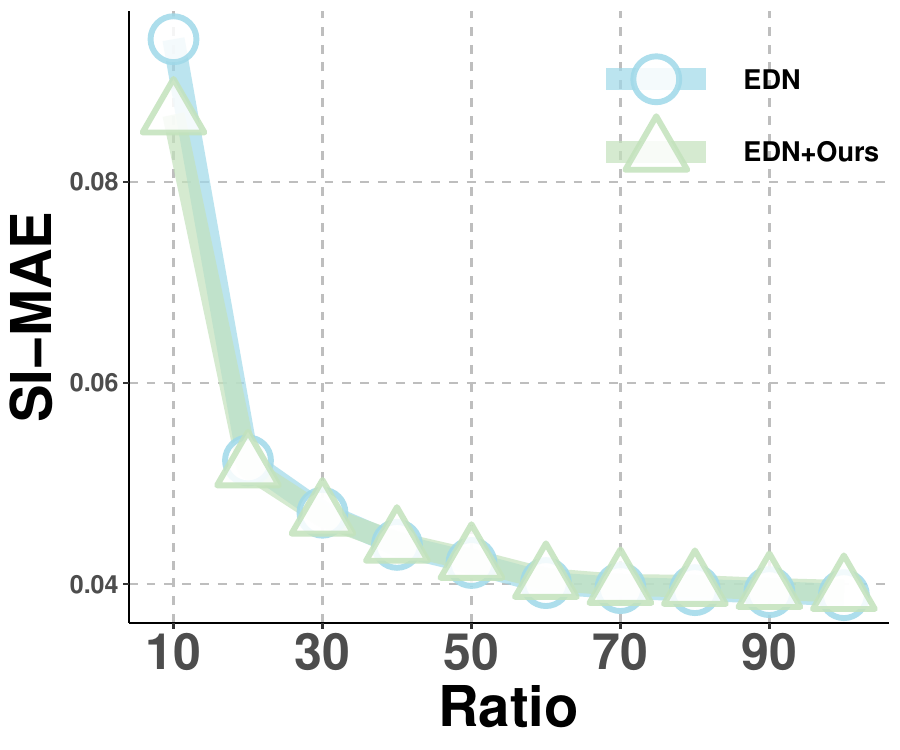}  
\label{fig:EDN_DUTS_ratio_line-appendix}
\end{minipage}
}
\subfigure[ECSSD]{   
\begin{minipage}{0.185\linewidth}
\includegraphics[width=\linewidth]{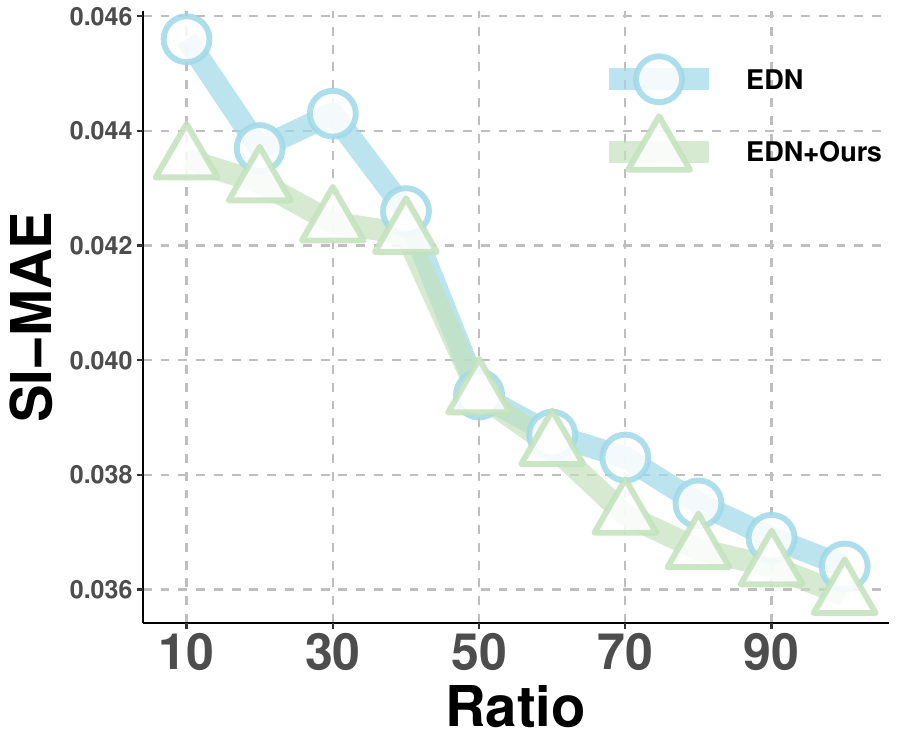}  
\label{fig:EDN_ECSSD_ratio_line}
\end{minipage}
}
\subfigure[DUT-OMRON]{   
\begin{minipage}{0.185\linewidth}
\includegraphics[width=\linewidth]{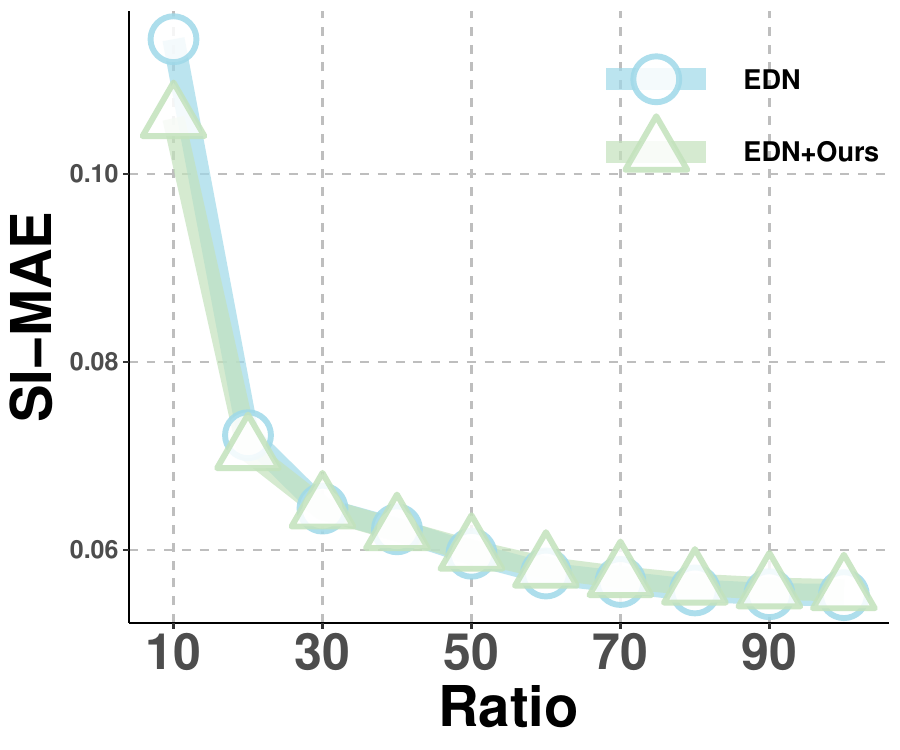}  
\label{fig:EDN_DUT-OMRON_ratio_line}
\end{minipage}
}
\subfigure[HKU-IS]{   
\begin{minipage}{0.185\linewidth}
\includegraphics[width=\linewidth]{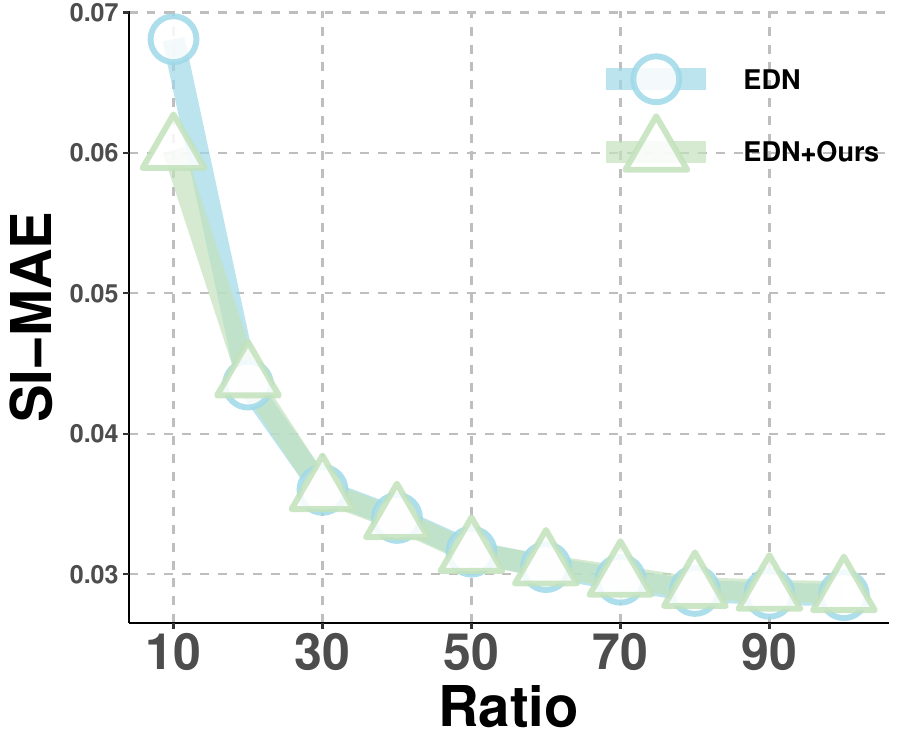}  
\label{fig:EDN_HKU-IS_ratio_line}
\end{minipage}
}
\caption{$\SMAE$ performance on objects with different sizes on five datasets, with EDN as the backbone.}    
\label{fig:fine-analysis-EDN-ratio-appendix}    
\end{figure*}

\begin{figure*}[ht]
\centering
\subfigure[MSOD]{   
\begin{minipage}{0.185\linewidth}
\includegraphics[width=\linewidth]{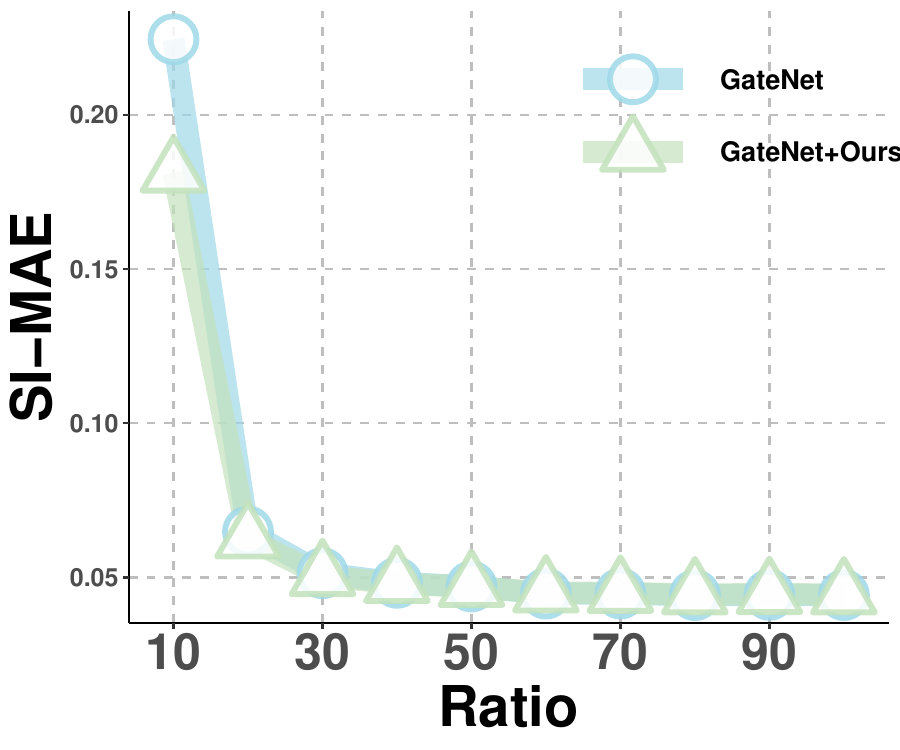}  
\label{fig:gatenet_msod_ratio_line}
\end{minipage}
}
\subfigure[DUTS]{   
\begin{minipage}{0.185\linewidth}
\includegraphics[width=\linewidth]{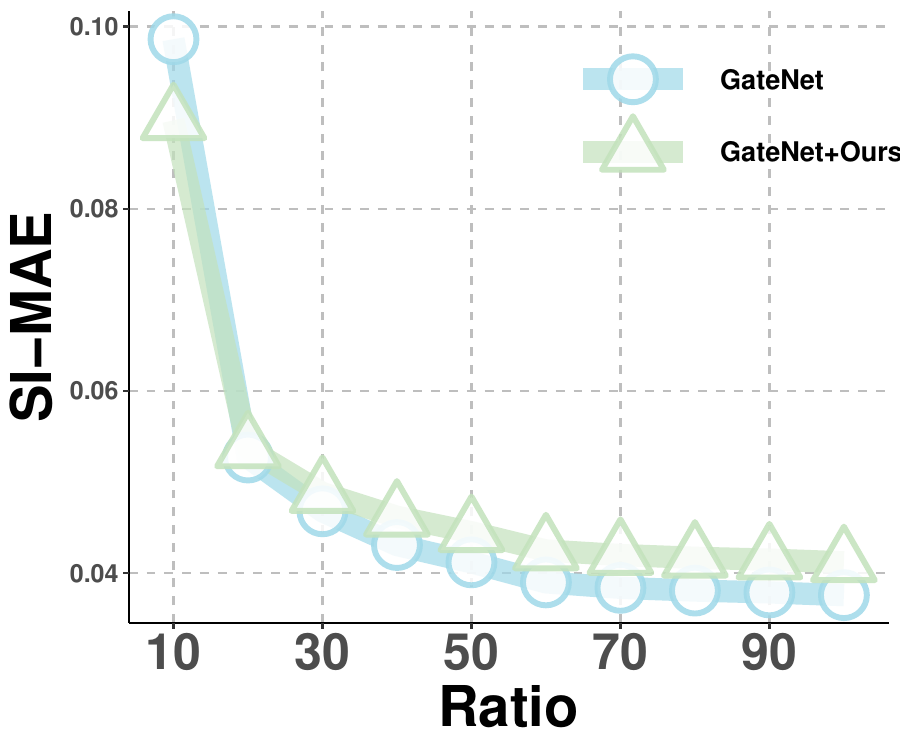}  
\label{fig:gatenet_DUTS_ratio_line}
\end{minipage}
}
\subfigure[ECSSD]{   
\begin{minipage}{0.185\linewidth}
\includegraphics[width=\linewidth]{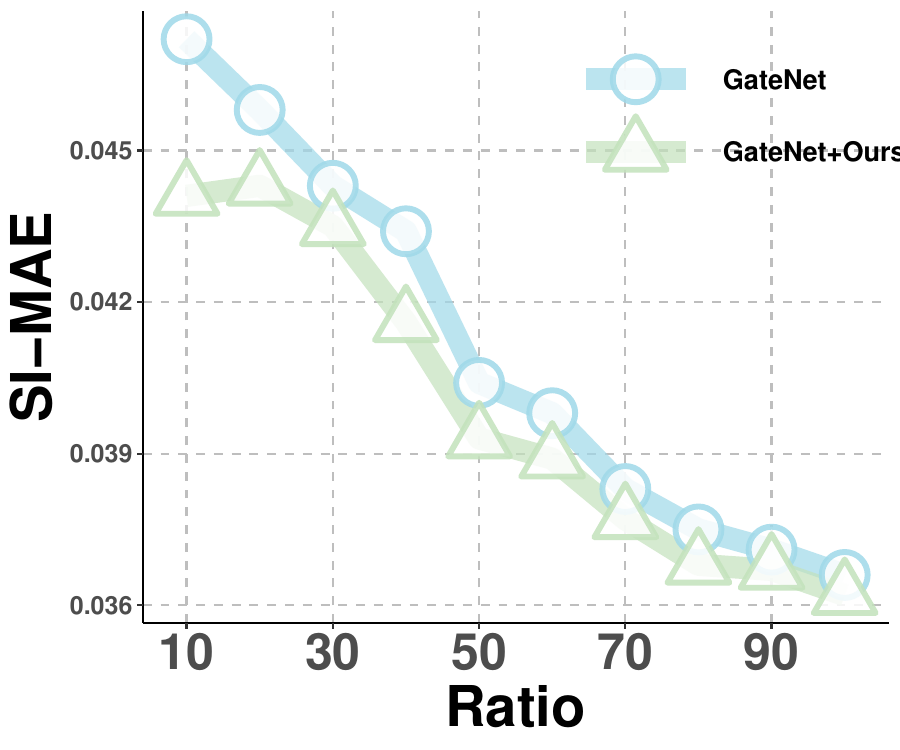}  
\label{fig:gatenet_ECSSD_ratio_line}
\end{minipage}
}
\subfigure[DUT-OMRON]{   
\begin{minipage}{0.185\linewidth}
\includegraphics[width=\linewidth]{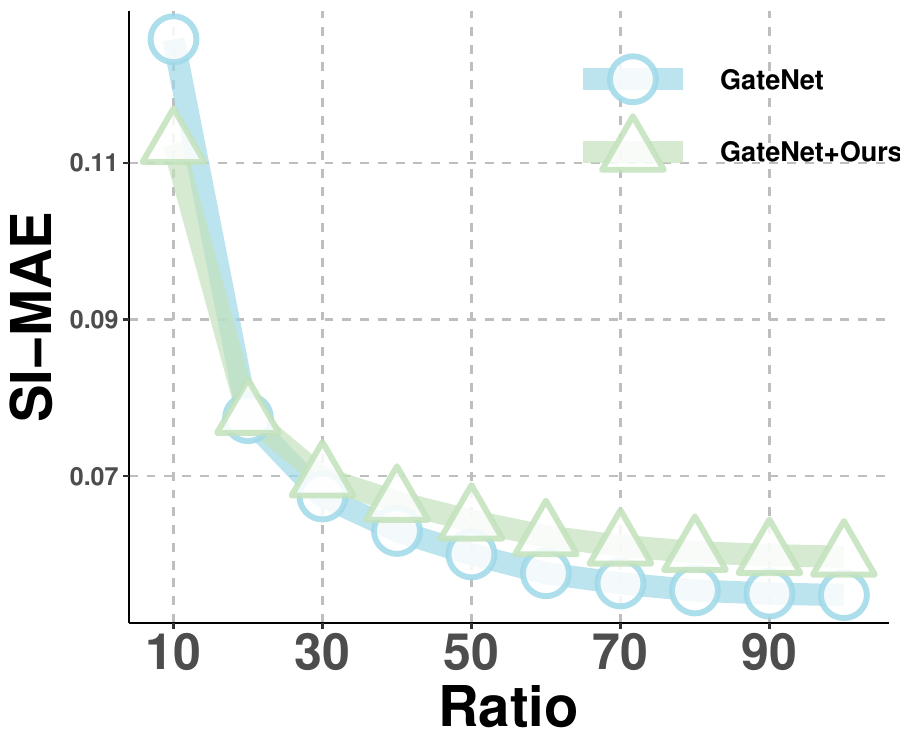}  
\label{fig:gatenet_DUT-OMRON_ratio_line}
\end{minipage}
}
\subfigure[HKU-IS]{   
\begin{minipage}{0.185\linewidth}
\includegraphics[width=\linewidth]{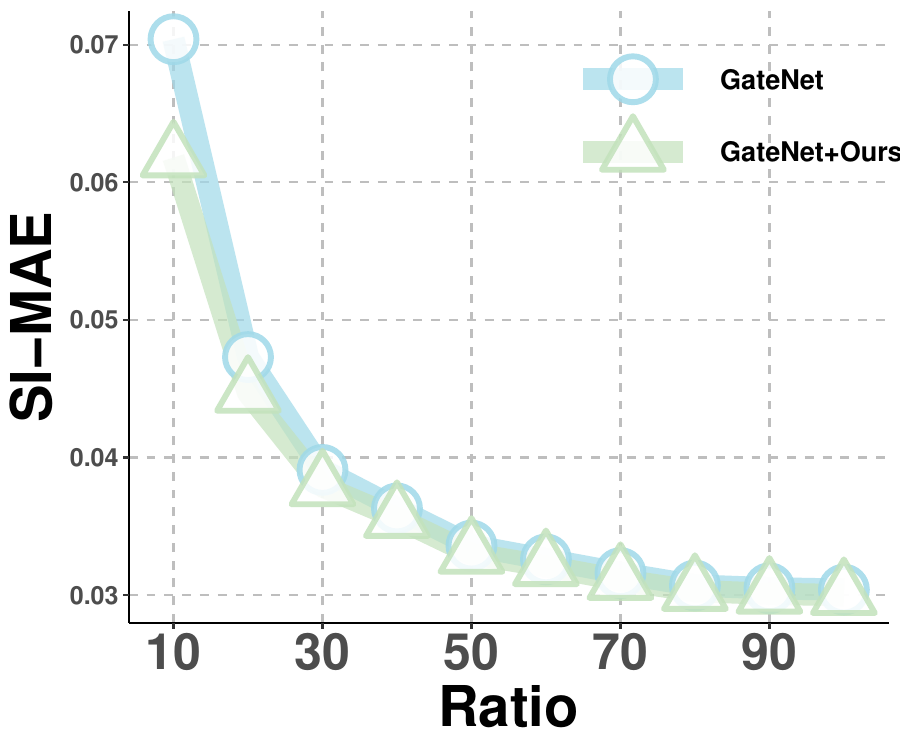}  
\label{fig:gatenet_HKU-IS_ratio_line}
\end{minipage}
}
\caption{$\SMAE$ performance on objects with different sizes on five datasets, with GateNet as the backbone.}    
\label{fig:fine-analysis-GateNet-ratio}    
\end{figure*}

\begin{figure*}[ht]
\centering
\subfigure[MSOD]{
\begin{minipage}{0.185\linewidth}
\includegraphics[width=\linewidth]{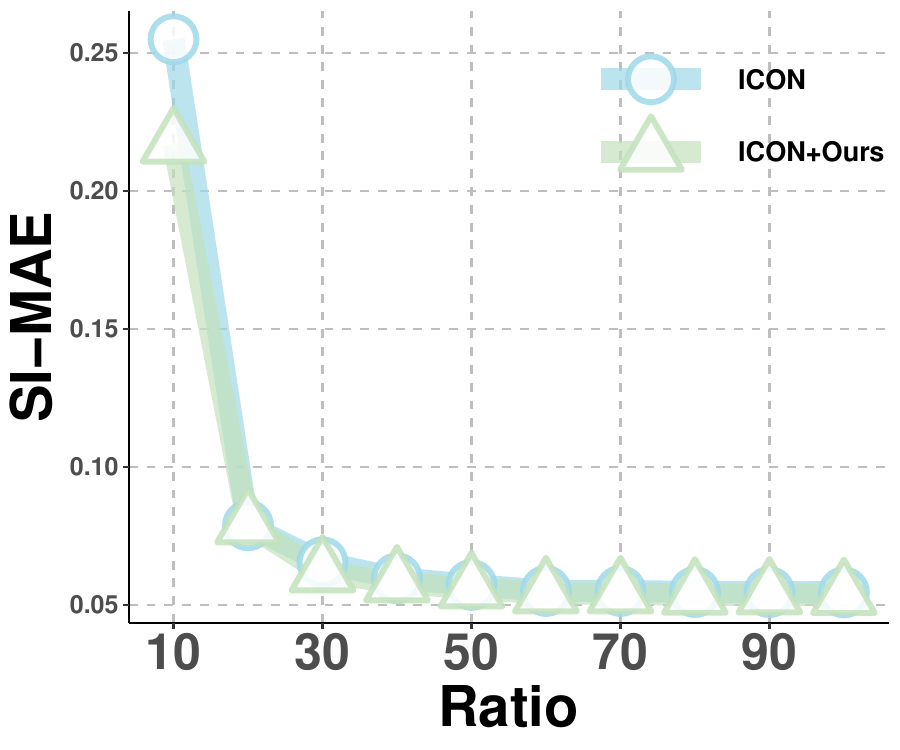}  
\label{fig:ICON_msod_ratio_line}
\end{minipage}
}
\subfigure[DUTS]{   
\begin{minipage}{0.185\linewidth}
\includegraphics[width=\linewidth]{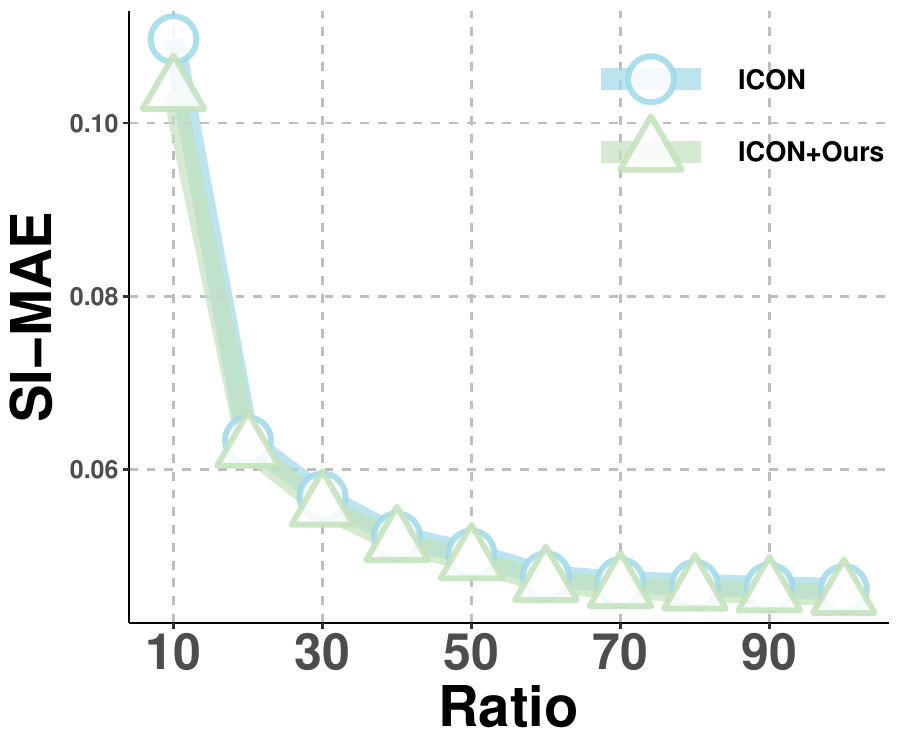}  
\label{fig:ICON_DUTS_ratio_line}
\end{minipage}
}
\subfigure[ECSSD]{   
\begin{minipage}{0.185\linewidth}
\includegraphics[width=\linewidth]{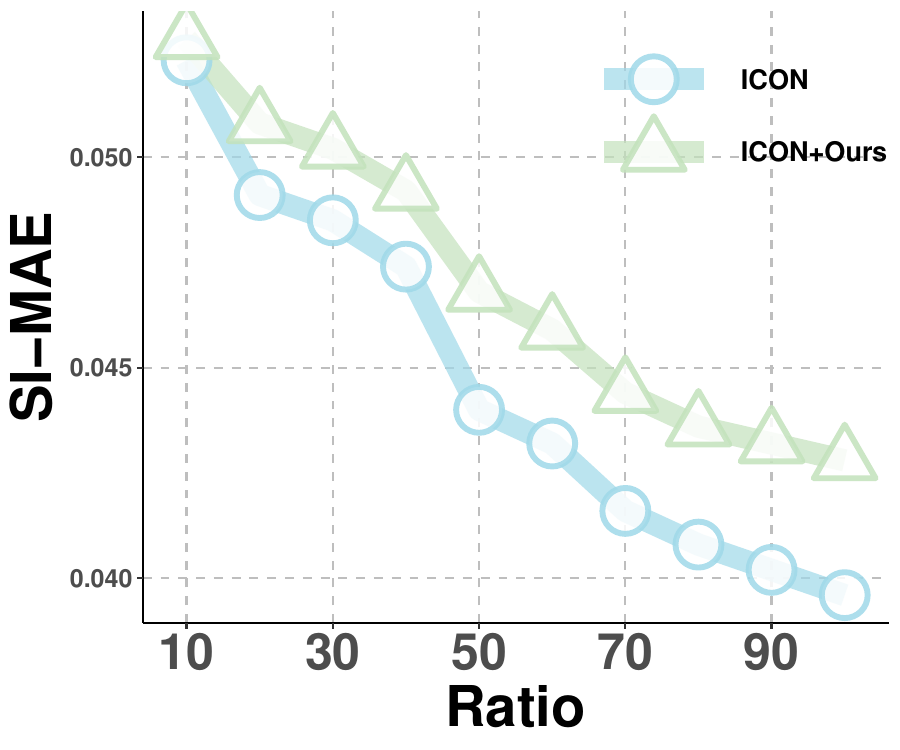}  
\label{fig:ICON_ECSSD_ratio_line}
\end{minipage}
}
\subfigure[DUT-OMRON]{   
\begin{minipage}{0.185\linewidth}
\includegraphics[width=\linewidth]{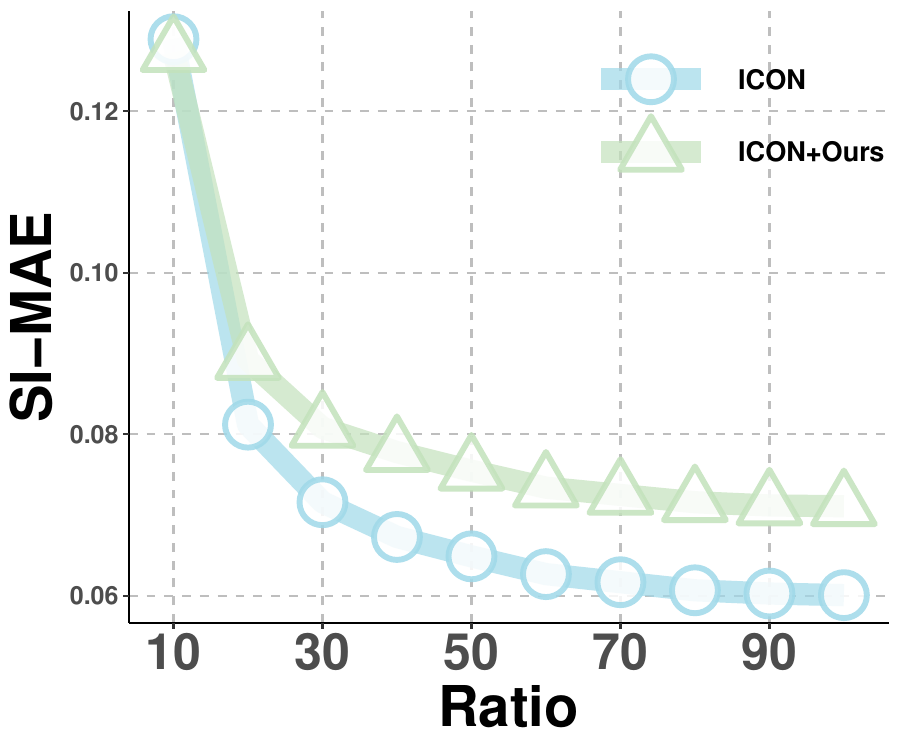}  
\label{fig:ICON_DUT-OMRON_ratio_line}
\end{minipage}
}
\subfigure[HKU-IS]{   
\begin{minipage}{0.185\linewidth}
\includegraphics[width=\linewidth]{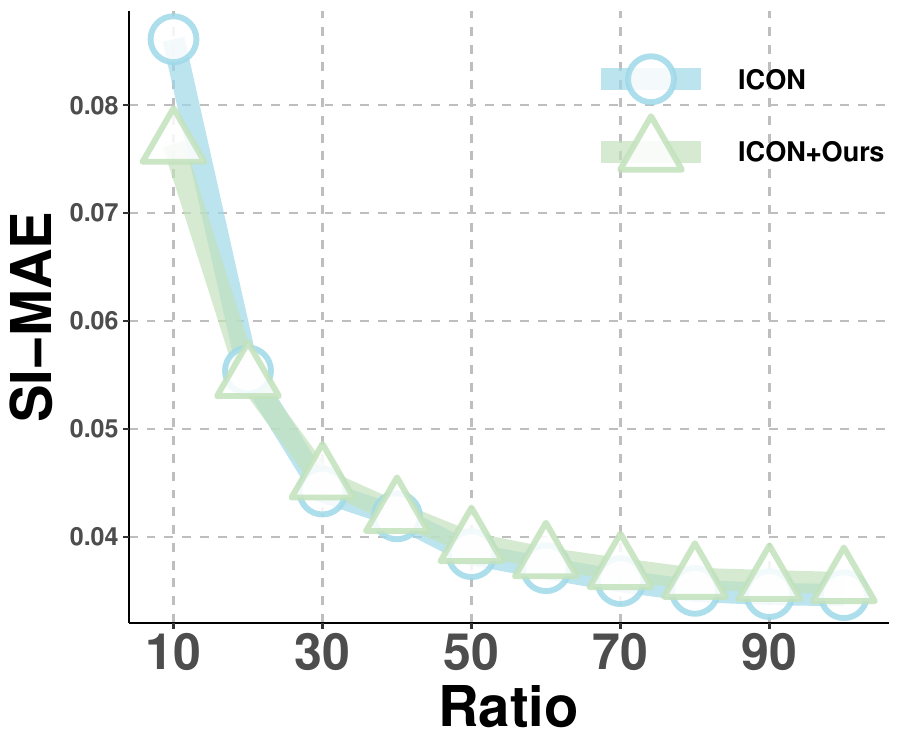}  
\label{fig:ICON_HKU-IS_ratio_line}
\end{minipage}
}
\caption{$\SMAE$ performance on objects with different sizes on five datasets, with ICON as the backbone.}    
\label{fig:fine-analysis-ICON-ratio}    
\end{figure*}

\begin{figure*}[ht]
\centering
\subfigure[MSOD]{   
\begin{minipage}{0.185\linewidth}
\includegraphics[width=\linewidth]{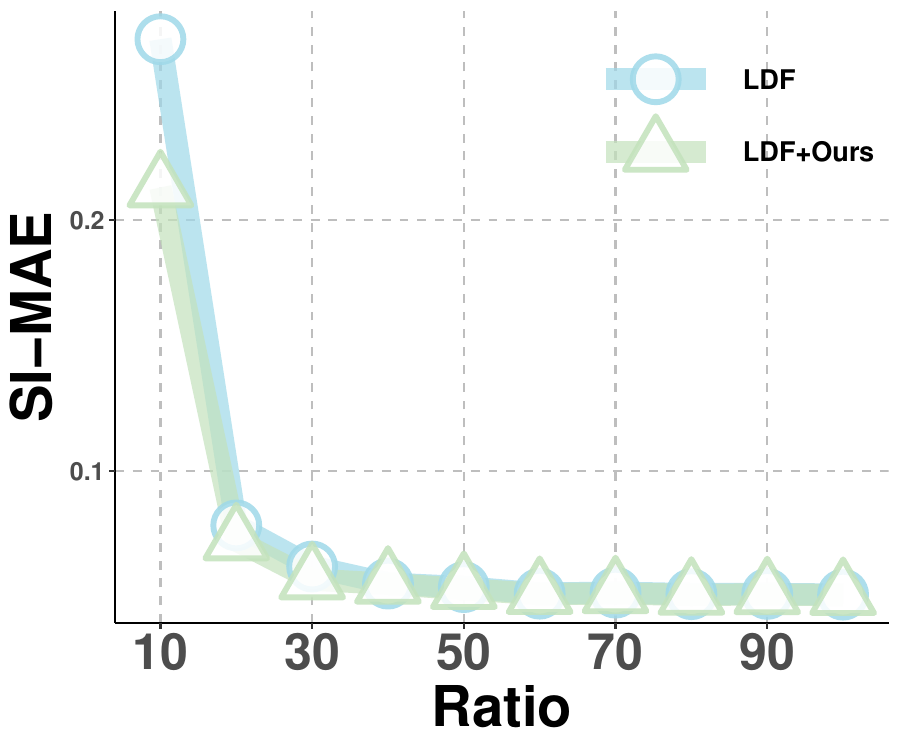}  
\label{fig:LDF_msod_ratio_line}
\end{minipage}
}
\subfigure[DUTS]{   
\begin{minipage}{0.185\linewidth}
\includegraphics[width=\linewidth]{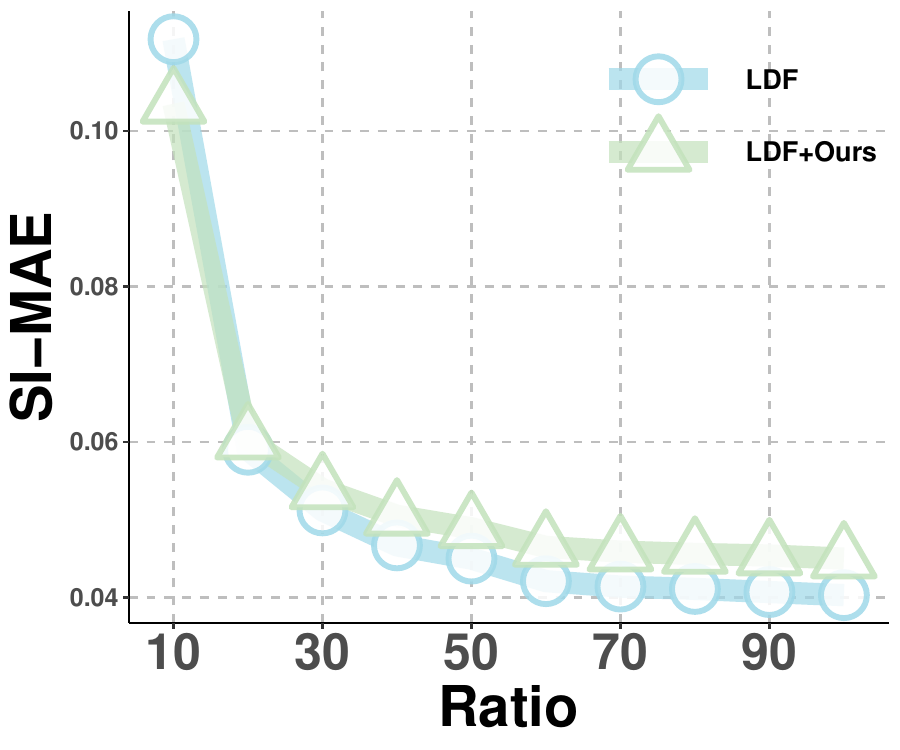}  
\label{fig:LDF_DUTS_ratio_line}
\end{minipage}
}
\subfigure[ECSSD]{   
\begin{minipage}{0.185\linewidth}
\includegraphics[width=\linewidth]{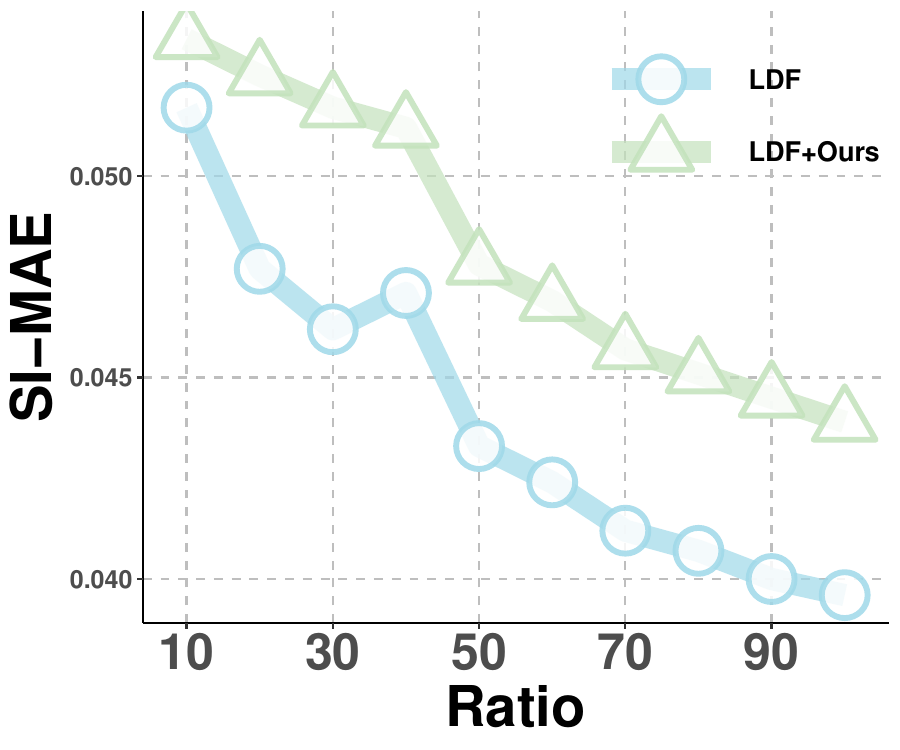}  
\label{fig:LDF_ECSSD_ratio_line}
\end{minipage}
}
\subfigure[DUT-OMRON]{   
\begin{minipage}{0.185\linewidth}
\includegraphics[width=\linewidth]{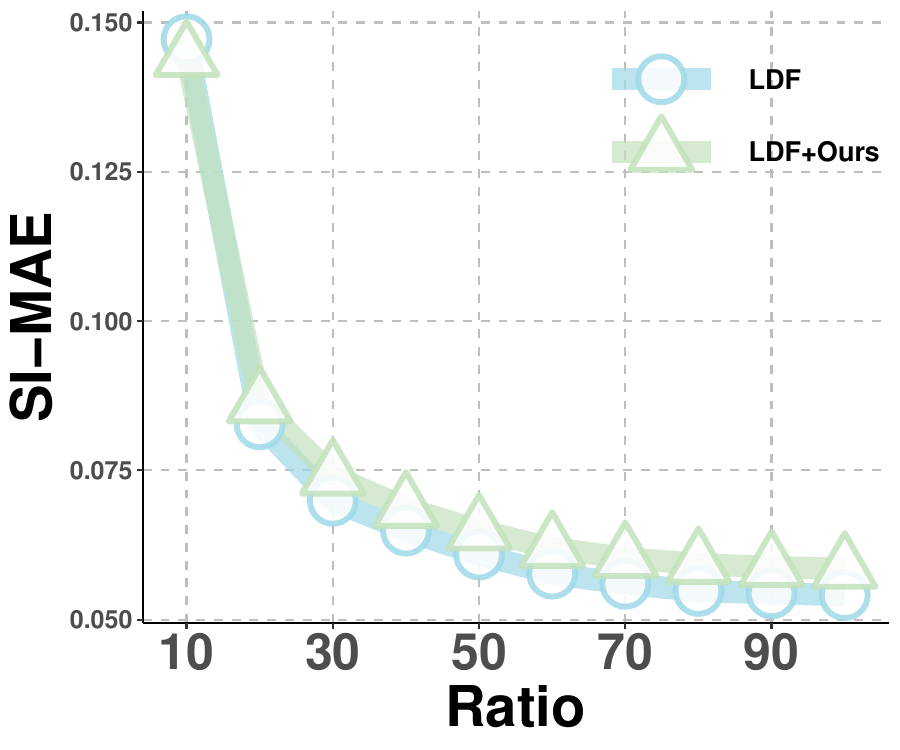}  
\label{fig:LDF_DUT-OMRON_ratio_line}
\end{minipage}
}
\subfigure[HKU-IS]{   
\begin{minipage}{0.185\linewidth}
\includegraphics[width=\linewidth]{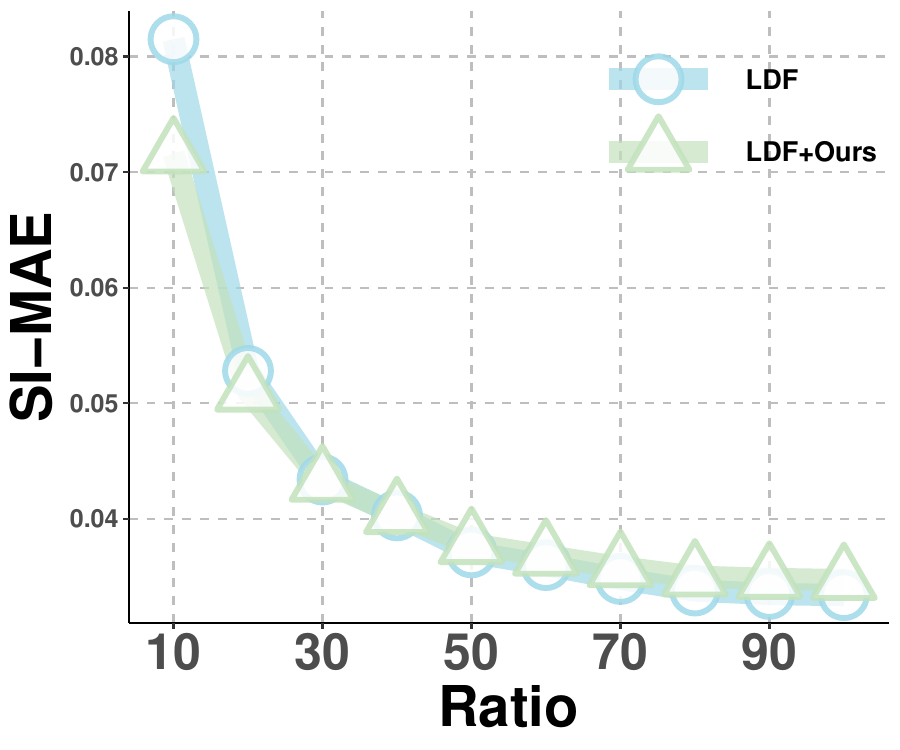}  
\label{fig:LDF_HKU-IS_ratio_line}
\end{minipage}
}
\caption{$\SMAE$ performance on objects with different sizes on five datasets, with LDF as the backbone.}    
\label{fig:fine-analysis-LDF-ratio}    
\end{figure*}
\begin{figure*}[ht]
\centering
\subfigure[MSOD]{   
\begin{minipage}{0.185\linewidth}
\includegraphics[width=\linewidth]{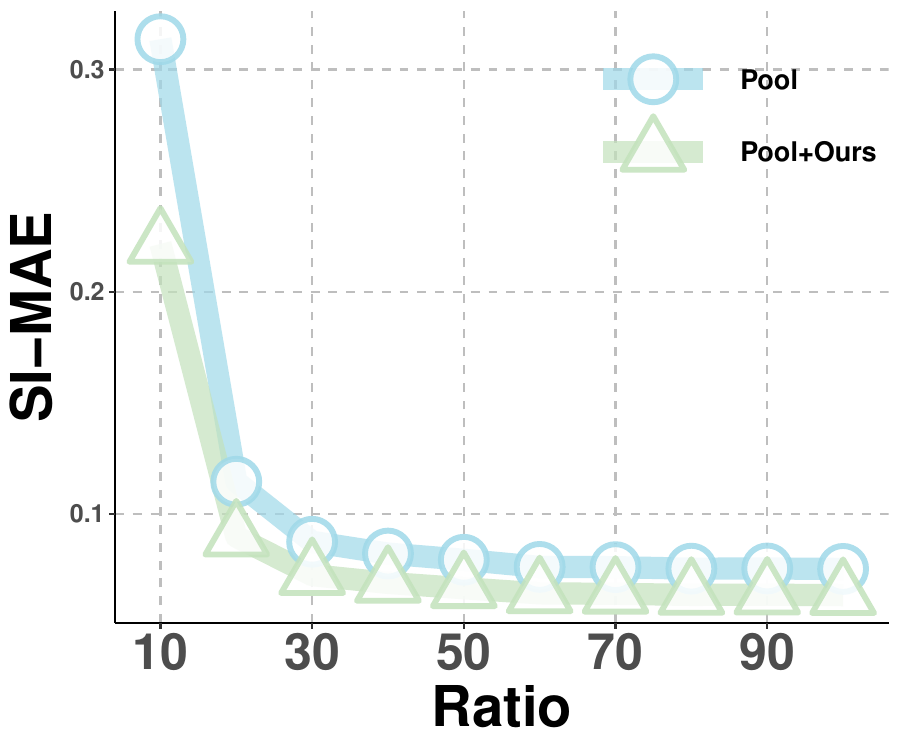}  
\label{fig:Pool_msod_ratio_line}
\end{minipage}
}
\subfigure[DUTS]{   
\begin{minipage}{0.185\linewidth}
\includegraphics[width=\linewidth]{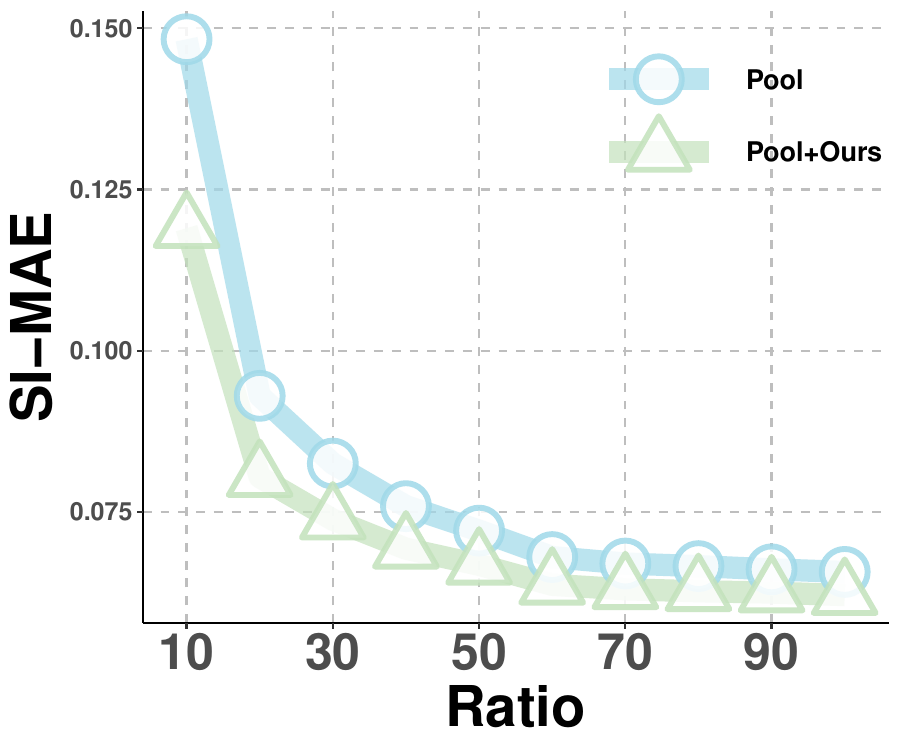}  
\label{fig:Pool_DUTS_ratio_line}
\end{minipage}
}
\subfigure[ECSSD]{   
\begin{minipage}{0.185\linewidth}
\includegraphics[width=\linewidth]{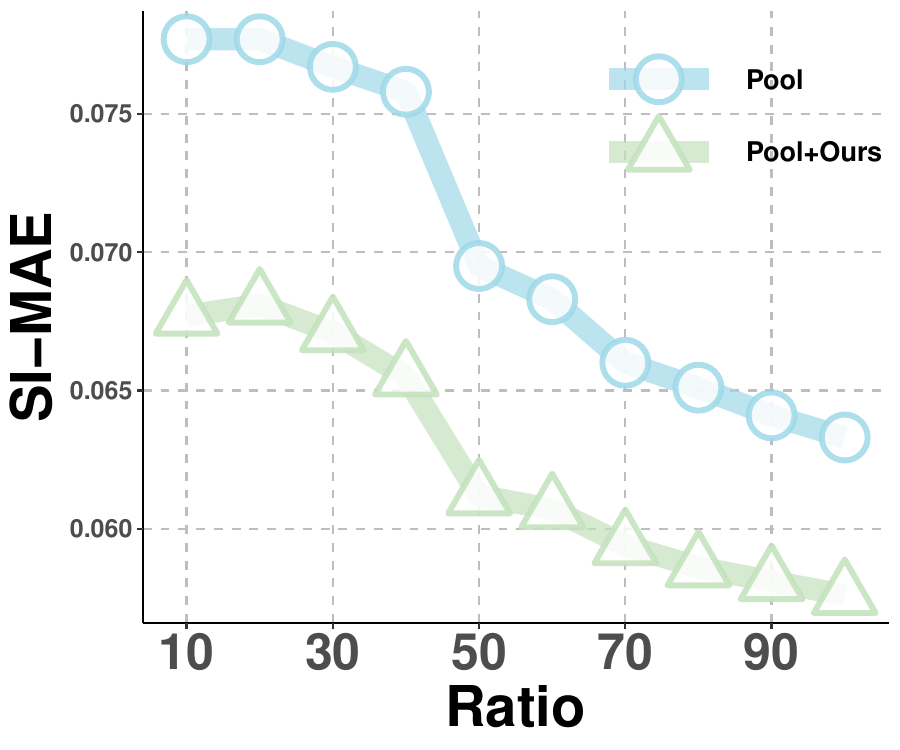}  
\label{fig:Pool_ECSSD_ratio_line}
\end{minipage}
}
\subfigure[DUT-OMRON]{   
\begin{minipage}{0.185\linewidth}
\includegraphics[width=\linewidth]{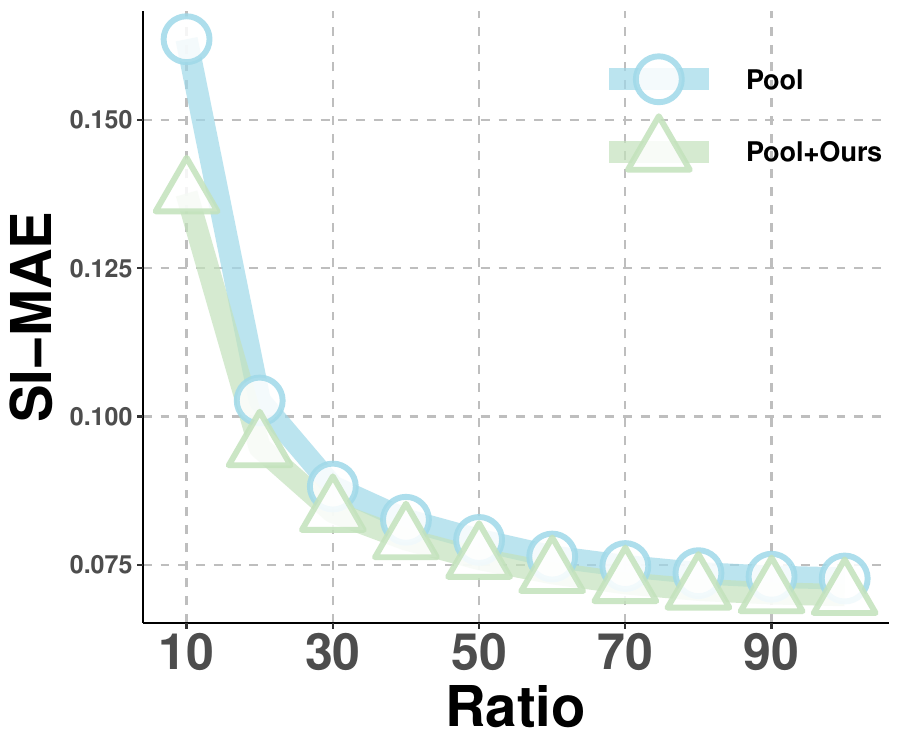}  
\label{fig:Pool_DUT-OMRON_ratio_line}
\end{minipage}
}
\subfigure[HKU-IS]{   
\begin{minipage}{0.185\linewidth}
\includegraphics[width=\linewidth]{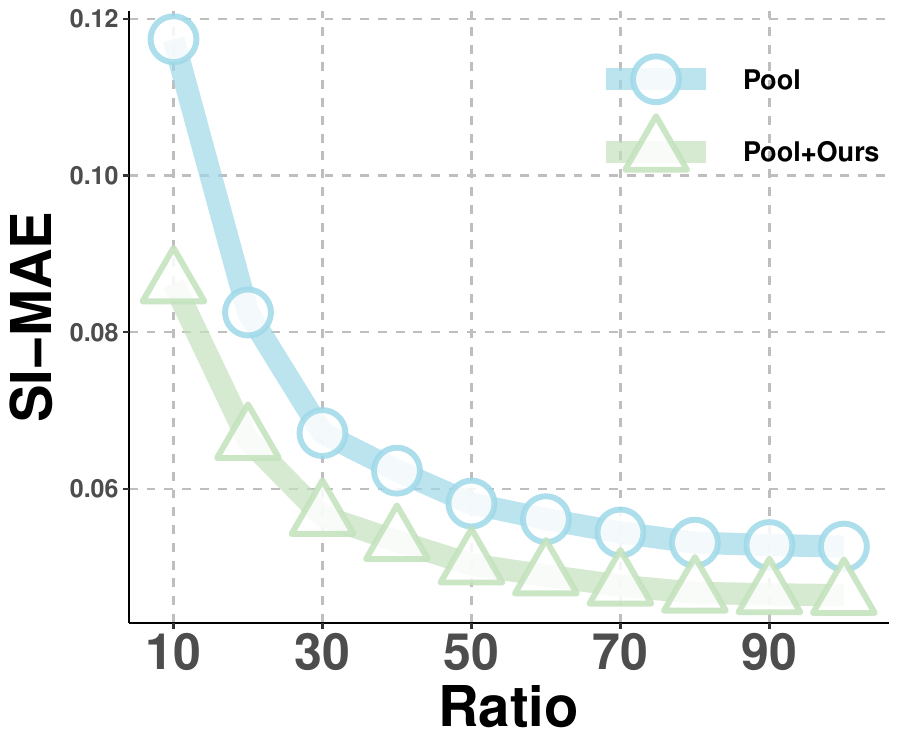}  
\label{fig:Pool_HKU-IS_ratio_line}
\end{minipage}
}
\caption{$\SMAE$ performance on objects with different sizes on five datasets, with PoolNet as the backbone.}    
\label{fig:fine-analysis-Pool-ratio}    
\end{figure*}
\FloatBarrier

\subsection{Performance with Respect to Object Numbers} \label{number-fine-grained_appendix}
Here we expand the number-relevant fine-grained analysis to other backbones and benchmarks.

\begin{figure*}[ht]
\centering
\subfigure[MSOD]{   
\begin{minipage}{0.185\linewidth}
\includegraphics[width=\linewidth]{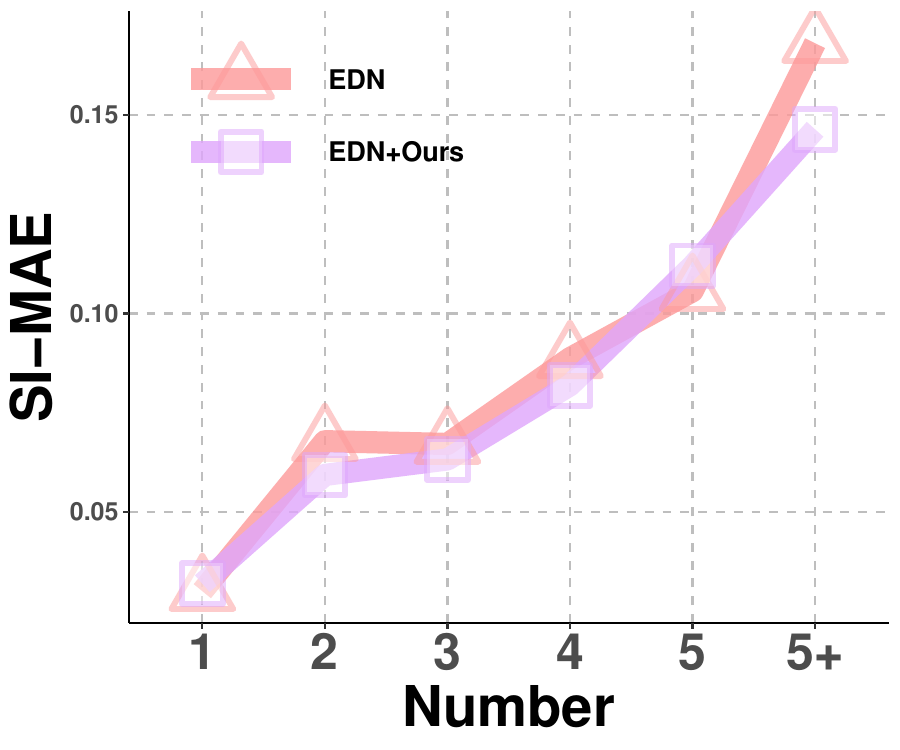}  
\label{fig:EDN_msod_num_line}
\end{minipage}
}
\subfigure[DUTS]{   
\begin{minipage}{0.185\linewidth}
\includegraphics[width=\linewidth]{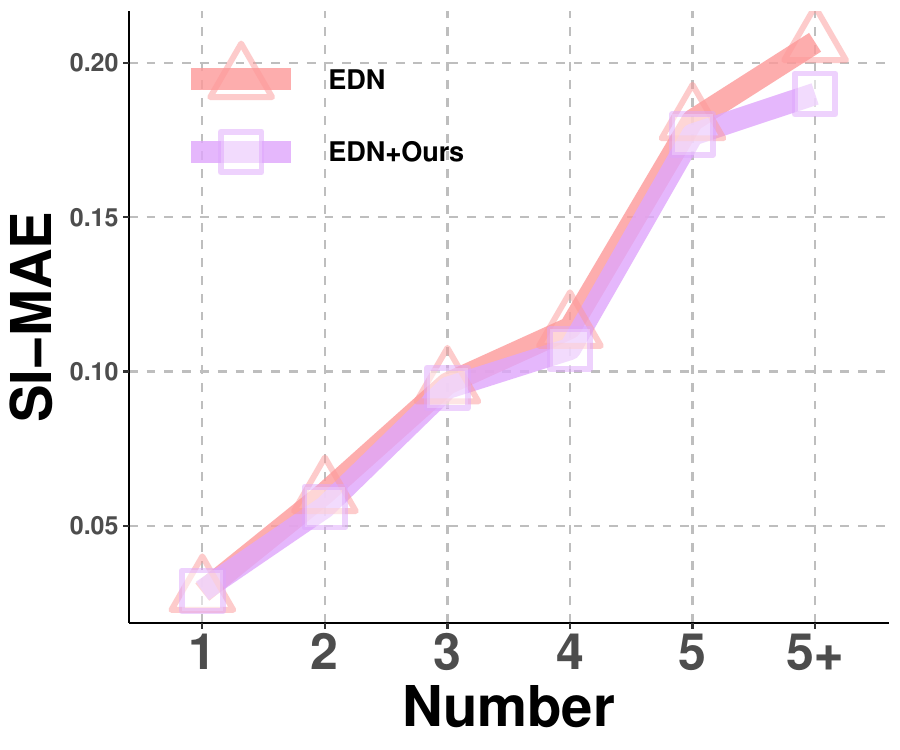}  
\label{fig:EDN_DUTS_num_line}
\end{minipage}
}
\subfigure[ECSSD]{   
\begin{minipage}{0.185\linewidth}
\includegraphics[width=\linewidth]{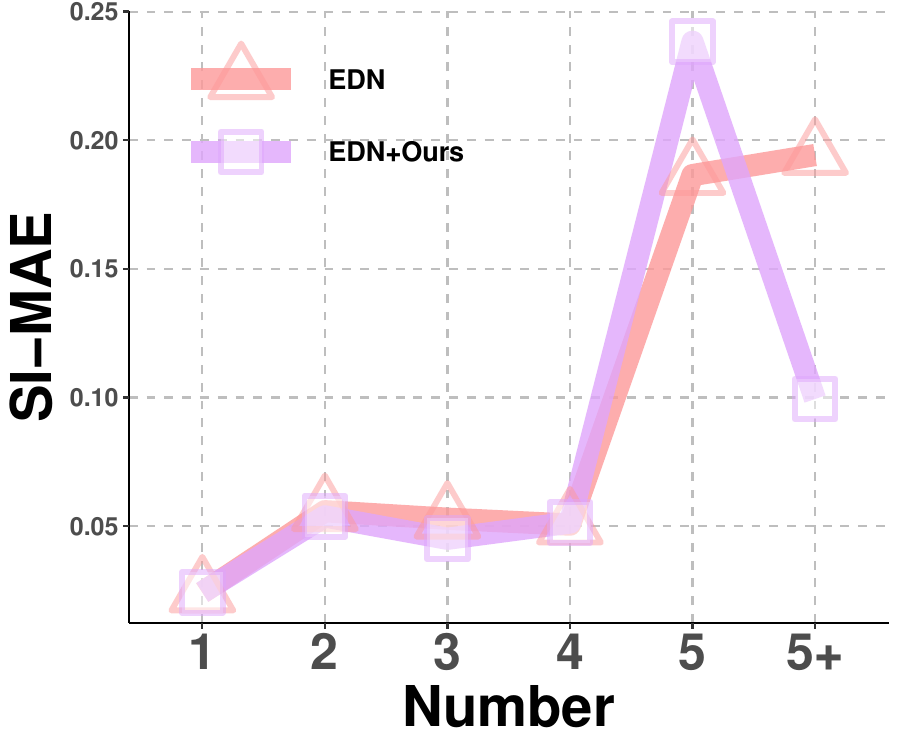}  
\label{fig:EDN_ECSSD_num_line}
\end{minipage}
}
\subfigure[DUT-OMRON]{   
\begin{minipage}{0.185\linewidth}
\includegraphics[width=\linewidth]{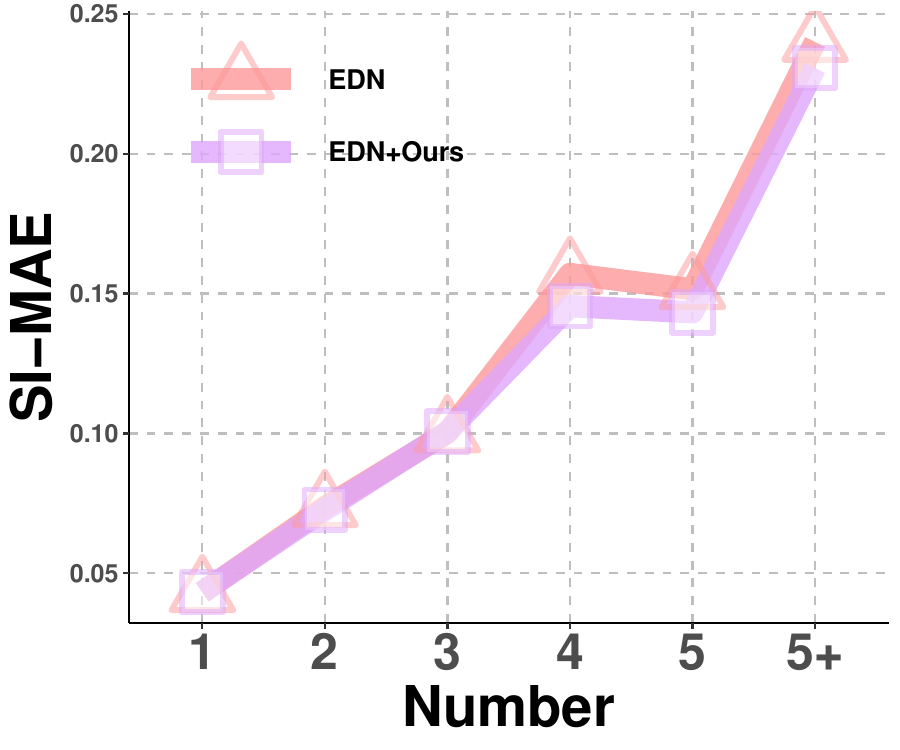}  
\label{fig:EDN_DUT-OMRON_num_line}
\end{minipage}
}
\subfigure[HKU-IS]{   
\begin{minipage}{0.185\linewidth}
\includegraphics[width=\linewidth]{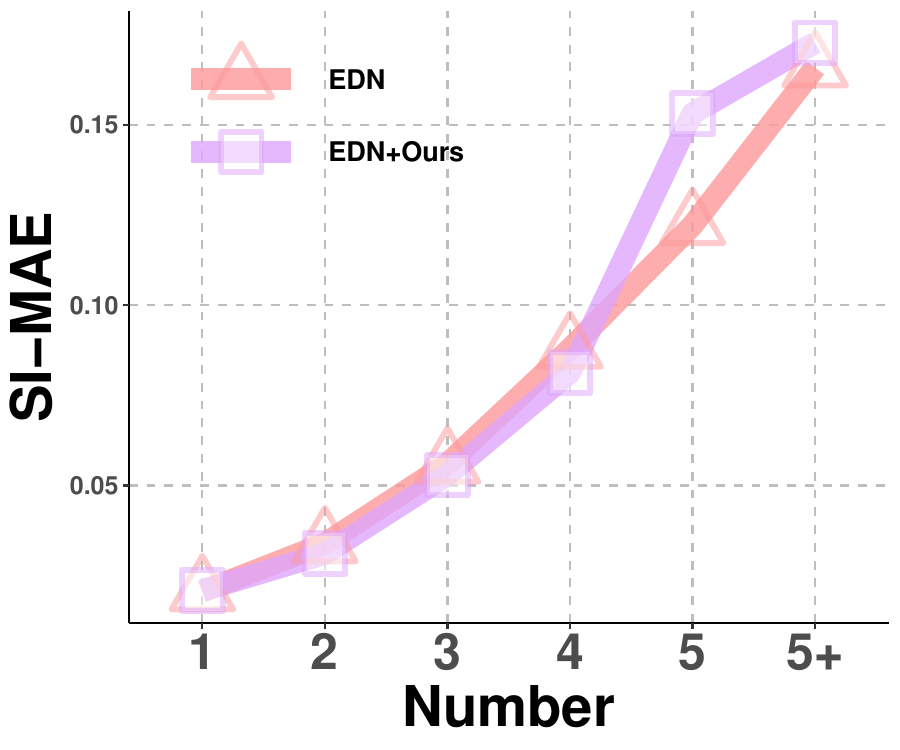}  
\label{fig:EDN_HKU-IS_num_line}
\end{minipage}
}
\caption{$\SMAE$ performance on objects with different object numbers on five datasets, with EDN as the backbone.}    
\label{fig:fine-analysis-EDN-num}    
\end{figure*}

\begin{figure*}[ht]
\centering
\subfigure[MSOD]{   
\begin{minipage}{0.185\linewidth}
\includegraphics[width=\linewidth]{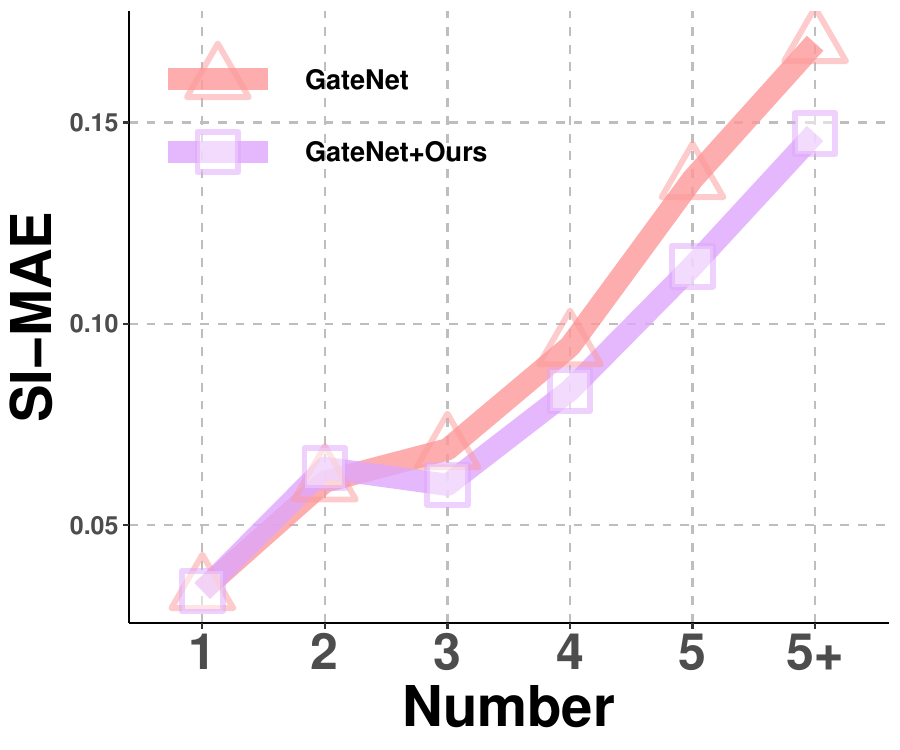}  
\label{fig:GateNet_msod_num_line}
\end{minipage}
}
\subfigure[DUTS]{   
\begin{minipage}{0.185\linewidth}
\includegraphics[width=\linewidth]{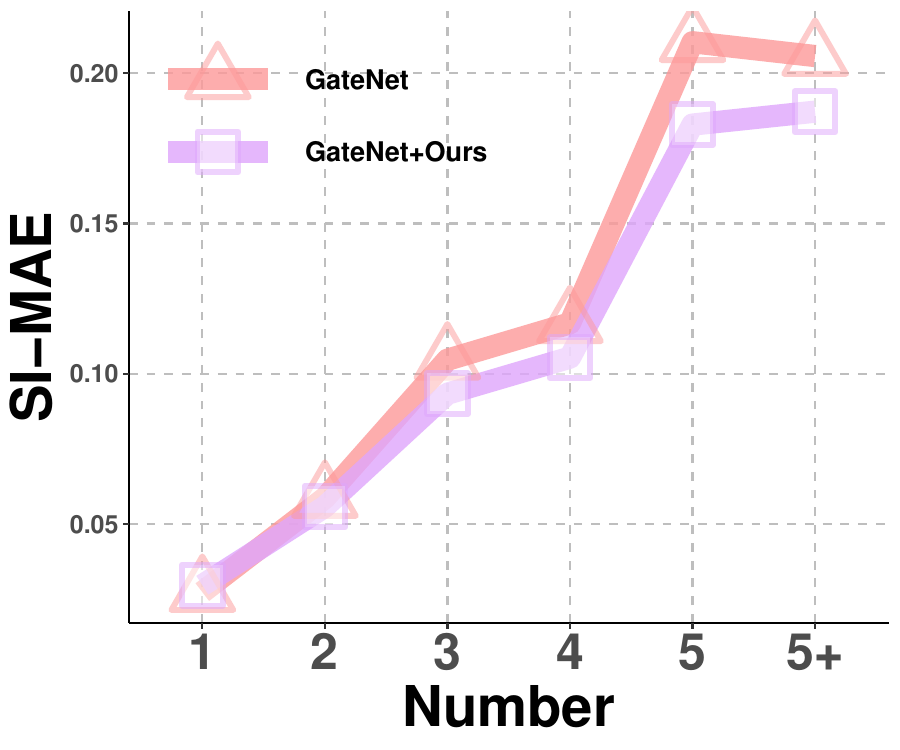}  
\label{fig:GateNet_DUTS_num_line}
\end{minipage}
}
\subfigure[ECSSD]{   
\begin{minipage}{0.185\linewidth}
\includegraphics[width=\linewidth]{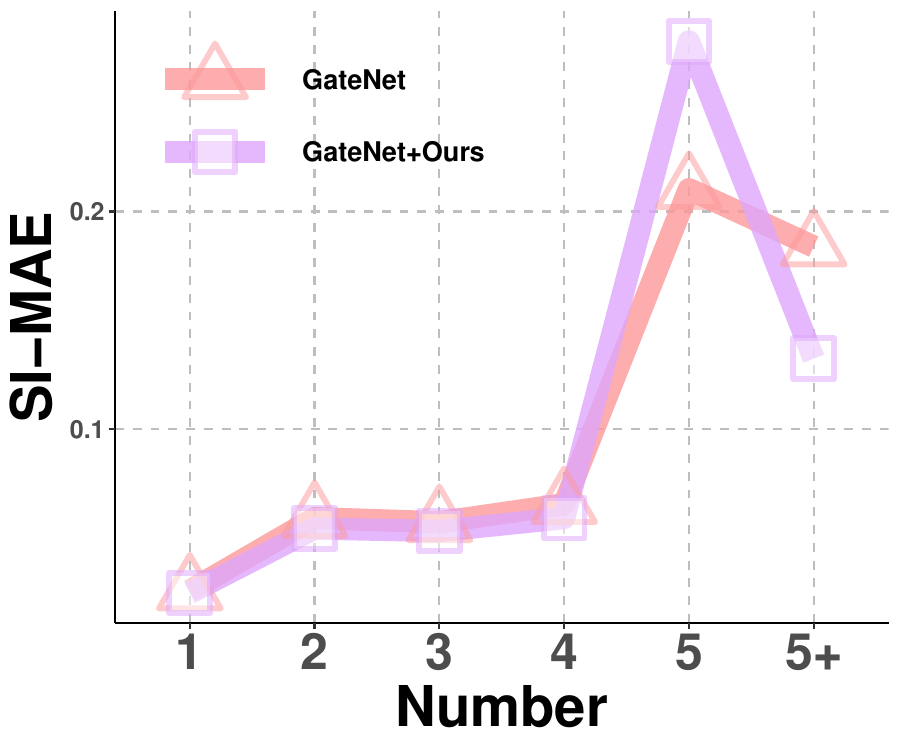}  
\label{fig:GateNet_ECSSD_num_line}
\end{minipage}
}
\subfigure[DUT-OMRON]{   
\begin{minipage}{0.185\linewidth}
\includegraphics[width=\linewidth]{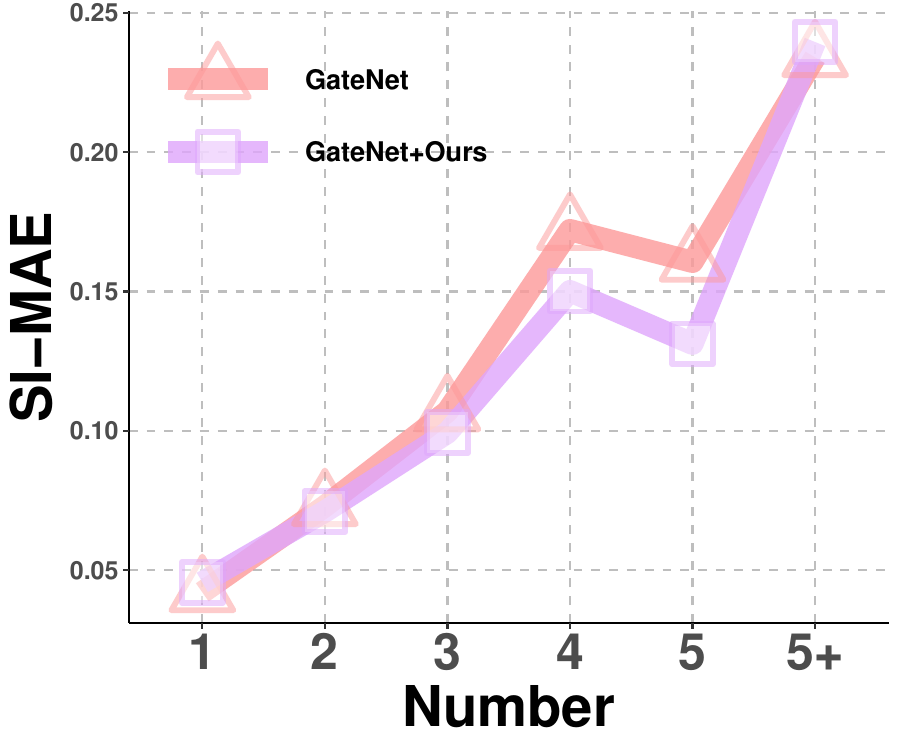}  
\label{fig:GateNet_DUT-OMRON_num_line}
\end{minipage}
}
\subfigure[HKU-IS]{   
\begin{minipage}{0.185\linewidth}
\includegraphics[width=\linewidth]{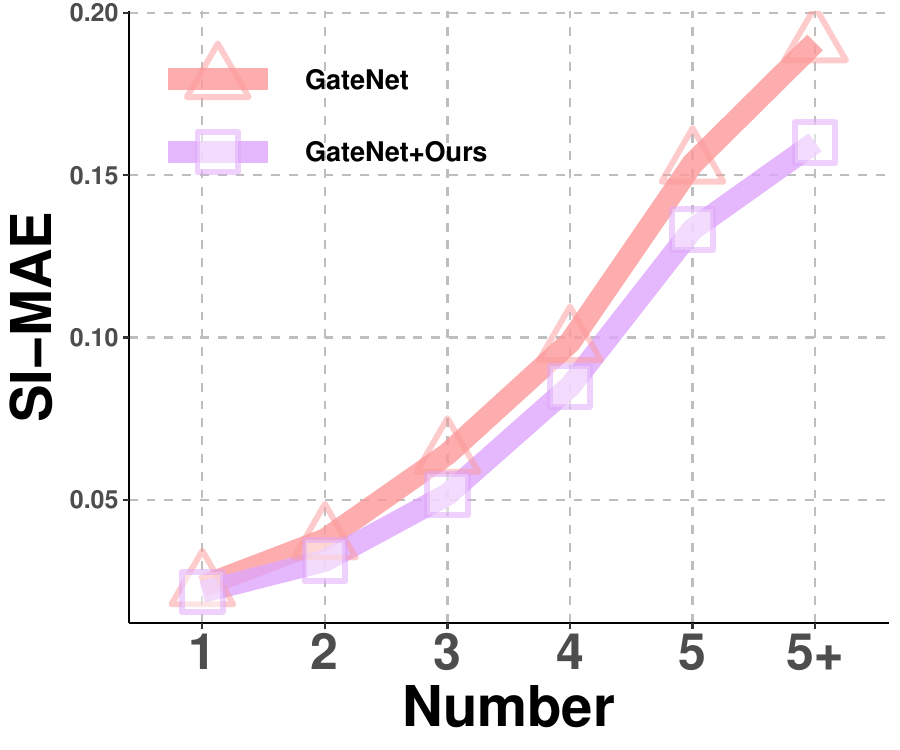}  
\label{fig:GateNet_HKU-IS_num_line}
\end{minipage}
}
\caption{$\SMAE$ performance on objects with different object numbers on five datasets, with GateNet as the backbone.}    
\label{fig:fine-analysis-GateNet-num}    
\end{figure*}

\begin{figure*}[ht]
\centering
\subfigure[MSOD]{   
\begin{minipage}{0.185\linewidth}
\includegraphics[width=\linewidth]{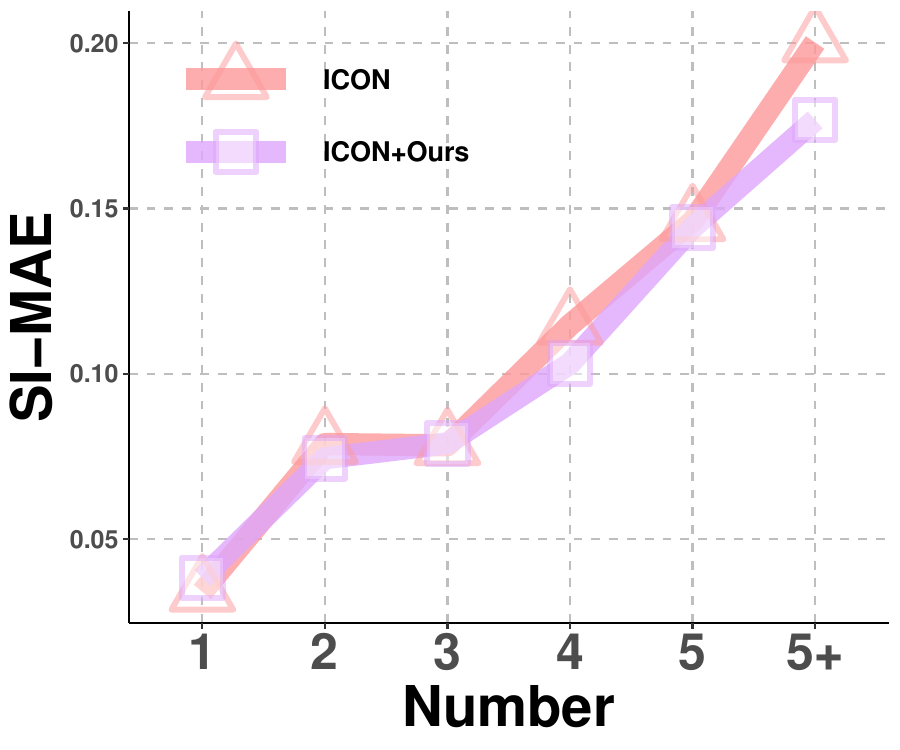}  
\label{fig:ICON_msod_num_line}
\end{minipage}
}
\subfigure[DUTS]{   
\begin{minipage}{0.185\linewidth}
\includegraphics[width=\linewidth]{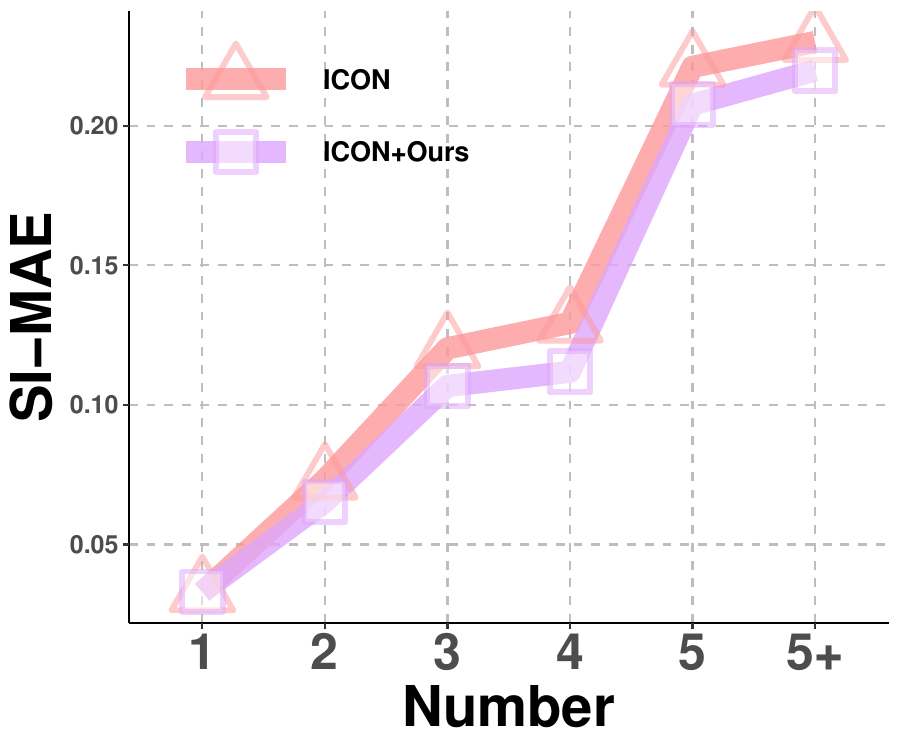}  
\label{fig:ICON_DUTS_num_line}
\end{minipage}
}
\subfigure[ECSSD]{   
\begin{minipage}{0.185\linewidth}
\includegraphics[width=\linewidth]{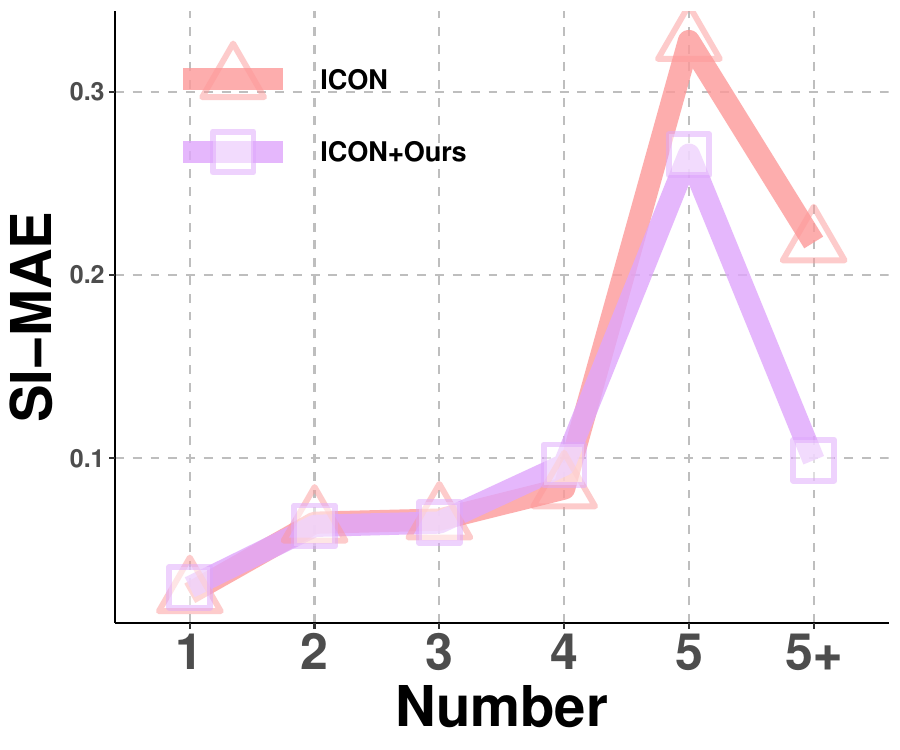}  
\label{fig:ICON_ECSSD_num_line}
\end{minipage}
}
\subfigure[DUT-OMRON]{   
\begin{minipage}{0.185\linewidth}
\includegraphics[width=\linewidth]{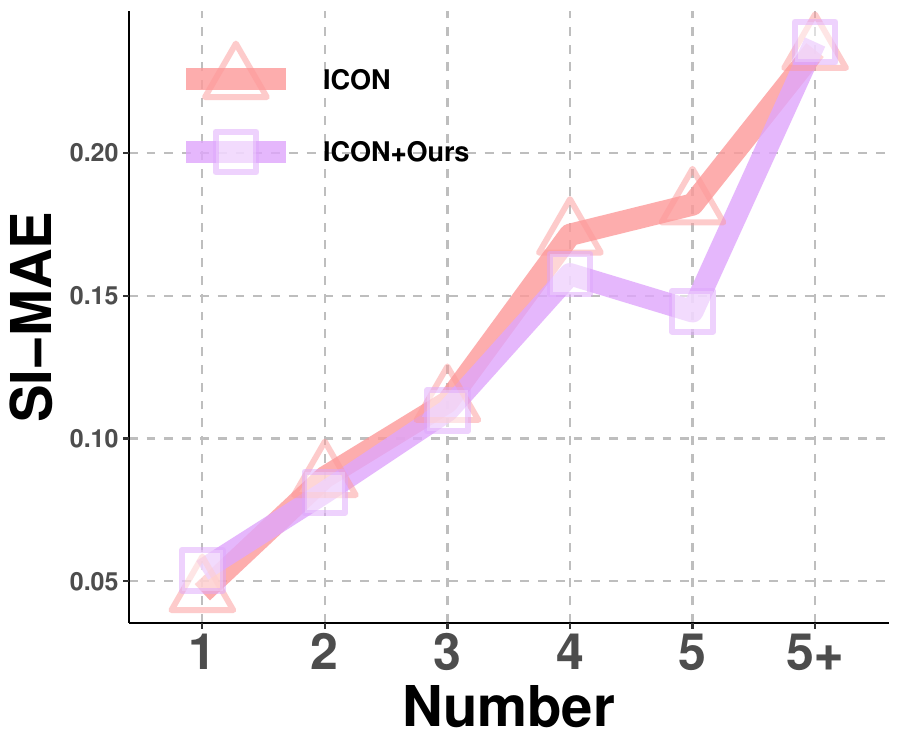}  
\label{fig:ICON_DUT-OMRON_num_line}
\end{minipage}
}
\subfigure[HKU-IS]{   
\begin{minipage}{0.185\linewidth}
\includegraphics[width=\linewidth]{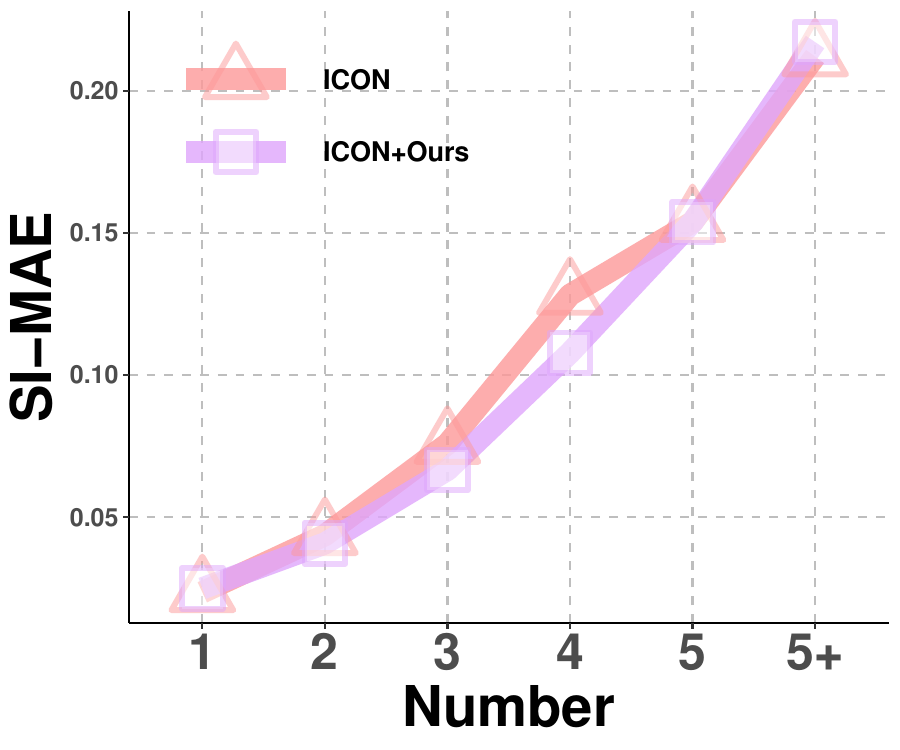}  
\label{fig:ICON_HKU-IS_num_line}
\end{minipage}
}
\caption{$\SMAE$ performance on objects with different object numbers on five datasets, with ICON as the backbone.}    
\label{fig:fine-analysis-ICON-num}    
\end{figure*}

\begin{figure*}[ht]
\centering
\subfigure[MSOD]{   
\begin{minipage}{0.185\linewidth}
\includegraphics[width=\linewidth]{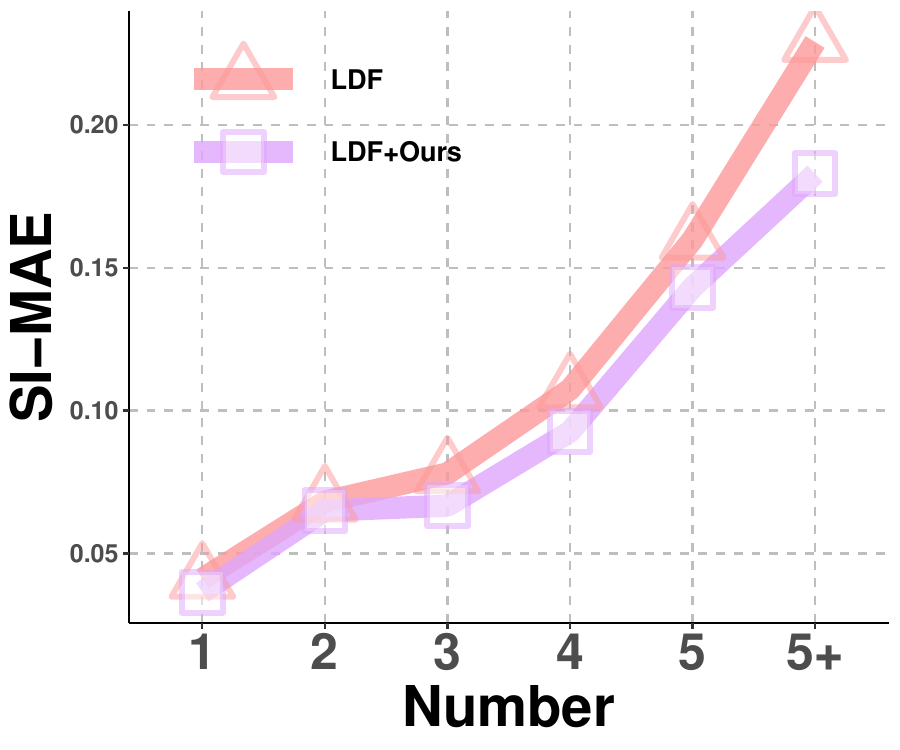}  
\label{fig:LDF_msod_num_line}
\end{minipage}
}
\subfigure[DUTS]{   
\begin{minipage}{0.185\linewidth}
\includegraphics[width=\linewidth]{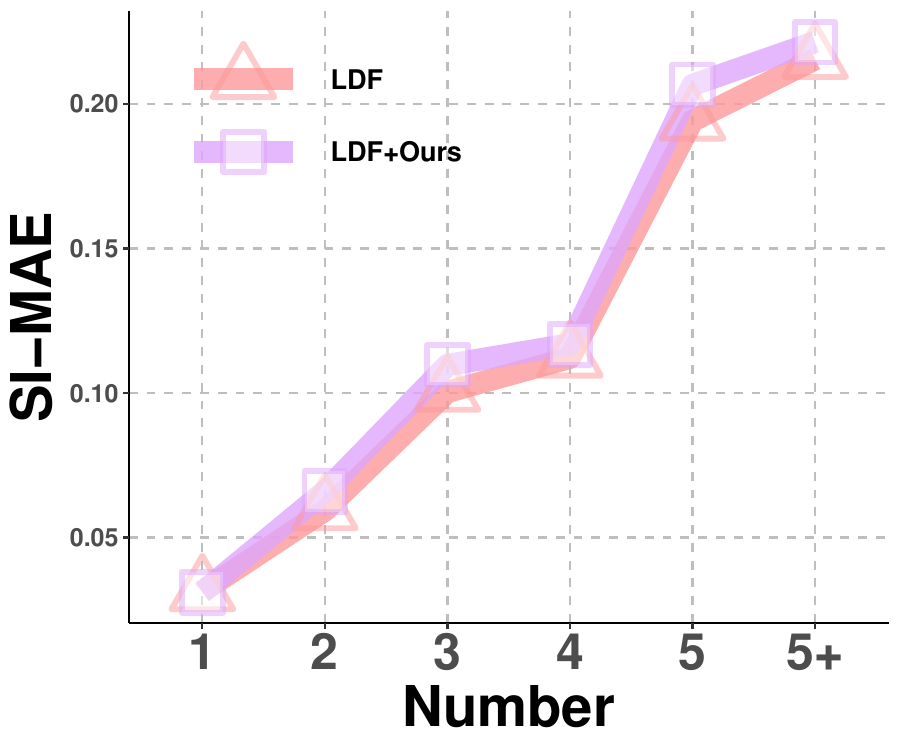}  
\label{fig:LDF_DUTS_num_line}
\end{minipage}
}
\subfigure[ECSSD]{   
\begin{minipage}{0.185\linewidth}
\includegraphics[width=\linewidth]{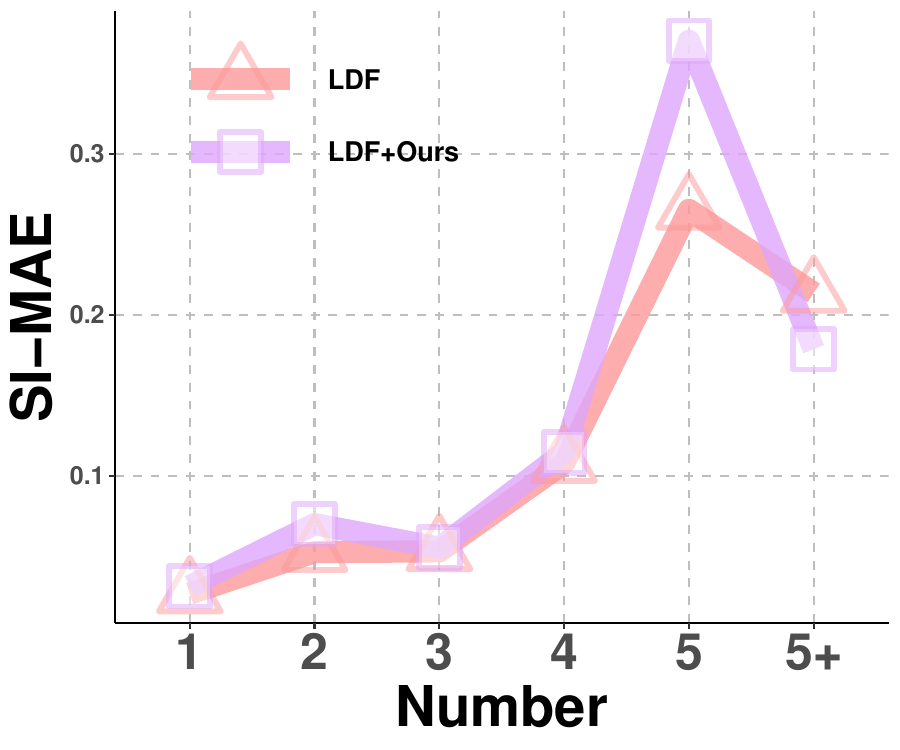}  
\label{fig:LDF_ECSSD_num_line}
\end{minipage}
}
\subfigure[DUT-OMRON]{   
\begin{minipage}{0.185\linewidth}
\includegraphics[width=\linewidth]{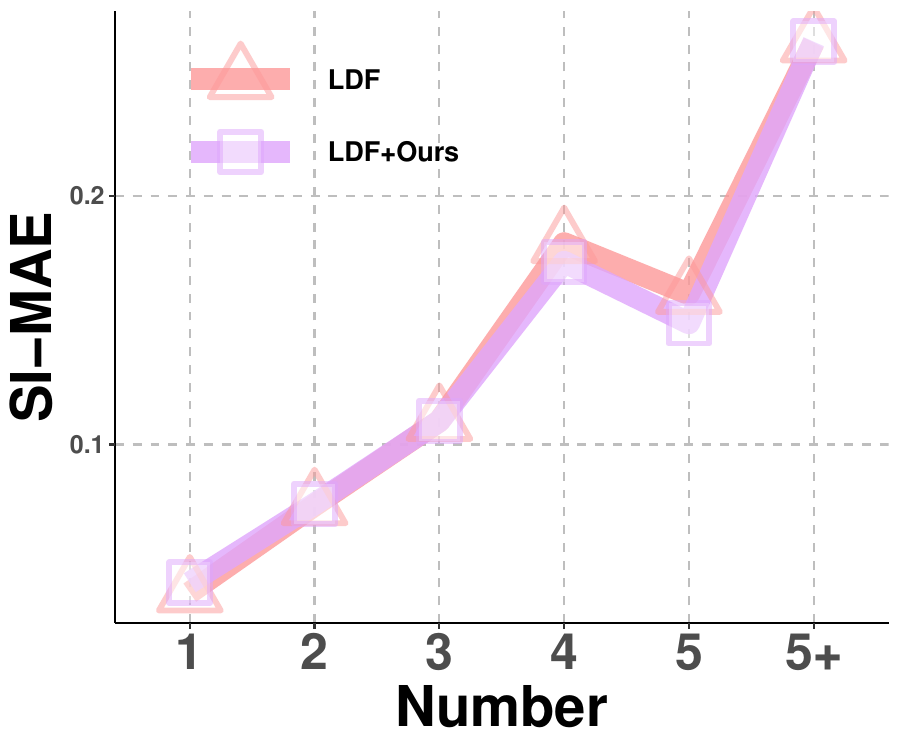}  
\label{fig:LDF_DUT-OMRON_num_line}
\end{minipage}
}
\subfigure[HKU-IS]{   
\begin{minipage}{0.185\linewidth}
\includegraphics[width=\linewidth]{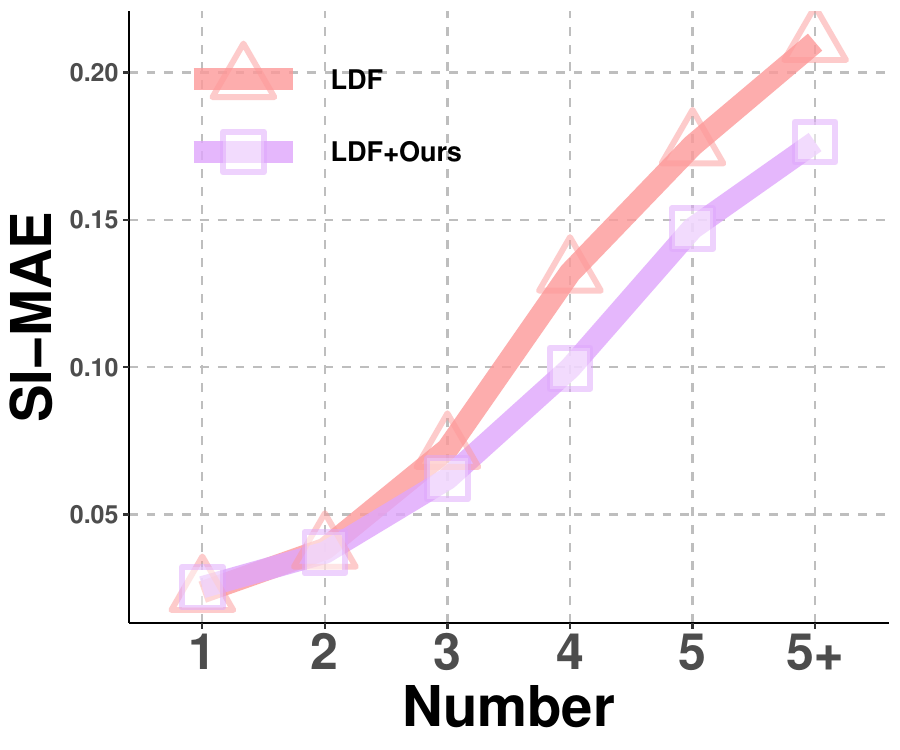}  
\label{fig:LDF_HKU-IS_num_line}
\end{minipage}
}
\caption{$\SMAE$ performance on objects with different object numbers on five datasets, with LDF as the backbone.}    
\label{fig:fine-analysis-LDF-num}    
\end{figure*}

\begin{figure*}[ht]
\centering
\subfigure[MSOD]{   
\begin{minipage}{0.185\linewidth}
\includegraphics[width=\linewidth]{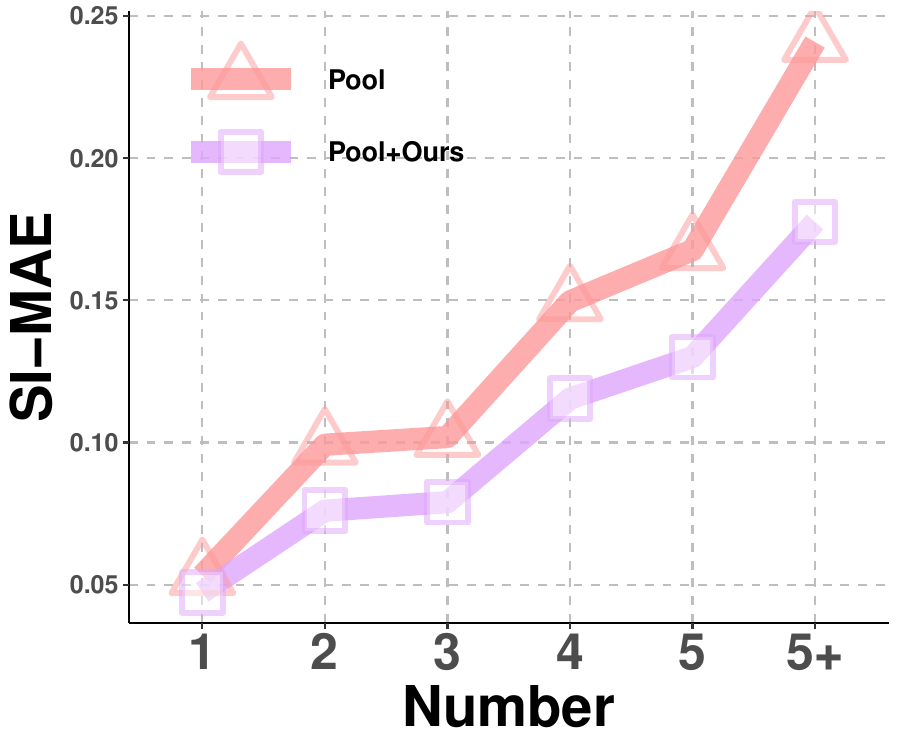}  
\label{fig:Pool_msod_num_line}
\end{minipage}
}
\subfigure[DUTS]{   
\begin{minipage}{0.185\linewidth}
\includegraphics[width=\linewidth]{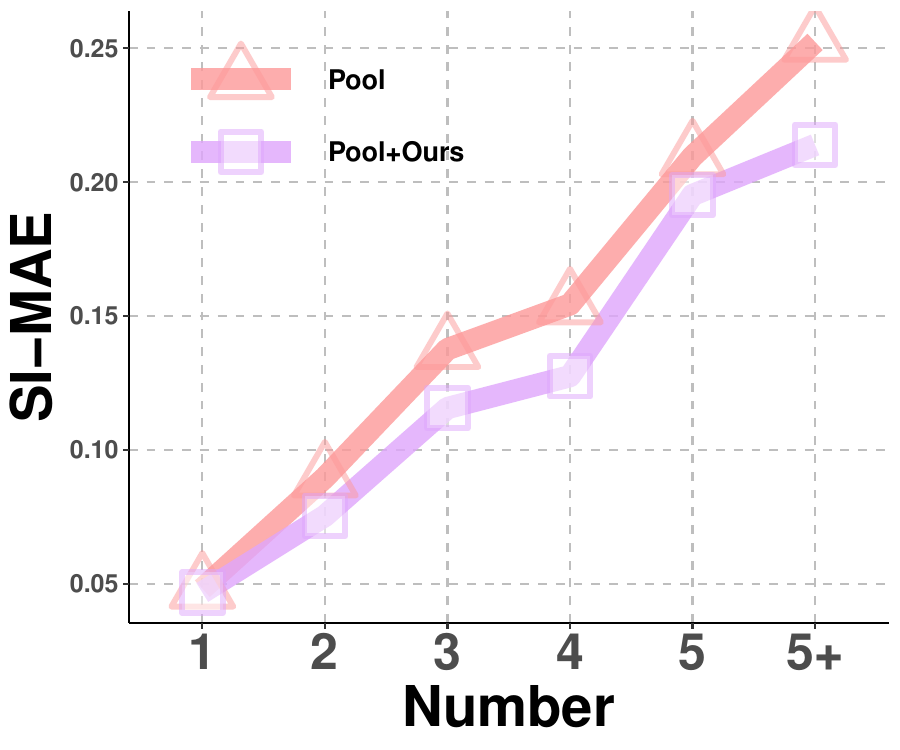}  
\label{fig:Pool_DUTS_num_line}
\end{minipage}
}
\subfigure[ECSSD]{   
\begin{minipage}{0.185\linewidth}
\includegraphics[width=\linewidth]{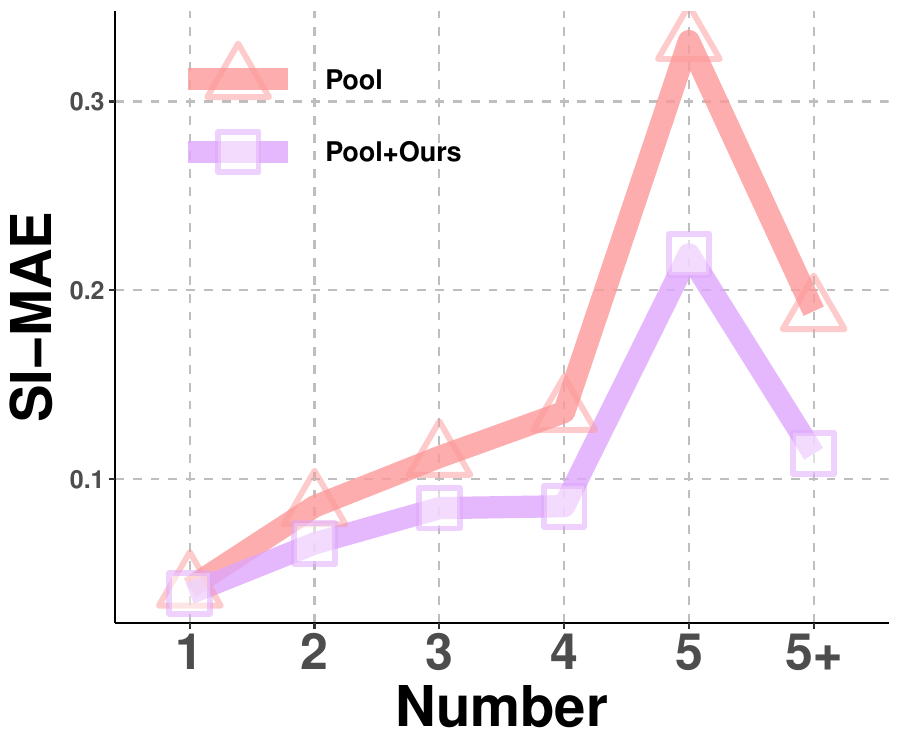}  
\label{fig:Pool_ECSSD_num_line}
\end{minipage}
}
\subfigure[DUT-OMRON]{   
\begin{minipage}{0.185\linewidth}
\includegraphics[width=\linewidth]{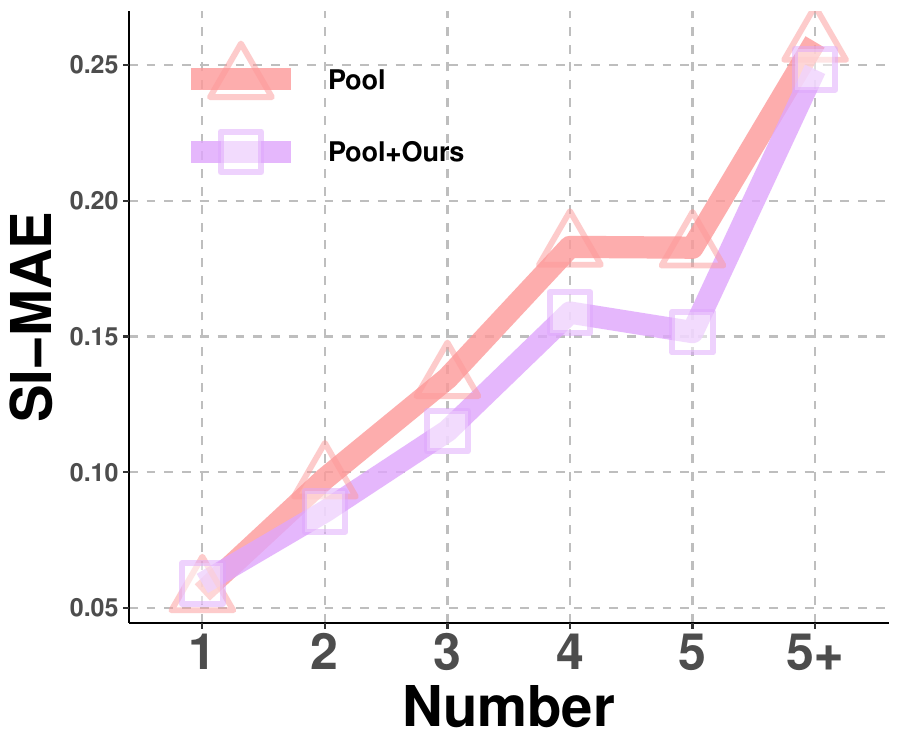}  
\label{fig:Pool_DUT-OMRON_num_line}
\end{minipage}
}
\subfigure[HKU-IS]{   
\begin{minipage}{0.185\linewidth}
\includegraphics[width=\linewidth]{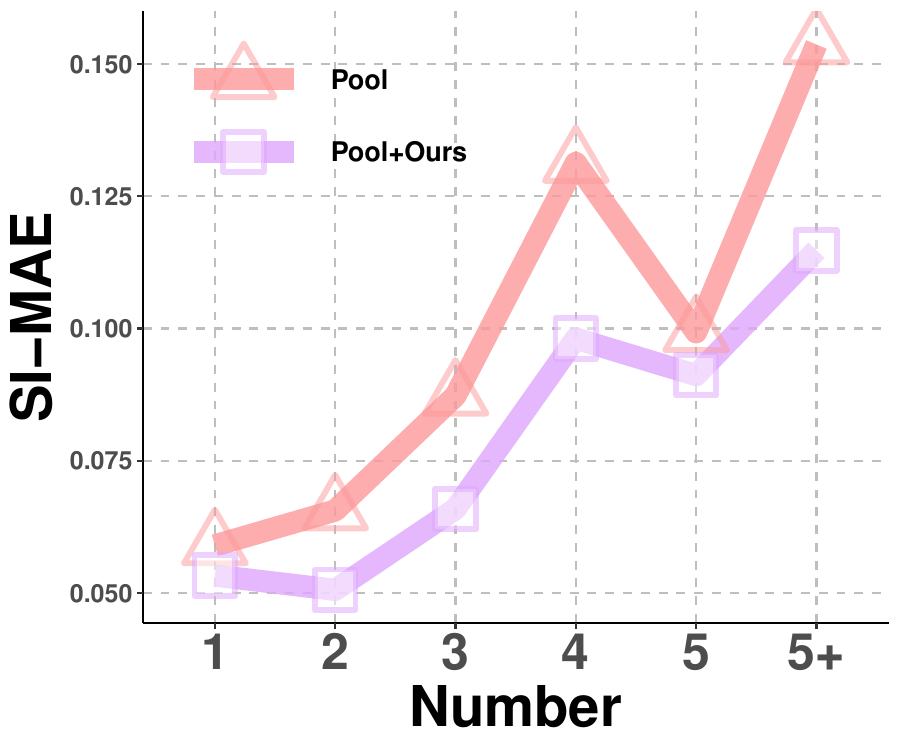}  
\label{fig:Pool_HKU-IS_num_line}
\end{minipage}
}
\caption{$\SMAE$ performance on objects with different object numbers on five datasets, with PoolNet as the backbone.}    
\label{fig:fine-analysis-Pool-num}    
\end{figure*}
\FloatBarrier

\subsection{Ablation Studies} \label{ablation_appendix}
Here we display the detailed results of the ablation studies in \cref{Ablation Studies}.
\begin{table*}[!h]
    \centering
    \caption{Ablation on the parameter $\alpha$ on MSOD(300 images). The best results are marked in bold.}
    \scalebox{0.8}{
    \begin{tabular}{c|c|cccccccc}
    \toprule
        Methods & $\alpha$ & $\MAE \downarrow $ & $\SMAE \downarrow$ & $\AUC \uparrow $ & $\F_m^\beta \uparrow$ & $\SF_m^{\beta} \uparrow $ & $\F_{max}\uparrow$ & $\SF_{max} \uparrow$ & $\E_m \uparrow$  \\
    \midrule
        EDN (A) & ResNet50 & 0.0467 & 0.0788 & 0.9196 & 0.7925 & 0.7635 & 0.8410 & 0.8321 & 0.8712 \\ 
        \textbf{+ \textit{Ours}} (B) & 0 & 0.2340 & 0.2000 & 0.9217 & 0.4790	& \textbf{0.8547} & 0.7695 & \textbf{0.9136} & 0.6199 \\
        \textbf{+ \textit{Ours}} (C) & 1 & 0.0544 & 0.0812 & 0.9360 & 0.7893	& 0.7850 & 0.8502 & 0.8912 & 0.8825 \\ 
        \textbf{+ \textit{Ours}} (D) & $\frac{S^{fore}}{S^{back}}$ & \textbf{0.0453} & \textbf{0.0724} & \textbf{0.9401} & \textbf{0.8057}	& 0.7990	& \textbf{0.8555} & 0.8619 & \textbf{0.8936} \\ 
    \bottomrule
    \end{tabular}
    }
    \label{tab:ablation_alpha_msod}
\end{table*}
\begin{table*}[!h]
    \centering
    \caption{Ablation on the parameter $\alpha$ on DUTS(5,019 images). The best results are marked in bold.}
    \scalebox{0.8}{
    \begin{tabular}{c|c|cccccccc}
    \toprule
        Methods & $\alpha$ & $\MAE \downarrow $ & $\SMAE \downarrow$ & $\AUC \uparrow $ & $\F_m^\beta \uparrow$ & $\SF_m^{\beta} \uparrow $ & $\F_{max}\uparrow$ & $\SF_{max} \uparrow$ & $\E_m \uparrow$  \\
    \midrule
        EDN (A) & ResNet50 & \textbf{0.0389} & 0.0388 & 0.9600 & \textbf{0.8288} & 0.8565 & 0.8752 & 0.9017 & 0.9033 \\
        \textbf{+ \textit{Ours}} (B) & 0 & 0.2318 & 0.1975 & 0.9354 & 0.4621 & \textbf{0.8730} & 0.7705 & \textbf{0.9235} & 0.6069 \\
        \textbf{+ \textit{Ours}} (C) & 1 & 0.0489 & 0.0460 & 0.9634 & 0.8146 & 0.8585 & 0.8807 & 0.9182 & 0.8954 \\
        \textbf{+ \textit{Ours}} (D) & $\frac{S^{fore}}{S^{back}}$ & 0.0392 & \textbf{0.0381} & \textbf{0.9658} & 0.8260 & 0.8672 & \textbf{0.8765} & 0.9119 & \textbf{0.9072} \\ 
    \bottomrule
    \end{tabular}
    }
    \label{tab:ablation_alpha_DUTS}
\end{table*}
\begin{table*}[!h]
    \centering
    \caption{Ablation on the parameter $\alpha$ on ECSSD(1,000 images). The best results are marked in bold.}
    \scalebox{0.8}{
    \begin{tabular}{c|c|cccccccc}    
    \toprule
        Methods & $\alpha$ & $\MAE \downarrow $ & $\SMAE \downarrow$ & $\AUC \uparrow $ & $\F_m^\beta \uparrow$ & $\SF_m^{\beta} \uparrow $ & $\F_{max}\uparrow$ & $\SF_{max} \uparrow$ & $\E_m \uparrow$  \\
    \midrule
        EDN (A) & ResNet50 & 0.0363 & 0.0271 & \textbf{0.9767} & \textbf{0.9089} & 0.9147 & \textbf{0.9531} & 0.9560 & 0.9338 \\ 
        \textbf{+ \textit{Ours}} (B) & 0 & 0.1656 & 0.1282 & 0.9633 & 0.6557 & \textbf{0.9236} & 0.9043 & \textbf{0.9587} & 0.7431 \\
        \textbf{+ \textit{Ours}} (C) & 1 & 0.0454 & 0.0340 & 0.9740 & 0.8986 & 0.9164 & 0.9457 & 0.9556 & 0.9282 \\
        \textbf{+ \textit{Ours}} (D) & $\frac{S^{fore}}{S^{back}}$ & \textbf{0.0358} & \textbf{0.0269} & 0.9762 & 0.9084 & 0.9216 & 0.9456 & 0.9543 & \textbf{0.9375} \\
    \bottomrule
    \end{tabular}
    }
    \label{tab:ablation_alpha_ECSSD}
\end{table*}
\begin{table*}[!h]
    \centering
    \caption{Ablation on the parameter $\alpha$ on DUT-OMRON(5,168 images). The best results are marked in bold.}
    \scalebox{0.8}{
    \begin{tabular}{c|c|cccccccc}
    \toprule
        Methods & $\alpha$ & $\MAE \downarrow $ & $\SMAE \downarrow$ & $\AUC \uparrow $ & $\F_m^\beta \uparrow$ & $\SF_m^{\beta} \uparrow $ & $\F_{max}\uparrow$ & $\SF_{max} \uparrow$ & $\E_m \uparrow$  \\
    \midrule
        EDN (A) & ResNet50 & \textbf{0.0514} & 0.0484 & 0.9292  & 0.7529 & 0.8224 & 0.8117 & 0.8798 & 0.8514 \\
        \textbf{+ \textit{Ours}} (B) & 0 & 0.2693 & 0.2305 & 0.9098 & 0.4231 & \textbf{0.8708} & 0.7104 & \textbf{0.9231} & 0.564 \\
        \textbf{+ \textit{Ours}} (C) & 1 & 0.0642 & 0.0555 & 0.9284 & 0.7442 & 0.8239 & \textbf{0.8190} & 0.9048 & 0.8508 \\
        \textbf{+ \textit{Ours}} (D) & $\frac{S^{fore}}{S^{back}}$ & 0.0557 & \textbf{0.0483} & \textbf{0.9382} & \textbf{0.7544} & 0.8381 & 0.8163 & 0.8912 & \textbf{0.8594} \\
    \bottomrule
    \end{tabular}
    }
    \label{tab:ablation_alpha_DUT-OMRON}
\end{table*}
\begin{table*}[!h]
    \centering
    \caption{Ablation on the parameter $\alpha$ on HKU-IS(4,447 images). The best results are marked in bold.}
    \scalebox{0.8}{
    \begin{tabular}{c|c|cccccccc}
    \toprule
        Methods & $\alpha$ & $\MAE \downarrow $ & $\SMAE \downarrow$ & $\AUC \uparrow $ & $\F_m^\beta \uparrow$ & $\SF_m^{\beta} \uparrow $ & $\F_{max}\uparrow$ & $\SF_{max} \uparrow$ & $\E_m \uparrow$  \\
    \midrule
        EDN (A) & ResNet50 & \textbf{0.0279} & 0.0294 & 0.9750 & \textbf{0.9004} & 0.9017 & \textbf{0.9417} & 0.9364 & 0.9429 \\
        \textbf{+ \textit{Ours}} (B) & 0 & 0.1822 & 0.1431 & 0.9613 & 0.6027 & \textbf{0.9132} & 0.8921 & \textbf{0.9541} & 0.7100 \\
        \textbf{+ \textit{Ours}} (C) & 1 & 0.0383 & 0.0364 & 0.9760 & 0.888 & 0.9007 & 0.9377 & 0.9478 & 0.9347 \\
        \textbf{+ \textit{Ours}} (D) & $\frac{S^{fore}}{S^{back}}$ & 0.0287 & \textbf{0.0289} & \textbf{0.9776} & 0.8986 & 0.9072 & 0.9375 & 0.9443 & \textbf{0.9442} \\
    \bottomrule
    \end{tabular}
    }
    \label{tab:ablation_alpha_HKU-IS}
\end{table*}

\subsection{Time Cost Comparison} \label{time_cost}
Size-invariant optimization generally modifies the computation process of the original loss functions without bringing in too much computational burden. According to \cref{eq:loss}:
\begin{equation}
     \mathcal{L} _{\mathsf{Sl}}(f)=\sum _{k=1}^{K}\ell(f _k^{fore})+\alpha\ell(f _{K+1}^{back}).
\end{equation}
If the time cost of the original loss functions is $\mathcal{O}(t)$, then the theoretical time cost for size-invariant optimization will almost be $\mathcal{O}((K+1)\bar{t})$, where $\mathcal{O}(\bar{t}$ is the average time cost of processing one frame. Particularly, as it is common practice to train the SOD model on dataset DUTS-TR, we report the average $\bar{K}$ over the dataset: $\bar{K} \approx 1.21$.  Also, we report the practical training time of different backbones when applying our method.

\begin{table*}[!ht]
    \centering
    \caption{Practical training cost per epoch. The results are displayed as mean $\pm$ std., with 'seconds' as the unit.}
    \begin{tabular}{c|l|l}
    \toprule
        Backbone & Original optimization & SI optimization \\ 
    \midrule
        EDN & 340.0 $\pm$ 3.3s & 543.8$\pm$0.7s \\ 
        PoolNet & 523.5$\pm$1.1s & 690.2$\pm$1.5s \\ 
        ICON & 162.0$\pm$0.5s & 340.1$\pm$0.1s \\ 
        GateNet & 561.2$\pm$0.8s & 1270.2$\pm$35.3s \\ 
        LDF & 109.2$\pm$0.6s & 244.5$\pm$1.4s \\
    \bottomrule
    \end{tabular}
\end{table*}

It is noteworthy that the time cost of calculating the connected components of the image is not included in the training process. All the calculations can be completed during the data pre-process. The pre-process mainly consists of two stages:

(a) Calculating the connected components of the image.

(b) Generating the weight mask according to the bounding boxes for components.

Here we display the practical pre-process time on some representative datasets, which shows that the bounding boxes and connected components can be obtained with acceptable efficiency.

\begin{table}[!ht]
    \centering
    \caption{Practical pre-process time for each dataset.}
    \begin{tabular}{l|l|l|l}
    \toprule
        Dataset & Stage(a) & Stage(b) & Total \\
    \midrule
        DUTS-TE(5,019) & 474.0s & 188.2s & 658.2s \\ 
        ECSSD(1,000) & 91.8s & 40.5s & 132.3s \\ 
        DUT-OMRON(5,168) & 470.7s & 220.1s & 690.8s \\ 
        HKU-IS(4,447) & 508.8s & 190.1s & 698.9s \\ 
        XPIE(10,000) & 1432.2s & 316.5s & 1748.7s \\
    \bottomrule
    \end{tabular}
\end{table}

\stopcontents[sections]
\end{document}